\pdfoutput=1   
\documentclass[runningheads]{llncs}

\usepackage{eccv}
\usepackage{lmodern}            

\usepackage{tikz}
\usetikzlibrary{shapes, arrows.meta, positioning, calc, fit, backgrounds}
\usepackage{eccvabbrv}
\usepackage{graphicx}
\usepackage{booktabs}
\usepackage{amsmath,amssymb}
\usepackage[inline]{enumitem}
\usepackage{subcaption}
\usepackage[accsupp]{axessibility}
\usepackage{xcolor}
\usepackage{multirow}
\usepackage[protrusion=true,expansion=false]{microtype} 
\usepackage{placeins}           
\usepackage{colortbl}           
\usepackage{tcolorbox}
\tcbuselibrary{listings,breakable}

\usepackage[breaklinks,colorlinks,citecolor=eccvblue]{hyperref}

\usepackage{orcidlink}
\usepackage{marvosym}           

\definecolor{placeholderBg}{RGB}{240,240,240}
\definecolor{planblue}{RGB}{31,119,180}
\definecolor{groundorange}{RGB}{214,125,36}
\definecolor{stagegreen}{RGB}{44,160,44}
\definecolor{guideG}{RGB}{219,98,128}   
\definecolor{guideU}{RGB}{70,120,210}   
\definecolor{guideI}{RGB}{80,186,150}   
\definecolor{guideD}{RGB}{155,115,195}  
\definecolor{guideE}{RGB}{220,180,40}   

\newcommand{\plan}{\textbf{\textsc{Planning}}}
\newcommand{\ground}{\textbf{\textsc{Grounding}}}
\newcommand{\guidenameC}{\textbf{\textcolor{guideG}{G}\textcolor{guideU}{U}\textcolor{guideI}{I}\textcolor{guideD}{D}\textcolor{guideE}{E}}}
\newcommand{\guidename}{\textbf{\textsc{Guide}}}

\newtcblisting{promptbox}{%
  colback=placeholderBg,
  colframe=placeholderBg,
  listing only,
  listing options={basicstyle=\footnotesize\ttfamily, columns=fullflexible, keepspaces=true},
  boxrule=0pt,
  arc=4pt,
  left=4pt, right=4pt, top=3pt, bottom=3pt,
  breakable,
}

\begin{document}

\title{\guidenameC{}: Resolving Domain Bias in GUI Agents through Real-Time Web Video Retrieval and Plug-and-Play Annotation}

\titlerunning{GUIDE}

\author{
  Rui Xie\inst{1,2},\;
  Zhi Gao\inst{2,3}\textsuperscript{\Letter},\;
  Chenrui Shi\inst{2,3},\;
  Zirui Shang\inst{2,3},\;
  Lu Chen\inst{1}\textsuperscript{\Letter},\\
  Qing Li\inst{2}\textsuperscript{\Letter}
}

\authorrunning{R. Xie et al.}

\institute{
  Shanghai Jiao Tong University, Shanghai, China \and
  State Key Laboratory for General Artificial Intelligence, BIGAI, Beijing, China \and
  Beijing Institute of Technology, Beijing, China\\[3pt]
  \normalfont\small \url{https://sharryXR.github.io/GUIDE/}
}

{%
\renewcommand{\thefootnote}{\Letter}%
\footnotetext{Corresponding authors.}%
}

\maketitle

\begin{abstract}
Large vision-language models have endowed GUI agents with strong general capabilities for interface understanding and interaction.
However, due to insufficient exposure to domain-specific software operation data during training, these agents exhibit significant \emph{domain bias}---they lack familiarity with the specific operation workflows (\emph{planning}) and UI element layouts (\emph{grounding}) of particular applications, limiting their real-world task performance.
In this paper, we present \guidenameC{} (\textbf{\textcolor{guideG}{G}}UI \textbf{\textcolor{guideU}{U}}nbiasing via \textbf{\textcolor{guideI}{I}}nstructional-Video \textbf{\textcolor{guideD}{D}}riven \textbf{\textcolor{guideE}{E}}xpertise), a training-free, plug-and-play framework that resolves GUI agent domain bias by autonomously acquiring domain-specific expertise from web tutorial videos through a retrieval-augmented automated annotation pipeline.
\guidename{} introduces two key innovations.
First, a \emph{subtitle-driven Video-RAG pipeline} unlocks video semantics through subtitle analysis, performing progressive three-stage retrieval---domain classification, topic extraction, and relevance matching---to identify task-relevant tutorial videos.
Second, a \emph{fully automated VLM pairwise video annotation pipeline} feeds consecutive keyframes enhanced with UI element detection into VLMs,
inferring the required \emph{planning} and \emph{grounding} knowledge that are injected into the agent's corresponding modules to address both manifestations of domain bias.
Extensive experiments on OSWorld demonstrate \guidename{}'s generality as a plug-and-play component for both multi-agent systems and single-model agents.
It yields \textbf{+4.47--+7.48 percentage-point} improvements and reduces execution steps---without modifying any model parameters or architecture---validating \guidename{} as an architecture-agnostic enhancement to bridge GUI agent domain bias.
Code, dataset, and additional results are available on our project page: \url{https://sharryXR.github.io/GUIDE/}.
\end{abstract}
\section{Introduction}
\label{sec:intro}

The rapid advancement of large vision-language models (VLMs)~\cite{gpt4o,gpt5,qwen3vl,claude46} has enabled a new generation of GUI agents~\cite{agents,cogagent,seeclick,cradle,ufo,sport,mmagenttuning} capable of autonomously operating desktop and mobile applications through visual perception and natural language reasoning.
These agents can interpret screenshots, understand task instructions, and generate executable actions such as clicks, keystrokes, and scrolls, demonstrating impressive \emph{general} capabilities in interface understanding, complex reasoning, and instruction following. However, when confronted with specific software tasks in real-world scenarios, current GUI agents exhibit significant \emph{domain bias}: 
they lack familiarity with the \textbf{planning-level} and \textbf{grounding-level} domain knowledge.

\begin{figure*}[tb]
\centering
\resizebox{\textwidth}{!}{%
\begin{tikzpicture}[
    x=1cm, y=1cm,
    font=\sffamily,
    badge/.style={fill=carrow, text=white, font=\bfseries\fontsize{8pt}{8pt}\selectfont, rounded corners=5pt, inner sep=4pt, align=center},
    stagebadge/.style={fill=cstage, text=white, font=\bfseries\fontsize{6.5pt}{7pt}\selectfont, rounded corners=4pt, inner sep=2.5pt},
    searchbar/.style={draw=carrow, fill=white, rounded corners=3pt, inner sep=3pt, font=\ttfamily\fontsize{6.5pt}{7pt}\selectfont, text=ctext},
    card/.style={draw={rgb,255:red,204;green,204;blue,204}, fill=white, rounded corners=2pt, inner sep=2pt},
    videoicon/.style={fill=cfail, rounded corners=1pt},
    ccicon/.style={draw=carrow, fill=white, rounded corners=0.5pt, inner sep=1pt, font=\fontsize{4.5pt}{4.5pt}\sffamily\bfseries, text=carrow},
    planbox/.style={draw=cplan, fill=cbgplan, rounded corners=4pt, line width=1pt},
    groundbox/.style={draw=cground, fill=cbgground, rounded corners=4pt, line width=1pt},
    arrowmain/.style={-{Triangle[length=5pt,width=5pt]}, line width=1.5pt, carrow},
    arrowplan/.style={-{Triangle[length=5pt,width=5pt]}, line width=1.5pt, draw=cplan, fill=cplan},
    arrowground/.style={-{Triangle[length=5pt,width=5pt]}, line width=1.5pt, draw=cground, fill=cground},
    subtext/.style={font=\fontsize{6.5pt}{7pt}\selectfont, text=ctext},
    mutedtext/.style={font=\fontsize{6pt}{6.5pt}\selectfont, text=cmuted}
]

\definecolor{cplan}{RGB}{31,119,180}
\definecolor{cground}{RGB}{214,125,36}
\definecolor{cstage}{RGB}{44,160,44}
\definecolor{cfail}{RGB}{220,50,32}
\definecolor{carrow}{RGB}{68,68,68}
\definecolor{ctext}{RGB}{26,26,26}
\definecolor{cmuted}{RGB}{136,136,136}
\definecolor{csub}{RGB}{23,114,131}
\definecolor{cbgplan}{RGB}{235,243,250}
\definecolor{cbgground}{RGB}{253,243,231}
\definecolor{bgcol1}{RGB}{242,251,242}
\definecolor{bgcol2}{RGB}{250,250,250}

\fill[bgcol1, rounded corners=6pt] (0, 0) rectangle (5.4, 7.5);
\node[badge] at (2.7, 7.5) {(1) Retrieval Agent};

\node[mutedtext, anchor=center] at (2.7, 7.1) {LLM generates query $\to$ YouTube Search};
\node[searchbar, text width=4.5cm, align=center] (search) at (2.7, 6.7) {\textcolor{cfail}{$\blacktriangleright$} How to adjust brightness in GIMP};
\node[font=\fontsize{6.5pt}{7pt}\bfseries\selectfont, text=cfail, anchor=center] at (2.7, 6.2) {50 Candidate YouTube Videos};

\fill[cstage, opacity=0.1, rounded corners=2pt] (0.8, 5.9) -- (4.6, 5.9) -- (3.7, 1.35) -- (1.7, 1.35) -- cycle;

\node[stagebadge] at (2.7, 5.5) {Stage 1: Domain Classification};
\node[ccicon, anchor=east] at (1.6, 5.05) {CC};
\node[subtext, anchor=west, align=left] at (1.65, 5.05) {\textcolor{csub}{\textit{"...click on}}\\[-0.5mm]\textcolor{csub}{\textit{Colors menu..."}}};
\node[subtext, anchor=center] at (2.7, 4.6) {$\hookrightarrow$ \textbf{GUI Demo} \textcolor{cstage}{$\checkmark$} \textcolor{cmuted}{(Filter)}};

\node[stagebadge] at (2.7, 4.1) {Stage 2: Topic Extraction};
\node[ccicon, anchor=center] at (1.65, 3.7) {TITLE};
\node[subtext, align=center] at (1.65, 3.25) {\textcolor{cmuted}{\textit{"GIMP}}\\[-0.5mm]\textcolor{cmuted}{\textit{Tutorial}}\\[-0.5mm]\textcolor{cmuted}{\textit{2024"}}};
\node[font=\fontsize{9pt}{9pt}\selectfont, text=carrow, anchor=center] at (2.7, 3.4) {$+$};
\node[font=\fontsize{9pt}{9pt}\selectfont, text=cstage, anchor=center] at (2.7, 2.75) {$\Downarrow$};
\node[font=\fontsize{6pt}{6.5pt}\bfseries\selectfont, text=cstage, anchor=center, align=center] at (2.7, 2.45) {Topic: \textit{"Adjust brightness via Colors"}};
\node[ccicon, anchor=center] at (3.75, 3.7) {CC};
\node[subtext, align=center] at (3.75, 3.25) {\textcolor{csub}{\textit{"...adjust}}\\[-0.5mm]\textcolor{csub}{\textit{brightness}}\\[-0.5mm]\textcolor{csub}{\textit{via Colors"}}};

\node[stagebadge] at (2.7, 1.9) {Stage 3: Relevance Matching};
\node[mutedtext, anchor=east] at (2.6, 1.55) {Top-K (K$\le$2)};
\draw[carrow, opacity=0.2, line width=3pt, line cap=round] (2.8, 1.55) -- (4.2, 1.55);
\draw[cstage, line width=3pt, line cap=round] (2.8, 1.55) -- (3.8, 1.55);
\draw[carrow, dashed, line width=0.5pt] (3.5, 1.35) -- (3.5, 1.75);

\node[card, minimum width=10mm, minimum height=7mm] (thumb1) at (2.1, 0.7) {};
\fill[videoicon] (1.75, 0.55) -- (1.75, 0.85) -- (2.0, 0.7) -- cycle;
\node[fill=cstage, text=white, inner sep=1pt, rounded corners=1pt, font=\fontsize{5pt}{5pt}\bfseries\selectfont, anchor=south east] at (thumb1.south east) {0.92};
\node[font=\fontsize{6pt}{6pt}\bfseries\selectfont, text=ctext] at (2.1, 0.2) {Video 1};

\node[card, minimum width=10mm, minimum height=7mm] (thumb2) at (3.3, 0.7) {};
\fill[videoicon] (2.95, 0.55) -- (2.95, 0.85) -- (3.2, 0.7) -- cycle;
\node[fill=cstage, text=white, inner sep=1pt, rounded corners=1pt, font=\fontsize{5pt}{5pt}\bfseries\selectfont, anchor=south east] at (thumb2.south east) {0.78};
\node[font=\fontsize{6pt}{6pt}\bfseries\selectfont, text=ctext] at (3.3, 0.2) {Video 2};

\fill[bgcol2, rounded corners=6pt] (6.0, 0) rectangle (11.4, 7.5);
\node[badge] at (8.7, 7.5) {(2) Annotation Agent};

\node[fill=white, draw=carrow!30, rounded corners=2pt, minimum width=2.2cm, minimum height=1.4cm, inner sep=0, outer sep=0] (win1) at (7.3, 6.3) {};
\fill[carrow!15, rounded corners=2pt] ([yshift=-4pt]win1.north west) rectangle (win1.north east);
\fill[cfail] ([xshift=4pt, yshift=-2pt]win1.north west) circle (0.8pt);
\fill[cground] ([xshift=8pt, yshift=-2pt]win1.north west) circle (0.8pt);
\fill[cstage] ([xshift=12pt, yshift=-2pt]win1.north west) circle (0.8pt);
\fill[carrow!5, rounded corners=1pt] ([yshift=0pt]win1.south west) rectangle ([xshift=6pt, yshift=-4pt]win1.north west);
\fill[carrow!20] (6.7, 6.0) rectangle (7.9, 6.7);
\fill[white] (6.9, 6.5) circle (1.5pt); 
\fill[white] (7.3, 6.0) -- (7.6, 6.4) -- (7.9, 6.0) -- cycle; 
\fill[white] (6.7, 6.0) -- (7.1, 6.5) -- (7.5, 6.0) -- cycle; 
\fill[carrow!20] (6.7, 5.8) rectangle (7.4, 5.85);
\fill[carrow!20] (6.7, 5.7) rectangle (7.8, 5.75);
\fill[carrow] (7.3, 6.1) -- (7.45, 5.95) -- (7.3, 5.95) -- cycle; 
\node[font=\fontsize{7pt}{7pt}\itshape\selectfont, text=carrow] at (7.3, 5.4) {$s_t$ (Before)};

\node[fill=white, draw=carrow!30, rounded corners=2pt, minimum width=2.2cm, minimum height=1.4cm, inner sep=0, outer sep=0] (win2) at (10.1, 6.3) {};
\fill[carrow!15, rounded corners=2pt] ([yshift=-4pt]win2.north west) rectangle (win2.north east);
\fill[cfail] ([xshift=4pt, yshift=-2pt]win2.north west) circle (0.8pt);
\fill[cground] ([xshift=8pt, yshift=-2pt]win2.north west) circle (0.8pt);
\fill[cstage] ([xshift=12pt, yshift=-2pt]win2.north west) circle (0.8pt);
\fill[carrow!5, rounded corners=1pt] ([yshift=0pt]win2.south west) rectangle ([xshift=6pt, yshift=-4pt]win2.north west);
\fill[carrow!20] (9.5, 6.0) rectangle (10.7, 6.7);
\fill[white] (9.7, 6.5) circle (1.5pt);
\fill[white] (10.1, 6.0) -- (10.4, 6.4) -- (10.7, 6.0) -- cycle;
\fill[white] (9.5, 6.0) -- (9.9, 6.5) -- (10.3, 6.0) -- cycle;
\fill[carrow!20] (9.5, 5.8) rectangle (10.2, 5.85);
\fill[carrow!20] (9.5, 5.7) rectangle (10.6, 5.75);
\fill[white, draw=carrow!50, rounded corners=1pt] (9.7, 5.9) rectangle (10.5, 6.4);
\fill[carrow!15] (9.7, 6.25) rectangle (10.5, 6.4);
\draw[carrow!30, line width=1pt] (9.8, 6.05) -- (10.4, 6.05); 
\fill[cground] (10.2, 6.05) circle (1.2pt); 
\draw[cground, thick, fill=cground, fill opacity=0.2] (10.1, 5.95) rectangle (10.3, 6.15); 
\fill[carrow] (10.2, 6.05) -- (10.35, 5.9) -- (10.2, 5.9) -- cycle; 
\node[font=\fontsize{7pt}{7pt}\itshape\selectfont, text=carrow] at (10.1, 5.4) {$s_{t+1}$ (After)};

\draw[-{Triangle[length=4pt,width=4pt]}, carrow, thick] (8.5, 6.3) -- node[above, font=\fontsize{6.5pt}{6.5pt}\bfseries\selectfont, text=carrow] {Action?} (8.9, 6.3);

\node[draw=carrow, dashed, rounded corners=4pt, fill=white, minimum width=5.0cm, minimum height=2.9cm] (idm) at (8.7, 3.65) {};
\node[font=\fontsize{8pt}{8pt}\itshape\bfseries\selectfont, fill=white, inner sep=1.5pt, text=ctext] at (8.7, 5.15) {$f_{VPA}$};


\node[fill=white, draw=carrow!50, minimum width=5mm, minimum height=3.5mm, rounded corners=1pt, inner sep=0, outer sep=0, label={[font=\fontsize{4.5pt}{4.5pt}\selectfont, text=carrow, yshift=0.5pt]below:$s_t$}] (mw1) at (6.6, 4.6) {};
\fill[carrow!20, rounded corners=1pt] ([yshift=-1.2mm]mw1.north west) rectangle (mw1.north east);
\fill[cfail] ([xshift=1pt, yshift=-0.6mm]mw1.north west) circle (0.4pt);
\fill[cground] ([xshift=2.5pt, yshift=-0.6mm]mw1.north west) circle (0.4pt);
\fill[cstage] ([xshift=4pt, yshift=-0.6mm]mw1.north west) circle (0.4pt);

\node[fill=white, draw=carrow!50, minimum width=5mm, minimum height=3.5mm, rounded corners=1pt, inner sep=0, outer sep=0, label={[font=\fontsize{4.5pt}{4.5pt}\selectfont, text=carrow, yshift=0.5pt]below:$s_{t+1}$}] (mw2) at (7.4, 4.6) {};
\fill[carrow!20, rounded corners=1pt] ([yshift=-1.2mm]mw2.north west) rectangle (mw2.north east);
\fill[cfail] ([xshift=1pt, yshift=-0.6mm]mw2.north west) circle (0.4pt);
\fill[cground] ([xshift=2.5pt, yshift=-0.6mm]mw2.north west) circle (0.4pt);
\fill[cstage] ([xshift=4pt, yshift=-0.6mm]mw2.north west) circle (0.4pt);

\draw[-stealth, carrow, line width=0.5pt] ([xshift=1pt]mw2.north east) -- ++(0.15,0) |- (8.4, 4.4);

\node[font=\fontsize{6pt}{6.5pt}\bfseries\selectfont, text=cground, align=center] (et) at (6.6, 3.8) {$E_t,$\\$E_{t+1}$};

\node[fill=none, minimum width=12mm, minimum height=8mm, outer sep=0] (et_bg) at (7.5, 3.8) {};
\draw[cground, fill=white, rounded corners=0.5pt] 
    ([xshift=-6mm,yshift=-4mm]et_bg.center) -- 
    ([xshift=-6mm,yshift=4mm]et_bg.center) -- 
    ([xshift=3mm,yshift=4mm]et_bg.center) -- 
    ([xshift=6mm,yshift=1mm]et_bg.center) -- 
    ([xshift=6mm,yshift=-4mm]et_bg.center) -- cycle;
\draw[cground, rounded corners=0.5pt] 
    ([xshift=3mm,yshift=4mm]et_bg.center) -- 
    ([xshift=3mm,yshift=1mm]et_bg.center) -- 
    ([xshift=6mm,yshift=1mm]et_bg.center);
\node[align=left, font=\fontsize{5pt}{5.5pt}\selectfont, text=cground] at (et_bg.center) {
    \textbf{elem $i$} \\[-0.3mm]
    \itshape desc $i$...
};

\draw[cground, dashed, rounded corners=2pt, line width=0.5pt] (6.2, 3.3) rectangle (8.2, 4.3);
\draw[-stealth, carrow, line width=0.5pt] (8.2, 3.8) -- (8.4, 3.8);

\node[ccicon, font=\fontsize{4.5pt}{4.5pt}\sffamily\bfseries, anchor=west] (cc2) at (6.3, 2.9) {CC};
\node[fill=cstage, text=white, rounded corners=2pt, font=\fontsize{5.5pt}{6pt}\bfseries\selectfont, inner sep=1.5pt, anchor=west] (ttop) at (6.9, 2.9) {$T_{topic}$};
\node[draw=cstage, text=cstage, rounded corners=2pt, font=\fontsize{5.5pt}{6pt}\bfseries\selectfont, inner sep=1.5pt, anchor=west] (csub) at (6.9, 2.4) {$C_{sub}$};

\draw[-stealth, carrow, line width=0.5pt] (ttop.east) -| (8.25, 3.2) -- (8.4, 3.2);
\draw[carrow, line width=0.5pt] (csub.east) -| (8.25, 3.2);

\node[draw=carrow, fill={rgb,255:red,245;green,245;blue,245}, rounded corners=2pt, minimum width=10mm, minimum height=10mm] (vlm) at (8.9, 3.8) {};
\node[font=\fontsize{5.5pt}{6pt}\bfseries\selectfont, text=carrow] at (8.9, 3.8) {VLM};

\node[card, font=\ttfamily\fontsize{5.5pt}{6pt}\selectfont, align=left, anchor=west, inner sep=1.5pt] (jsonout) at (9.65, 3.8) {
\textbf{\{Meaningful:}\textcolor{cstage}{$\checkmark$}\\
 \textbf{Thght\&Act:}\\
 \hspace{1mm}\textcolor{cmuted}{1. Env Observ}\\
 \hspace{2mm}\textcolor{cmuted}{\& Elem Guess}\\
 \hspace{1mm}\textcolor{cmuted}{2. Action Plan}\\
 \hspace{1mm}\textcolor{cfail}{3. Grounded}\\
 \hspace{2mm}\textcolor{cfail}{Action(NLP)}\textbf{\}}
};
\draw[-stealth, carrow, line width=0.5pt] (vlm.east) -- (jsonout.west);

\node[draw=carrow, fill={rgb,255:red,245;green,245;blue,245}, rounded corners=4pt, font=\fontsize{7.5pt}{8pt}\bfseries\selectfont, inner sep=3pt, text=ctext] (llm) at (8.7, 1.9) {LLM Split};

\node[planbox, text width=2.5cm, align=left, inner sep=3pt] (pbox) at (7.35, 0.65) {\textbf{\fontsize{6.5pt}{7pt}\selectfont \textcolor{cplan}{Planning Knowledge}}\\[1mm]\fontsize{5.5pt}{7pt}\selectfont $\bullet$ Execution Flow\\$\bullet$ Key Considerations\\$\bullet$ Coordinate-free};

\node[groundbox, text width=2.5cm, align=left, inner sep=3pt] (gbox) at (10.05, 0.65) {\textbf{\fontsize{6.5pt}{7pt}\selectfont \textcolor{cground}{Grounding Knowledge}}\\[1mm]\fontsize{5.5pt}{7pt}\selectfont $\bullet$ $\le$15 UI Elements\\$\bullet$ Appearance/Pos\\$\bullet$ Predicted Func};

\draw[arrowplan] (llm.south) -- (pbox.north);
\draw[arrowground] (llm.south) -- (gbox.north);

\node[badge] at (14.7, 7.5) {(3) GUI Agent Integration};
\node[font=\fontsize{7.5pt}{8pt}\itshape\selectfont, text=cmuted] at (14.7, 6.9) {Agent-agnostic natural language knowledge};

\fill[cbgplan, rounded corners=4pt] (12.2, 4.5) rectangle (17.2, 6.5);
\node[font=\fontsize{8pt}{8pt}\bfseries\selectfont, text=ctext, anchor=north west] at (12.3, 6.4) {Mode A: Multi-Agent (AgentS3)};

\node[fill=white, draw={rgb,255:red,204;green,204;blue,204}, rounded corners=2pt, minimum width=2.3cm, minimum height=1.3cm, outer sep=0, inner sep=0] (worker) at (13.5, 5.4) {};
\fill[cplan, rounded corners=1pt] ([xshift=0.5pt, yshift=0.5pt]worker.south west) rectangle ([xshift=2.5pt, yshift=-0.5pt]worker.north west);
\node[font=\fontsize{6.5pt}{7pt}\selectfont, align=center, text=ctext] at (13.5, 5.75) {System Prompt};
\node[fill=cplan, text=white, rounded corners=2pt, font=\ttfamily\fontsize{5.5pt}{6pt}\selectfont, inner sep=1.5pt] at (13.5, 5.2) {\{video\_planning\}};

\node[fill=white, draw={rgb,255:red,204;green,204;blue,204}, rounded corners=2pt, minimum width=2.3cm, minimum height=1.3cm, outer sep=0, inner sep=0] (gndagent) at (15.9, 5.4) {};
\fill[cground, rounded corners=1pt] ([xshift=0.5pt, yshift=0.5pt]gndagent.south west) rectangle ([xshift=2.5pt, yshift=-0.5pt]gndagent.north west);
\node[font=\fontsize{6.5pt}{7pt}\selectfont, align=center, text=ctext] at (15.9, 5.75) {Coord. Prompt};
\node[fill=cground, text=white, rounded corners=2pt, font=\ttfamily\fontsize{5.5pt}{6pt}\selectfont, inner sep=1.5pt] at (15.9, 5.2) {\{video\_grounding\}};

\fill[cbgground, rounded corners=4pt] (12.2, 2.1) rectangle (17.2, 4.3);
\node[font=\fontsize{8pt}{8pt}\bfseries\selectfont, text=ctext, anchor=north west] at (12.3, 4.2) {Mode B: Single-Model Agent};

\node[fill=white, draw={rgb,255:red,204;green,204;blue,204}, rounded corners=2pt, minimum width=4.7cm, minimum height=1.5cm, outer sep=0, inner sep=0] (unified) at (14.7, 3.1) {};
\fill[cplan, rounded corners=1pt] ([xshift=0.5pt, yshift=0.5pt]unified.west) rectangle ([xshift=2.5pt, yshift=-0.5pt]unified.north west);
\fill[cground, rounded corners=1pt] ([xshift=0.5pt, yshift=0.5pt]unified.south west) rectangle ([xshift=2.5pt, yshift=-0.5pt]unified.west);

\node[font=\fontsize{6.5pt}{7pt}\selectfont, align=center, text=ctext] at (14.7, 3.6) {Unified System Prompt};
\node[fill=cplan, text=white, rounded corners=2pt, font=\ttfamily\fontsize{5.5pt}{6pt}\selectfont, inner sep=1.5pt] at (13.7, 3.2) {\{video\_planning\}};
\node[font=\fontsize{7pt}{7pt}\selectfont, text=ctext] at (14.7, 3.2) {+};
\node[fill=cground, text=white, rounded corners=2pt, font=\ttfamily\fontsize{5.5pt}{6pt}\selectfont, inner sep=1.5pt] at (15.7, 3.2) {\{video\_grounding\}};

\node[fill={rgb,255:red,255;green,230;blue,153}, draw={rgb,255:red,214;green,182;blue,0}, text=ctext, rounded corners=2pt, font=\fontsize{6pt}{6pt}\bfseries\selectfont, inner sep=1.5pt] at (14.7, 2.7) {Guided CoT Template};

\node[card, minimum width=15mm, minimum height=10mm] (vm) at (14.7, 0.9) {};
\draw[carrow, opacity=0.3] (14.1, 1.2) -- (15.3, 1.2) (14.1, 1.0) -- (14.6, 1.0) (14.1, 0.8) -- (14.8, 0.8);
\node[font=\fontsize{8pt}{8pt}\bfseries\selectfont, text=ctext] at (14.7, 0.2) {GUI Task Execution};

\draw[-{Triangle[length=5pt,width=5pt]}, carrow, line width=1.5pt] (14.7, 2.1) -- (14.7, 1.4);
\draw[-{Triangle[length=5pt,width=5pt]}, carrow, line width=1.5pt, rounded corners=4pt] (17.2, 5.5) -- (17.6, 5.5) -- (17.6, 0.9) -- (15.5, 0.9);

\node[font=\fontsize{7.5pt}{8pt}\itshape\selectfont, text=cmuted, anchor=south east] at (17.4, -0.3) {Reference, not directive};

\begin{pgfonlayer}{background}
    \draw[arrowmain, rounded corners=6pt] (4.0, 0.8) -- (5.6, 0.8) -- (5.6, 6.3) -- (5.9, 6.3);
    \node[font=\fontsize{7pt}{7pt}\bfseries\selectfont, text=carrow, fill=white, inner sep=1pt] at (5.6, 3.6) {\rotatebox{-90}{Retrieved Videos}};

    \draw[arrowplan, rounded corners=4pt] (pbox.east) -- (11.5, 0.65) -- (11.5, 4.5) -- (12.0, 4.5);
    \draw[arrowground, rounded corners=4pt] (gbox.east) -- (11.8, 0.65) -- (11.8, 4.3) -- (12.0, 4.3);
\end{pgfonlayer}
\node[font=\fontsize{7pt}{7pt}\bfseries\selectfont, text=cplan, fill=white, inner sep=1pt] at (11.5, 3.1) {\rotatebox{-90}{Planning}};
\node[font=\fontsize{7pt}{7pt}\bfseries\selectfont, text=cground, fill=white, inner sep=1pt] at (11.8, 2.4) {\rotatebox{-90}{Grounding}};

\end{tikzpicture}
}
\caption{\textbf{Overview of \guidenameC{}.}
  \textbf{(1)}~A \emph{Retrieval Agent} filters YouTube candidates via three subtitle-driven stages to select top-$K$ videos.
  \textbf{(2)}~An \emph{Annotation Agent} applies $f_{\mathrm{VPA}}$ (Eq.\,\ref{eq:pairwise-annotation}) on keyframe pairs $s_t$/$s_{t+1}$ with UI element graphs $E_t$/$E_{t+1}$, topic $T_{\mathrm{topic}}$, and subtitle context $C_{\mathrm{sub}}$, producing \textcolor{planblue}{planning} and \textcolor{groundorange}{grounding} knowledge.
  \textbf{(3)}~Knowledge is injected into the \emph{GUI Agent} in a plug-and-play manner---supporting multi-agent (Mode~A) and single-model (Mode~B) architectures.}
\label{fig:overview}
\end{figure*}

\begin{itemize}[leftmargin=1.5em,itemsep=3pt]
\item \textbf{Planning-level domain bias.}
The agent 
is unfamiliar with the specific operation workflows of a given application---it cannot accurately determine which concrete steps are needed, their sequential order, or which menus and panels to navigate.
For instance, an agent may understand the semantics of ``adjust image brightness'' but not know that in GIMP the correct path is \texttt{Colors\,$\to$\,Brightness-Contrast} rather than the \texttt{Image\,$\to$\,Adjustments} menu familiar from other editors (\eg, Adobe Photoshop).

\item \textbf{Grounding-level domain bias.}
The agent 
is unfamiliar with the UI layouts and control styles of a specific application---it struggles to locate task-relevant interactive elements.
For example, an agent recognizes the general concept of a ``menu'' yet cannot locate the \texttt{Colors} entry in the GIMP menu bar or identify the brightness slider within its dialog.
\end{itemize}

\noindent
This domain bias is fundamentally an \emph{alignment gap} between general model capabilities and domain-specific task requirements,
\emph{i.e.}, the model does not lack capability but lacks domain knowledge.
Conventional solutions such as manual dataset annotation, expert rule authoring, or domain-specific fine-tuning are costly~\cite{osworld}, narrow in coverage, and unable to keep pace with continuously evolving software interfaces.

In this paper we present \guidenameC{} (\textbf{\textcolor{guideG}{G}}UI \textbf{\textcolor{guideU}{U}}nbiasing via \textbf{\textcolor{guideI}{I}}nstructional-Video \textbf{\textcolor{guideD}{D}}riven \textbf{\textcolor{guideE}{E}}xpertise), a novel plug-and-play framework whose core objective is to enable \emph{autonomous learning} for GUI agents---leveraging the vast, continuously growing repository of GUI tutorial videos on the internet to bridge the domain bias, aligning general model capabilities to domain-specific task proficiency.

As illustrated in \cref{fig:overview}, \guidename{} orchestrates three collaborating agents:

\noindent
\textbf{(1)~Retrieval Agent} (\cref{sec:video-rag}).
Given a task instruction, the retrieval agent searches YouTube for candidate tutorial videos and applies a progressive three-stage subtitle-driven filtering pipeline---domain classification, topic extraction, and relevance matching---to precisely identify the most task-relevant GUI demonstration videos.

\noindent
\textbf{(2)~Annotation Agent} (\cref{sec:annotation}).
The annotation agent processes selected videos through a fully automated VLM pairwise video annotation pipeline: extracting keyframes, detecting UI elements, and inferring transferable annotation information that is then decomposed into structured \plan{} and \ground{} knowledge.


\noindent
\textbf{(3)~GUI Agent integration} (\cref{sec:integration}).
\plan{} knowledge is injected into the GUI agent's task planning module, and \ground{} knowledge is injected into the coordinate generation module, precisely targeting the two manifestations of domain bias without modifying the agent's architecture or model weights.

\smallskip
\noindent\textbf{Contributions.}
\begin{enumerate}[leftmargin=1.5em,itemsep=1pt,label=(\roman*)]
\item We propose an \emph{autonomous learning} paradigm for GUI agents that leverages internet video resources to bridge \emph{domain bias}, aligning general capabilities to domain-specific tasks without manual annotation or model fine-tuning.
\item We design a \emph{subtitle-driven Video-RAG pipeline} that exploits the unique subtitle modality of videos, achieving progressive content-level retrieval precision far beyond title-based keyword matching.
\item We construct a fully automated VLM pairwise video annotation pipeline with a \emph{transferable annotation format}, producing the \plan{} and \ground{} knowledge that targets the two manifestations of GUI agent domain bias.
\item We show \textbf{+4.47--+7.48 percentage-point} OSWorld~\cite{osworld} gains across three agent architectures, plus WindowsAgentArena (WAA)~\cite{windowsagentarena} transfer and a same-backbone Watch\,\&\,Learn-style raw-trajectory control, validating \guidename{} as a scalable plug-and-play solution.
\end{enumerate}

\section{Related Work}
\label{sec:related}

\paragraph{GUI Agents.}
Vision-language models have enabled GUI agents to operate across desktop~\cite{cogagent,agents,agents2,showui,dart}, web~\cite{webagent}, and mobile~\cite{seeclick,mobileenv} environments via direct visual perception and action generation.
However, OSWorld~\cite{osworld}---a benchmark of 369 real-world tasks spanning diverse applications---reveals that even state-of-the-art agents suffer severe performance drops on out-of-distribution software, exposing a persistent \emph{domain bias} that generalist training alone cannot overcome.

\paragraph{Retrieval-Augmented Generation for GUI Agents.}
RAG~\cite{rag} augments language models with external knowledge at inference time; ReAct~\cite{react} further integrates reasoning with acting.
Recent works extend this paradigm to GUI agents: RAG-GUI~\cite{raggui} retrieves web tutorials, KG-RAG~\cite{kgrag} constructs knowledge graphs from UI transitions for plan retrieval, and OS-Symphony~\cite{ossymphony} synthesizes visually grounded tutorials on demand.
A complementary line injects video-based knowledge without automated retrieval: Mobile-Agent-V~\cite{mobileagentv} and ShowUI-Aloha~\cite{showuialoha} feed pre-assigned demonstration videos into agents, but rely on \emph{manually curated, task-specific} assignments that do not scale.
None of these retrieves and processes \emph{raw tutorial videos from the open web} at test time---the core capability of \guidename{}.

\paragraph{Video Materials for GUI Agents.}
Screen recordings and tutorial videos are increasingly exploited as data sources for GUI agents.
Large-scale datasets such as AITW~\cite{aitw} and GUI-World~\cite{guiworld} provide annotated interaction episodes, while AssistGUI~\cite{assistgui} uses screen recordings for desktop task planning.
TongUI~\cite{tongui} and VideoAgentTrek~\cite{videoagenttrek} further mine YouTube tutorials at scale to synthesize GUI trajectories for model fine-tuning---the former via VLM zero-shot annotation, the latter via an inverse dynamics module with LLM-generated inner monologues.
Watch and Learn~\cite{watchlearn}, the most closely related work, trains a lightweight inverse dynamics model on consecutive frames, generates LLM reasoning traces, and supports both offline fine-tuning and inference-time retrieval as in-context exemplars.
\guidename{} differs in three respects:
\emph{(i)}~\textbf{live web retrieval}---\guidename{} searches YouTube in real time for each task, whereas Watch and Learn retrieves from a pre-built offline corpus;
\emph{(ii)}~\textbf{subtitle-driven Video-RAG}---progressive semantic filtering (domain classification $\to$ topic extraction $\to$ relevance matching) via the subtitle modality, achieving content-level precision beyond keyword search;
\emph{(iii)}~\textbf{richer annotation}---a VLM annotator jointly consumes keyframe pairs with UI element graphs, video topic, and subtitle context to produce structured knowledge (strategic reasoning, visual element descriptions, coordinate-free workflow), rather than labeling every frame pair with a lightweight IDM.

\section{Method}
\label{sec:method}

\subsection{Overview}
\label{sec:overview}

Given a task instruction~$\mathcal{T}$ (\eg, ``set conditional formatting for column~B in LibreOffice Calc'') and the associated application name, \guidename{} automatically retrieves relevant GUI tutorial videos from the web, extracts transferable domain knowledge, and injects it into a GUI agent to bridge its domain bias for the specific task.
Architecturally, \guidename{} is a \emph{multi-agent collaborative system} comprising three specialized agents (\cref{fig:overview}):
a \emph{Retrieval Agent} that identifies task-relevant tutorial videos via subtitle-driven Video-RAG (\cref{sec:video-rag}),
an \emph{Annotation Agent} that extracts transferable knowledge through pairwise video annotation (\cref{sec:annotation}), and
the downstream \emph{GUI Agent} into which the extracted knowledge is injected in a plug-and-play manner (\cref{sec:integration}).

\subsection{Subtitle-Driven Video-RAG Pipeline}
\label{sec:video-rag}

Extending RAG~\cite{rag} from text to \emph{videos} poses unique challenges: video titles are noisy, candidate types are heterogeneous (\eg, lectures vs.\ hands-on tutorials), and the operational knowledge within frames is semantically opaque.
\guidename{} addresses these challenges through a key insight: \emph{subtitles} (including auto-generated captions) naturally contain narrated operation steps, UI element names, and task-purpose explanations, serving as a textual semantic bridge between video content and task requirements.
As illustrated in \cref{fig:rag-pipeline}, the pipeline progressively narrows 50+ candidates to at most $K{=}2$ relevant videos through initial pre-filtering followed by three subtitle-driven stages.

\begin{figure*}[tb]
\centering
\resizebox{\textwidth}{!}{%
\begin{tikzpicture}[
    x=1cm, y=1cm,
    font=\sffamily,
    badge/.style={fill=carrow, text=white, font=\bfseries\fontsize{9pt}{9pt}\selectfont, rounded corners=4pt, inner sep=4pt, align=center},
    stagebadge/.style={fill=cstage, text=white, font=\bfseries\fontsize{8.5pt}{8.5pt}\selectfont, rounded corners=3pt, inner sep=3pt},
    searchbar/.style={draw=carrow, fill=white, rounded corners=3pt, inner sep=4pt, font=\ttfamily\fontsize{8pt}{8pt}\selectfont, text=ctext},
    card/.style={draw={rgb,255:red,204;green,204;blue,204}, fill=white, rounded corners=2pt, inner sep=2pt},
    videoicon/.style={fill=cfail, rounded corners=1.5pt}, 
    ccicon/.style={draw=carrow, fill=white, rounded corners=1pt, inner sep=1.5pt, font=\fontsize{7pt}{7pt}\sffamily\bfseries, text=carrow},
    arrowmain/.style={-{Triangle[length=5pt,width=5pt]}, line width=1.5pt, carrow},
    arrowsmall/.style={-{Triangle[length=4pt,width=4pt]}, line width=1.0pt, carrow}, 
    subtext/.style={font=\fontsize{8pt}{9pt}\selectfont, text=ctext},
    mutedtext/.style={font=\fontsize{7.5pt}{8.5pt}\selectfont, text=cmuted, align=center, text width=1cm, inner sep=1pt},
    stagebox/.style={draw=cstage, fill=white, rounded corners=4pt, line width=1pt},
    cc_label/.style={fill=cstage!15, text=cstage, rounded corners=2pt, font=\fontsize{7.5pt}{8pt}\bfseries\selectfont, inner sep=2.5pt},
    passbox/.style={draw=cstage, fill=bgcol1, rounded corners=2pt, line width=0.5pt},
    failbox/.style={draw=cfail, fill={rgb,255:red,255;green,245;blue,245}, rounded corners=2pt, line width=0.5pt},
    promptbox_mono/.style={fill=placeholderBg, rounded corners=2pt, inner sep=3pt, font=\ttfamily\fontsize{7.5pt}{9pt}\selectfont, text=carrow, align=center}
]

\definecolor{cplan}{RGB}{31,119,180}
\definecolor{cground}{RGB}{214,125,36}
\definecolor{cstage}{RGB}{44,160,44}
\definecolor{cfail}{RGB}{220,50,32}
\definecolor{carrow}{RGB}{68,68,68}
\definecolor{ctext}{RGB}{26,26,26}
\definecolor{cmuted}{RGB}{136,136,136}
\definecolor{bgcol1}{RGB}{242,251,242}
\definecolor{bgcol2}{RGB}{250,250,250}
\definecolor{placeholderBg}{RGB}{240,240,240}

\fill[bgcol1, rounded corners=6pt] (0, 0.0) -- (0, 6.5) -- (19.4, 5.5) -- (19.4, 1.0) -- cycle;
\fill[bgcol1!50, rounded corners=4pt] (0.2, 0.2) -- (0.2, 6.3) -- (19.2, 5.3) -- (19.2, 1.2) -- cycle;

\fill[bgcol2!80, rounded corners=4pt] (0, -0.6) rectangle (19.4, -0.1);
\node[ccicon, anchor=west] (ccicon_bot) at (0.3, -0.35) {CC};
\node[text=carrow!90, font=\itshape\fontsize{8.5pt}{8.5pt}\selectfont, anchor=west] at (ccicon_bot.east) {Subtitles serve as a continuous textual semantic bridge between raw pixels and task requirements.};

\node[searchbar, minimum width=2.6cm, minimum height=0.7cm] at (1.4, 5.2) {};
\fill[videoicon] (0.2, 5.05) rectangle (0.55, 5.35);
\fill[white] (0.3, 5.12) -- (0.45, 5.2) -- (0.3, 5.28) -- cycle;
\node[font=\ttfamily\fontsize{6.5pt}{7.5pt}\selectfont, text=ctext, anchor=west, align=left] at (0.6, 5.2) {How to adjust\\[-0.5mm]brightness...};

\node[card, minimum width=2.6cm, minimum height=1.6cm] at (1.3, 3.3) {};
\node[card, minimum width=2.6cm, minimum height=1.6cm] at (1.35, 3.25) {};
\node[card, minimum width=2.6cm, minimum height=1.6cm] (stack) at (1.4, 3.2) {};
\fill[videoicon] ([xshift=-16pt, yshift=-11pt]stack.center) rectangle ([xshift=16pt, yshift=11pt]stack.center);
\fill[white] ([xshift=-5pt, yshift=-6pt]stack.center) -- ([xshift=8pt, yshift=0pt]stack.center) -- ([xshift=-5pt, yshift=6pt]stack.center) -- cycle;

\node[stagebadge] at (1.4, 4.3) {Initial};
\node[subtext, text width=2.6cm, align=center] at (1.4, 2.1) {\textbf{50+ candidates}};

\draw[arrowmain] (2.8, 3.2) -- node[above=2pt, font=\fontsize{8.5pt}{8.5pt}\bfseries\selectfont, text=carrow] {Pre-filter} node[below=3pt, mutedtext, align=center, text width=1.7cm] {Duration $<$ 3000s\\[1mm]Valid title} (4.6, 3.2);
\node[stagebox, minimum width=3.6cm, minimum height=4.8cm] (s1box) at (6.4, 3.2) {};
\node[cc_label] at (6.4, 5.8) {Subtitle-driven}; 
\node[stagebadge] at (6.4, 5.3) {Stage 1};
\node[font=\fontsize{9pt}{9pt}\bfseries\selectfont, text=ctext] at (6.4, 4.8) {Domain Classify};

\node[mutedtext, anchor=west, text width=1cm] at (4.7, 4.5) {PASS:}; 
\node[passbox, minimum width=3.2cm, minimum height=1.2cm] (pbox) at (6.4, 3.75) {}; 
\node[ccicon, anchor=north west] at ([xshift=4pt, yshift=-4pt]pbox.north west) {CC}; 
\node[subtext, anchor=north west, text width=2.4cm, align=left] at ([xshift=22pt, yshift=-4pt]pbox.north west) {\textit{...click on the colors menu...}};
\node[fill=cstage, text=white, font=\fontsize{6pt}{6pt}\bfseries\selectfont, rounded corners=2pt, inner sep=2pt, anchor=south east] at ([xshift=-4pt, yshift=4pt]pbox.south east) {GUI Demo $\checkmark$};

\node[mutedtext, anchor=west, text width=1cm] at (4.7, 2.9) {FAIL:}; 
\node[failbox, minimum width=3.2cm, minimum height=1.7cm] (fbox) at (6.4, 1.85) {}; 
\node[ccicon, anchor=north west] at ([xshift=4pt, yshift=-4pt]fbox.north west) {CC}; 
\node[subtext, anchor=north west, text width=2.4cm, align=left] at ([xshift=22pt, yshift=-4pt]fbox.north west) {\textit{...welcome to my vlog...}\\[1.5mm]\textcolor{cmuted}{(No GUI steps)}};
\node[fill=cfail, text=white, font=\fontsize{6pt}{6pt}\bfseries\selectfont, rounded corners=2pt, inner sep=2pt, anchor=south east] at ([xshift=-4pt, yshift=4pt]fbox.south east) {Non-GUI $\times$};

\draw[arrowmain] (8.2, 3.2) -- node[above=1pt, mutedtext] {~\textasciitilde15 $\to$ \\ \textasciitilde8 videos} (9.2, 3.2);

\node[stagebox, minimum width=2.8cm, minimum height=4.8cm] (s2box) at (10.6, 3.2) {};
\node[cc_label] at (10.6, 5.8) {Subtitle-driven};
\node[stagebadge] at (10.6, 5.3) {Stage 2};
\node[font=\fontsize{9pt}{9pt}\bfseries\selectfont, text=ctext] at (10.6, 4.8) {Topic Extract};

\node[draw=carrow!50, fill=placeholderBg, rounded corners=2pt, minimum width=2.4cm, minimum height=0.6cm] at (10.6, 3.9) {};
\node[font=\fontsize{7.5pt}{7.5pt}\bfseries\selectfont, text=carrow] at (10.6, 3.9) {TITLE};

\node[draw=cstage, fill=bgcol1, rounded corners=2pt, minimum width=2.4cm, minimum height=0.6cm] at (10.6, 3.1) {};
\node[font=\fontsize{7.5pt}{7.5pt}\bfseries\selectfont, text=cstage] at (10.6, 3.1) {CC (Subtitle)};

\draw[arrowsmall] (10.6, 2.7) -- (10.6, 2.3);

\node[draw=cstage, fill=bgcol1, rounded corners=3pt, line width=1pt, minimum width=2.4cm, minimum height=0.9cm] at (10.6, 1.7) {};
\node[subtext, font=\fontsize{7.5pt}{8.5pt}\bfseries\selectfont, text=cstage, text width=2.2cm, align=center] at (10.6, 1.7) {Topic: Adjust brightness via Colors};

\draw[arrowmain] (12.0, 3.2) -- node[above=1pt, mutedtext] {~\textasciitilde8 $\to$ \\ \textasciitilde8 videos} (13.0, 3.2);

\node[stagebox, minimum width=3.2cm, minimum height=4.8cm] (s3box) at (14.6, 3.2) {};
\node[cc_label] at (14.6, 5.8) {Subtitle-driven};
\node[stagebadge] at (14.6, 5.3) {Stage 3};
\node[font=\fontsize{9pt}{9pt}\bfseries\selectfont, text=ctext] at (14.6, 4.8) {Relevance};

\node[promptbox_mono, minimum width=2.8cm] at (14.6, 3.8) {
\textbf{\textcolor{cstage}{TOPIC}}: \{topic\}.\\
TITLE: \{title\}.\\
\textbf{\textcolor{cstage}{TOPIC}}: \{topic\}
};
\node[font=\fontsize{6.5pt}{6.5pt}\selectfont, text=cstage] at (14.6, 2.9) {$\uparrow$ Semantic Anchors $\uparrow$};

\shade[left color=bgcol1, right color=cstage, rounded corners=2pt] (13.4, 2.1) rectangle (15.8, 2.4);
\node[mutedtext, anchor=east,text width=0.5cm] at (13.4, 2.25) {0};
\node[mutedtext, anchor=west,text width=0.5cm] at (15.8, 2.25) {1};
\draw[carrow, dashed, line width=0.8pt] (14.6, 1.9) -- (14.6, 2.6); 

\fill[cstage, rounded corners=2pt] (15.5, 2.25) circle (2.2pt); 
\fill[cstage, rounded corners=2pt] (15.0, 2.25) circle (2.2pt); 
\fill[cfail, rounded corners=2pt] (14.0, 2.25) circle (2.2pt);  

\node[mutedtext, text width=2.8cm] at (14.6, 1.5) {\textbf{Top-K} Selection ($K \le 2$)}; 

\draw[arrowmain] (16.2, 3.2) -- node[above=1pt, mutedtext] {~\textasciitilde8 $\to$ \\ $\le$ 2 videos} (17.3, 3.2);

\node[card, minimum width=1.8cm, minimum height=1.2cm] (out1) at (18.2, 4.1) {};
\fill[videoicon] ([xshift=4pt, yshift=-4pt]out1.north west) rectangle ([xshift=16pt, yshift=-12pt]out1.north west);
\fill[white] ([xshift=7.5pt, yshift=-5.5pt]out1.north west) -- ([xshift=13.5pt, yshift=-8pt]out1.north west) -- ([xshift=7.5pt, yshift=-10.5pt]out1.north west) -- cycle;
\node[font=\fontsize{6pt}{6pt}\selectfont, text=carrow, anchor=west] at ([xshift=18pt, yshift=-8pt]out1.north west) {Video 1};
\node[fill=cstage, text=white, inner sep=1.5pt, rounded corners=1pt, font=\fontsize{5pt}{5pt}\bfseries\selectfont, anchor=south east] at (out1.south east) {0.92};

\node[card, minimum width=1.8cm, minimum height=1.2cm] (out2) at (18.2, 2.5) {};
\fill[videoicon] ([xshift=4pt, yshift=-4pt]out2.north west) rectangle ([xshift=16pt, yshift=-12pt]out2.north west);
\fill[white] ([xshift=7.5pt, yshift=-5.5pt]out2.north west) -- ([xshift=13.5pt, yshift=-8pt]out2.north west) -- ([xshift=7.5pt, yshift=-10.5pt]out2.north west) -- cycle;
\node[font=\fontsize{6pt}{6pt}\selectfont, text=carrow, anchor=west] at ([xshift=18pt, yshift=-8pt]out2.north west) {Video 2};
\node[fill=cstage, text=white, inner sep=1.5pt, rounded corners=1pt, font=\fontsize{5pt}{5pt}\bfseries\selectfont, anchor=south east] at (out2.south east) {0.78};

\node[stagebadge] at (18.2, 5.3) {Final $\le$ 2}; 
\node[mutedtext, align=center, text width=1.8cm] at (18.2, 1.4) {\textit{To Fig. 3}};

\draw[carrow, dashed, opacity=0.2, line width=1pt, rounded corners=6pt] (0, 0.0) -- (0, 6.5) -- (19.4, 5.5) -- (19.4, 1.0) -- cycle;

\end{tikzpicture}
}
\caption{\textbf{Subtitle-driven Video-RAG pipeline.}
  From 50+ YouTube candidates, a metadata pre-filter removes outliers, then three subtitle-driven stages progressively narrow results:
  \textbf{(a)}~domain classification filters non-GUI content,
  \textbf{(b)}~topic extraction (title\,$+$\,subtitle $\to$ semantic descriptor), and
  \textbf{(c)}~dual-anchored relevance matching (topic as primary anchor, score 0--1).
  Final top-$K$ ($K{\le}2$) videos proceed to annotation (\cref{fig:annotation}).}
\label{fig:rag-pipeline}
\end{figure*}
\subsubsection{Initial Retrieval and Pre-filtering.}
Given $\mathcal{T}$ and the application name, an LLM generates a search-friendly query (\eg, ``How to adjust brightness in GIMP'').
To broaden recall, a simplified variant is also produced by stripping filler words (\eg, ``how to'', ``tutorial''), yielding up to two query variants.
These are executed via yt-dlp~\cite{ytdlp} YouTube search to collect $50{+}$ candidate URLs, which are deduplicated and subjected to lightweight metadata pre-filtering (duration ${<}\,3000$\,s, valid title characters).

\subsubsection{Stage~1: Subtitle-Assisted Domain Classification.}
For each surviving candidate, the system downloads its subtitle file and performs multi-step cleaning: removing VTT/SRT markers, timestamps, HTML tags, and duplicate lines, then re-segmenting at sentence boundaries.
The cleaned text (truncated to 10\,000 characters) is jointly input with the video title to an LLM, which classifies whether the video contains \emph{actual GUI operation demonstrations} rather than theoretical explanations, reviews, or entertainment.

The critical value: \emph{subtitles provide content evidence that titles alone cannot}.
A title such as ``Excel Tips'' is ambiguous; subtitles containing ``click on the Format menu'' or ``select the cell range'' unambiguously indicate a hands-on GUI tutorial.

\subsubsection{Stage~2: Subtitle-Driven Topic Extraction.}
For confirmed GUI tutorials, the system extracts a refined \emph{topic}---a 12--30 word text precisely describing the software, task, and key operations.
This extraction is based on joint title\,$+$\,subtitle analysis, with the LLM distilling the actual operational semantics from the narration rather than relying on the potentially misleading title.
For example, a video titled ``Excel Tutorial 2024'' whose subtitles actually demonstrate LibreOffice~Calc operations might yield the topic ``Creating and formatting a data table with conditional formatting in LibreOffice Calc''---correctly reflecting the true content rather than the misleading title.

\subsubsection{Stage~3: Topic-Title Joint Relevance Matching.}
For each candidate, the system constructs a \emph{dual-anchored prompt} that positions the subtitle-derived topic as the primary semantic anchor:
\begin{quote}
  \small\ttfamily
  TOPIC (higher priority): \{topic\}. TITLE: \{title\}. TOPIC: \{topic\}
\end{quote}
\noindent
The topic is repeated at both ends of the context window as a dual-anchored attention signal, ensuring the LLM prioritizes content-level semantics over the potentially noisy title.
All candidates are batch-scored for relevance (0.0--1.0) against the task description.

An \emph{adaptive top-$K$} strategy retains the top-ranked video unconditionally; subsequent videos require relevance $\geq 0.5$, up to $K{=}2$, ensuring at least one reference while preventing low-quality videos from degrading performance.

\FloatBarrier   
\vspace{-1mm}
\subsection{Fully Automated Annotation Pipeline}
\label{sec:annotation}

The second core innovation of \guidename{} is a fully automated pipeline that converts retrieved tutorial videos into structured, transferable knowledge (\cref{fig:annotation}).
The pipeline follows an \emph{inverse-dynamics-style} pairwise annotation paradigm: given two consecutive interface states (keyframes), a VLM infers the actions and intent that likely occurred between them, without training a separate dynamics model.

\begin{figure*}[!t]
\centering
\resizebox{0.85\textwidth}{!}{%
\begin{tikzpicture}[
    x=1cm, y=1cm,
    font=\sffamily,
    rowlabel/.style={font=\fontsize{9pt}{9pt}\bfseries\selectfont, anchor=north west, inner sep=0, text=ctext},
    rectmodule/.style={draw=carrow, fill=white, rounded corners=4pt, align=center, inner sep=4pt},
    rectvlm/.style={draw=carrow, line width=1pt, fill=white, rounded corners=6pt, align=center, inner sep=4pt},
    arrowmain/.style={-{Triangle[length=5pt,width=5pt]}, line width=1.5pt, carrow, shorten >=4pt, shorten <=4pt},
    arrowsub/.style={-{Triangle[length=4pt,width=4pt]}, line width=1.0pt, carrow, shorten >=4pt, shorten <=4pt},
    arrowscurve/.style={-{Triangle[length=5pt,width=5pt]}, line width=1.5pt, carrow, shorten >=4pt, shorten <=4pt},
    labeltext/.style={font=\fontsize{8pt}{8pt}\bfseries\selectfont, text=ctext},
    mutedtext/.style={font=\fontsize{7pt}{7pt}\selectfont, text=cmuted},
    mathtext/.style={font=\fontsize{7.5pt}{7.5pt}\itshape\selectfont, text=ctext},
    card/.style={draw={rgb,255:red,204;green,204;blue,204}, fill=white, rounded corners=2pt, inner sep=2pt},
    pilltag/.style={fill=cstage, text=white, font=\fontsize{7pt}{7pt}\bfseries\selectfont, rounded corners=4pt, inner sep=2.5pt},
    bubbletext/.style={draw=cstage, fill=white, rounded corners=4pt, inner sep=3pt, font=\fontsize{7pt}{7pt}\selectfont, text=cstage},
]

\definecolor{cplan}{RGB}{31,119,180}
\definecolor{cground}{RGB}{214,125,36}
\definecolor{cstage}{RGB}{44,160,44}
\definecolor{cfail}{RGB}{220,50,32}
\definecolor{carrow}{RGB}{68,68,68}
\definecolor{ctext}{RGB}{26,26,26}
\definecolor{cmuted}{RGB}{136,136,136}
\definecolor{cbgperception}{RGB}{245,249,255} 
\definecolor{cbginference}{RGB}{255,254,248} 
\definecolor{placeholderBg}{RGB}{240,240,240}
\definecolor{cbgplan}{RGB}{235,243,250}
\definecolor{cbgground}{RGB}{253,243,231}

\pgfdeclarelayer{background}
\pgfdeclarelayer{arrows} 
\pgfsetlayers{background,main,arrows}

\begin{pgfonlayer}{background}
    \fill[cbgperception, rounded corners=4pt] (0, 11.2) rectangle (17.4, 14.2);
\end{pgfonlayer}
\node[rowlabel] at (0.2, 13.9) {(a) Perception Frontend};

\node[card, minimum width=14mm, minimum height=10mm] (video) at (1.0, 12.6) {};
\fill[cfail, rounded corners=1.5pt] ([xshift=-4mm, yshift=-2.5mm]video.center) rectangle ([xshift=4mm, yshift=2.5mm]video.center);
\fill[white] ([xshift=-1.5mm, yshift=-1.5mm]video.center) -- ([xshift=2.5mm, yshift=0mm]video.center) -- ([xshift=-1.5mm, yshift=1.5mm]video.center) -- cycle;
\node[labeltext, anchor=north] at (video.south) {Tutorial Video};

\node[rectmodule, minimum width=2.4cm] (whisper) at (3.8, 12.6) {
    \textbf{\fontsize{8pt}{8pt}\selectfont Whisper ASR} \\[-0.5mm]
    \textcolor{cmuted}{\fontsize{6.5pt}{6.5pt}\selectfont (base model)}
};
\draw[cmuted, fill=white, rounded corners=0.5pt] (5.1, 12.3) -- (5.1, 12.9) -- (5.5, 12.9) -- (5.7, 12.7) -- (5.7, 12.3) -- cycle;
\draw[cmuted, rounded corners=0.5pt] (5.5, 12.9) -- (5.5, 12.7) -- (5.7, 12.7);
\node[font=\fontsize{4pt}{4pt}\bfseries\selectfont, text=cmuted] at (5.4, 12.6) {VTT};

\node[rectmodule, minimum width=2.8cm] (keyframe) at (7.8, 13.0) {
    \textbf{\fontsize{8pt}{8pt}\selectfont Keyframe Extraction} \\[-0.5mm]
    \textcolor{cmuted}{\fontsize{6.5pt}{6.5pt}\selectfont MOG2, fg > 10K px}
};

\node[fill=white, draw=carrow!30, rounded corners=1.5pt, minimum width=10mm, minimum height=7mm, inner sep=0, outer sep=0] (fb) at (7.0, 12.0) {};
\fill[carrow!15, rounded corners=1.5pt] ([yshift=-2mm]fb.north west) rectangle (fb.north east);
\fill[cfail] ([xshift=1.5mm, yshift=-1mm]fb.north west) circle (0.4pt);
\fill[cground] ([xshift=2.5mm, yshift=-1mm]fb.north west) circle (0.4pt);
\fill[cstage] ([xshift=3.5mm, yshift=-1mm]fb.north west) circle (0.4pt);
\draw[cstage, line width=1pt] (6.8, 11.8) -- (7.2, 11.8) (7.0, 11.6) -- (7.0, 12.0); 

\node[fill=white, draw=carrow!30, rounded corners=1.5pt, minimum width=10mm, minimum height=7mm, inner sep=0, outer sep=0] (fa) at (8.6, 12.0) {};
\fill[carrow!15, rounded corners=1.5pt] ([yshift=-2mm]fa.north west) rectangle (fa.north east);
\fill[cfail] ([xshift=1.5mm, yshift=-1mm]fa.north west) circle (0.4pt);
\fill[cground] ([xshift=2.5mm, yshift=-1mm]fa.north west) circle (0.4pt);
\fill[cstage] ([xshift=3.5mm, yshift=-1mm]fa.north west) circle (0.4pt);
\fill[cfail!20] (8.35, 11.6) rectangle (8.85, 11.9); 

\node[minimum width=2.2cm, minimum height=1cm] (thumbs1) at (11.0, 12.6) {};
\begin{scope}[shift={(10.2, 12.6)}]
    \foreach \i in {0,1,2,3} {
        \node[draw=carrow!30, fill=white, minimum width=4mm, minimum height=3mm, outer sep=0.5pt, rounded corners=0.5pt] (t\i) at (0.5*\i, 0) {};
        \fill[carrow!15, rounded corners=0.5pt] ([yshift=-1.2mm]t\i.north west) rectangle (t\i.north east);
        \fill[cfail] ([xshift=0.7mm, yshift=-0.6mm]t\i.north west) circle (0.3pt);
        \fill[cground] ([xshift=1.2mm, yshift=-0.6mm]t\i.north west) circle (0.3pt);
        \fill[cstage] ([xshift=1.7mm, yshift=-0.6mm]t\i.north west) circle (0.3pt);
    }
    \node[mutedtext, anchor=north] at (0.75, -0.2) {\{$s_0, s_1, \dots, s_n$\}};
\end{scope}

\node[rectmodule, minimum width=1.6cm] (omni) at (13.6, 12.6) {
    \textbf{\fontsize{8pt}{8pt}\selectfont Omni-}\\[-0.5mm]
    \textbf{\fontsize{8pt}{8pt}\selectfont Parser}
};

\node[minimum width=1.8cm, minimum height=1cm] (thumbs2) at (16.2, 12.6) {};
\begin{scope}[shift={(15.6, 12.6)}]
    \foreach \i in {0,1,2} {
        \node[draw=carrow!30, fill=white, minimum width=4mm, minimum height=3mm, outer sep=0pt, rounded corners=0.5pt] (ot\i) at (0.3*\i, -0.2*\i) {};
        \fill[carrow!15, rounded corners=0.5pt] ([yshift=-1.2mm]ot\i.north west) rectangle (ot\i.north east);
        \fill[cfail] ([xshift=0.7mm, yshift=-0.6mm]ot\i.north west) circle (0.3pt);
        \fill[cground] ([xshift=1.2mm, yshift=-0.6mm]ot\i.north west) circle (0.3pt);
        \fill[cstage] ([xshift=1.7mm, yshift=-0.6mm]ot\i.north west) circle (0.3pt);
        \draw[cground, line width=0.5pt] ([xshift=0.5pt, yshift=-1pt]ot\i.center) rectangle ([xshift=2.5pt, yshift=1pt]ot\i.center);
        \draw[cstage, line width=0.5pt] ([xshift=-2.5pt, yshift=0pt]ot\i.center) rectangle ([xshift=-0.5pt, yshift=-1.5pt]ot\i.center);
    }
    \draw[cground, fill=white, rounded corners=0.2pt] (0.8, 0.1) -- (0.8, 0.6) -- (1.1, 0.6) -- (1.2, 0.5) -- (1.2, 0.1) -- cycle;
    \draw[cground, rounded corners=0.2pt] (1.1, 0.6) -- (1.1, 0.5) -- (1.2, 0.5);
    \node[font=\fontsize{4pt}{4pt}\bfseries\selectfont, text=cground] at (1.0, 0.35) {\{.\}};
    \node[mutedtext, anchor=north] at ([yshift=-1mm]ot1.south) {\{$E_0, \dots, E_n$\}};
\end{scope}

\begin{pgfonlayer}{background}
    \fill[cbginference, rounded corners=4pt] (0, 5.5) rectangle (17.4, 10.3);
\end{pgfonlayer}
\node[rowlabel] at (0.2, 10.0) {(b) Pairwise Video Annotation};

\node[draw=carrow, dashed, rounded corners=4pt, line width=0.8pt, minimum width=6.0cm, minimum height=3.6cm, anchor=west] (boxb) at (0.4, 7.5) {};

\node[fill=white, draw=carrow!30, rounded corners=2pt, minimum width=16mm, minimum height=11mm, inner sep=0, outer sep=0] (st) at (1.8, 8.0) {};
\fill[carrow!15, rounded corners=2pt] ([yshift=-2.5mm]st.north west) rectangle (st.north east);
\fill[cfail] ([xshift=2.2mm, yshift=-1.25mm]st.north west) circle (1.2pt); 
\fill[cground] ([xshift=4.2mm, yshift=-1.25mm]st.north west) circle (1.2pt); 
\fill[cstage] ([xshift=6.2mm, yshift=-1.25mm]st.north west) circle (1.2pt); 
\node[mutedtext, anchor=south] at (st.north) {$s_t$ (Before)};
\draw[cground, line width=1pt] (1.3, 7.8) rectangle (1.9, 8.1); 

\node[fill=white, draw=carrow!30, rounded corners=2pt, minimum width=16mm, minimum height=11mm, inner sep=0, outer sep=0] (st1) at (4.0, 8.0) {};
\fill[carrow!15, rounded corners=2pt] ([yshift=-2.5mm]st1.north west) rectangle (st1.north east);
\fill[cfail] ([xshift=2.2mm, yshift=-1.25mm]st1.north west) circle (1.2pt); 
\fill[cground] ([xshift=4.2mm, yshift=-1.25mm]st1.north west) circle (1.2pt); 
\fill[cstage] ([xshift=6.2mm, yshift=-1.25mm]st1.north west) circle (1.2pt); 
\node[mutedtext, anchor=south] at (st1.north) {$s_{t+1}$ (After)};
\draw[cground, line width=1pt] (3.5, 7.8) rectangle (4.1, 8.1); 
\fill[cfail!20] (3.5, 7.4) rectangle (4.1, 7.6); 
\draw[cfail, line width=0.8pt] (3.5, 7.4) rectangle (4.1, 7.6);

\draw[cground, fill=white, rounded corners=0.5pt] (5.25, 7.65) -- (5.25, 8.45) -- (5.65, 8.45) -- (5.85, 8.25) -- (5.85, 7.65) -- cycle;
\draw[cground, rounded corners=0.5pt] (5.65, 8.45) -- (5.65, 8.25) -- (5.85, 8.25);
\draw[cground, fill=white, rounded corners=0.5pt] (5.1, 7.5) -- (5.1, 8.3) -- (5.5, 8.3) -- (5.7, 8.1) -- (5.7, 7.5) -- cycle;
\draw[cground, rounded corners=0.5pt] (5.4, 8.3) -- (5.4, 8.1) -- (5.6, 8.1);
\node[font=\fontsize{5pt}{5pt}\bfseries\selectfont, text=cground] at (5.4, 7.9) {\{...\}};
\node[mutedtext, anchor=south] at (5.4, 8.5) {$E_t, E_{t+1}$};

\node[pilltag] (ttopic) at (3.8, 6.9) {"GIMP brightness adjustment"};
\node[mutedtext, anchor=east] at (ttopic.west) {$T_{topic}$};
\node[bubbletext] (csub) at (3.8, 6.2) {\textit{"...then click Colors menu..."}};
\node[mutedtext, anchor=east] at (csub.west) {$C_{sub}$};

\node[rectvlm, minimum width=3.4cm, minimum height=1.6cm] (vlm) at (9.0, 8.3) {
    \textbf{\fontsize{10pt}{10pt}\itshape\selectfont $f_{VPA}$} \\[-0.5mm]
    \textcolor{cmuted}{\fontsize{7pt}{7pt}\selectfont (VLM: Qwen3-VL / Seed-1.8)}
};
\node[draw=carrow!30, fill=placeholderBg, rounded corners=2pt, inner sep=4pt, align=center] at (9.0, 6.5) {
    $a_t = f_{VPA}(s_t, E_t, s_{t+1}, E_{t+1},$\\[1mm]
    $T_{topic}, C_{sub})$
};

\node[draw=carrow!50, fill=white, rounded corners=2pt, minimum width=4.8cm, align=left, inner sep=5pt, font=\fontsize{7pt}{8.5pt}\selectfont] (json) at (14.0, 7.5) {
    \textbf{\fontsize{8.5pt}{10pt}\selectfont JSON Output}\\[2pt]
    \textbf{\{}\\
    \hspace*{1.5mm}\textcolor{ctext}{"Meaningful":} \hfill \textcolor{cstage}{$\checkmark$} \textcolor{cmuted}{/ $\times$}\\[1.5pt]
    \hspace*{1.5mm}\colorbox{cstage!15}{\textcolor{cfail}{$\star$} \textbf{\textcolor{cstage}{"Thought \& Action NLP":}}}\\[0.5pt]
    \hspace*{3mm}{\textcolor{cmuted}{\fontsize{6pt}{7pt}\selectfont  Env Observation \& Element Guesses}}\\[0.5pt]
    \hspace*{3mm}{\textcolor{cmuted}{\fontsize{6pt}{7pt}\selectfont  Current Step Action Planning}}\\[0.5pt]
    \hspace*{3mm}\colorbox{cstage!15}{\textbf{\textcolor{cstage}{"I will type 'EOMONTH'..."}}}\\[0.5pt]
    \textbf{\}}
};

\node[rowlabel] at (0.2, 4.8) {(c) Knowledge Decomposition};

\node[minimum width=2.4cm, minimum height=2.0cm, anchor=west] (boxc) at (0.4, 2.9) {};
\begin{scope}[shift={(1.4, 2.9)}]
    \foreach \i in {0,1,2} {
        \node[draw=carrow!30, fill=white, minimum width=16mm, minimum height=11mm, inner sep=0, outer sep=0, rounded corners=2pt] (jc\i) at (0.3*\i, -0.2*\i) {};
        \fill[carrow!15, rounded corners=2pt] ([yshift=-2.5mm]jc\i.north west) rectangle (jc\i.north east);
        \node[font=\fontsize{6pt}{6pt}\bfseries\selectfont, text=carrow] at (jc\i.center) {\{...\}};
        \fill[cfail] ([xshift=2.2mm, yshift=-1.25mm]jc\i.north west) circle (1.2pt);
        \fill[cground] ([xshift=4.2mm, yshift=-1.25mm]jc\i.north west) circle (1.2pt);
        \fill[cstage] ([xshift=6.2mm, yshift=-1.25mm]jc\i.north west) circle (1.2pt);
    }
    \node[labeltext, anchor=north, align=center, text width=3.0cm] at ([yshift=-1mm]0.4, -0.8) {Concatenated trajectory\\[1mm] \textit{\textcolor{cmuted}{\fontsize{7pt}{7pt}\selectfont "All meaningful Thought \& Action NLP descriptions"}}};
\end{scope}

\node[rectmodule, minimum width=2.8cm, minimum height=1.4cm] (llm) at (6.0, 2.9) {
    \textbf{\fontsize{8.5pt}{8.5pt}\selectfont LLM Decompose}
};

\node[draw=cplan, fill=cbgplan, rounded corners=4pt, line width=1pt, minimum width=7.4cm, minimum height=1.8cm, align=left, inner sep=0pt, anchor=west] (pbox) at (9.4, 4.0) {};
\node[fill=cplan, text=white, font=\fontsize{7.5pt}{7.5pt}\bfseries\selectfont, rounded corners=3pt, inner sep=4pt, anchor=west] (plabel) at ([xshift=6pt]pbox.west) {Planning};
\node[anchor=west, align=left, font=\fontsize{7.5pt}{9.5pt}\selectfont] at ([xshift=4pt]plabel.east) {
    $\bullet$ Execution Flow \& Planning\\
    $\bullet$ Key Considerations (Expert insights)\\
    $\bullet$ Coordinate-free Abstraction
};

\node[draw=cground, fill=cbgground, rounded corners=4pt, line width=1pt, minimum width=7.4cm, minimum height=1.8cm, align=left, inner sep=0pt, anchor=west] (gbox) at (9.4, 1.8) {};
\node[fill=cground, text=white, font=\fontsize{7.5pt}{7.5pt}\bfseries\selectfont, rounded corners=3pt, inner sep=4pt, anchor=west] (glabel) at ([xshift=6pt]gbox.west) {Grounding};
\node[anchor=west, align=left, font=\fontsize{7.5pt}{9.5pt}\selectfont] at ([xshift=4pt]glabel.east) {
    $\bullet$ $\le$15 Key UI Elements\\
    $\bullet$ Icon/Control $\cdot$ Appearance \& Position\\
    $\bullet$ Predicted Function
};

\begin{pgfonlayer}{arrows}
    
    \draw[arrowsub] (video.east) -- (whisper.west);
    \draw[arrowsub] (5.7, 12.6) -- (keyframe.west);
    \draw[arrowsub] (keyframe.east) -- (thumbs1.west);
    \draw[arrowsub] (thumbs1.east) -- (omni.west);
    \draw[arrowsub] (omni.east) -- (thumbs2.west);

    \draw[arrowsub] (st.east) -- node[above=0.5pt, font=\fontsize{6.5pt}{6.5pt}\bfseries\selectfont, text=carrow] {Action?} (st1.west);
    \draw[arrowmain] (boxb.east) -- (vlm.west); 
    \draw[arrowmain] (vlm.east) -- (json.west); 

    \draw[arrowsub] (boxc.east) -- (llm.west);
    \draw[arrowsub] (llm.east) -- (8.2, 2.9) |- (pbox.west);
    \draw[arrowsub] (llm.east) -- (8.2, 2.9) |- (gbox.west);

    \draw[arrowscurve, rounded corners=6pt] (16.2, 12.0) -- (16.2, 10.6) -- (-0.2, 10.6) -- (-0.2, 7.5) -- (boxb.west);
    \node[font=\fontsize{7.5pt}{8pt}\bfseries\selectfont, text=carrow, fill=white, inner sep=2pt, anchor=center] at (8.5, 10.6) {State pairs ($s_t, s_{t+1}$) + element graphs ($E_t, E_{t+1}$)};

    \draw[arrowscurve, rounded corners=6pt] (json.south) -- (14.0, 5.2) -- (-0.2, 5.2) -- (-0.2, 2.9) -- (boxc.west);
    \node[font=\fontsize{7.5pt}{8pt}\bfseries\selectfont, text=carrow, fill=white, inner sep=2pt, anchor=center] at (8.5, 5.2) {Per-frame annotations \{$a_t$\}};

\end{pgfonlayer}

\end{tikzpicture}
}
\caption{\textbf{Fully automated annotation pipeline.}
    Retrieved videos (\cref{fig:rag-pipeline}) are converted into structured knowledge through three phases.
    \textbf{(a)}~\emph{Perception Frontend}: Whisper ASR $\to$ keyframe extraction (MOG2) $\to$ OmniParser~\cite{omniparser} UI element graphs $E_t$.
    \textbf{(b)}~\emph{Pairwise Video Annotation}: $f_{\mathrm{VPA}}$ (Eq.\,\ref{eq:pairwise-annotation}) takes keyframe pairs ($s_t, s_{t+1}$), element graphs ($E_t, E_{t+1}$), topic $T_{\mathrm{topic}}$, and subtitle context $C_{\mathrm{sub}}$ to produce structured annotations.
    \textbf{(c)}~\emph{Knowledge Decomposition}: per-frame annotations are aggregated and decomposed into \textcolor{planblue}{planning} and \textcolor{groundorange}{grounding} knowledge.}
\label{fig:annotation}
\end{figure*}
\subsubsection{Keyframe Extraction: State Sampling.}
\label{sec:keyframe}
The pipeline first transcribes the video audio using Whisper~\cite{whisper} (base model, word-level timestamps), producing VTT subtitle files with sentence-level segment merging.
It then extracts keyframes using a subtitle-timestamp-aligned background subtraction method: the video is segmented according to subtitle intervals, and within each segment the MOG2 algorithm~\cite{mog2} detects frames with significant visual changes (foreground pixel count $>$ 10\,000).
Consecutive change sequences are analyzed to extract start and end frames of each transition, yielding a discrete sequence of interface states---the ``state pairs'' for pairwise video annotation.

\subsubsection{UI Element Detection: Enhancing Perception.}
\label{sec:omniparser}
Each keyframe is processed by OmniParser~\cite{omniparser}, which detects interactive UI elements (buttons, text fields, menus, \etc) and produces bounding-box annotations with structured JSON descriptions (position, type, text label).
This \emph{element-level prior knowledge} substantially improves VLM inference by providing structured UI understanding beyond raw visual perception.

\vspace{-2mm}
\subsubsection{VLM Pairwise Video Annotation.}
We formalize the annotation step as a pairwise video annotation problem.
Let $s_t$ and $s_{t+1}$ denote two consecutive keyframe images, $E_t$ and $E_{t+1}$ the corresponding OmniParser element graphs (bounding boxes, types, and text labels), $T_{\mathrm{topic}}$ the video topic extracted during retrieval, and $C_{\mathrm{sub}}$ the local subtitle context (preceding, current, and following sentences at the corresponding timestamp).
The annotation agent acts as a VLM pairwise annotation function:
\begin{equation}
  a_t \;=\; f_{\mathrm{VPA}}\!\bigl(s_t,\, E_t,\, s_{t+1},\, E_{t+1},\, T_{\mathrm{topic}},\, C_{\mathrm{sub}}\bigr),
  \label{eq:pairwise-annotation}
\end{equation}
where $f_{\mathrm{VPA}}$ denotes the VLM-based pairwise video annotator and $a_t$ is the structured annotation output.
Consecutive keyframe pairs are processed in a sliding-window fashion (frames 0\,--\,1, 1\,--\,2, 2\,--\,3, \etc).
The annotator visually compares the two frames, detects element-level changes via $E_t$/$E_{t+1}$ comparison, assesses task-meaningfulness using $T_{\mathrm{topic}}$ and $C_{\mathrm{sub}}$, and infers actions with strategic intent.
\vspace{-2mm}
\subsubsection{Transferable Annotation Format.}
\label{sec:annotation-format}
Coordinate-level video labels are not transferable across environments.
Instead, $a_t$ contains a \texttt{Meaningful} flag and a natural-language \texttt{ThoughtAction} field:
\vspace{-2mm}
\paragraph{Semantic Noise Filtering (\texttt{Meaningful}).}
A boolean indicating whether the frame-pair change is task-relevant.
This field leverages the joint semantic judgment of VLM~$+$~OmniParser to \emph{complement traditional keyframe extraction}, which detects only pixel-level changes (mouse movement, window flicker) but cannot distinguish meaningful operations from visual noise.
Human verification confirms that this mechanism correctly filters out $>$91\% of non-GUI frames and idle no-action frames; the complete evaluation is provided in the supplementary material.
\vspace{-2mm}
\paragraph{Strategic Narrative (\texttt{Thought\,\&\,Action\,NLP}).}
The core annotation field, written from the first-person executor perspective, integrating step-by-step execution reasoning---not just ``what was done'' but ``why,'' the decision logic, and contextual judgment.
It includes \emph{UI element visual descriptions} (color, shape, position, text labels) and function inferences (\ie, guesses about each element's purpose based on appearance and context), as well as \emph{grounded actions} (click, type, scroll, \etc) in natural language form.
This strategic knowledge is inherently transferable across interface variations.
\vspace{-2mm}
\subsubsection{Knowledge Decomposition.}
\label{sec:decomposition}
Per-frame annotations are consolidated into per-video trajectories by concatenating the meaningful natural-language action descriptions.
An LLM then decomposes each trajectory into two knowledge types, precisely targeting the two manifestations of domain bias:

\smallskip\noindent
\textbf{Planning knowledge}---provides domain-specific operational workflow:
\begin{itemize}[leftmargin=1.5em,nosep]
\item \emph{Execution Flow \& Planning}: a coherent narrative of the task workflow---logical step sequences, stage objectives, and progression.
\item \emph{Key Considerations}: high-value insights (1--2 sentences each) distilled from the tutorial creator's expertise, helping avoid common pitfalls.
\item \emph{Coordinate-free Abstraction}: deliberately excludes low-level UI coordinates, decoupling workflow logic from specific display resolutions or layout versions.
\end{itemize}

\smallskip\noindent
\textbf{Grounding knowledge}---provides domain-specific UI element descriptions:
\begin{itemize}[leftmargin=1.5em,nosep]
\item Up to 15 key interactive elements, each with: \emph{Icon/Control} (name), \emph{Appearance \& Position} (visual attributes, screen-relative location), and \emph{Predicted Function} (inferred functionality and typical usage).
\item Use visual appearance and semantic function rather than absolute coordinates, enabling element identification across interface states.
\end{itemize}

\subsection{Plug-and-Play Agent Integration}
\label{sec:integration}

A key design principle of \guidename{} is \emph{agent-agnosticism}: the extracted \plan{} and \ground{} knowledge are expressed entirely in natural language and are independent of any specific GUI agent architecture, enabling plug-and-play integration with both single-model agents and multi-agent systems without modifying model parameters or architecture.

\paragraph{General Integration Principle.}
\plan{} knowledge is injected into the agent's \emph{planning context} (task decomposition and action decisions); \ground{} knowledge is injected into the \emph{grounding context} (instruction-to-coordinate mapping).
In all cases the knowledge is framed as \emph{reference material}, not directives: the agent verifies suggestions against the current screenshot and prioritizes its own observations when conflicts arise (\eg, due to software version shifts or UI layout differences between the tutorial and the target environment), providing \emph{robustness against domain shift}.
When $K{>}1$ videos are retrieved, each video's knowledge is individually labeled and concatenated, providing multi-perspective references.

\paragraph{Mode~A: Multi-Agent System (e.g., AgentS3~\cite{agents3}).}
We integrate with AgentS3, consisting of a \emph{Worker}, a \emph{Grounding Agent}, and a \emph{Code Agent}.
The two knowledge streams are routed to architecturally matched modules:
\plan{} knowledge is appended to the Worker's system prompt as a ``Video Planning Reference'' section alongside the screenshot and interaction history, guiding task decomposition;
\ground{} knowledge is supplied to the Grounding Agent as supplementary context for each action query, aiding coordinate prediction.
This separation mirrors the system's native division of labor; full prompts are provided in the supplementary material.

\paragraph{Mode~B: Single-Model Agent (e.g., Qwen3-VL~\cite{qwen3vl}).}
Both \plan{} and \ground{} knowledge are unified into a single system prompt under an ``External Knowledge'' section.
A \emph{structured Thought template}, acting as a \emph{guided Chain-of-Thought}~\cite{cot}, enforces active knowledge referencing at each step:
the model first assesses task similarity against the retrieved demo, then locates its current stage within the workflow, derives the next sub-task from \plan{} knowledge, and matches \ground{} element descriptions against the current screenshot before selecting an action.
When only one channel is available, the template degrades gracefully to a partial CoT.

\section{Experiments}
\label{sec:experiments}

We evaluate \guidename{} on OSWorld, studying its impact across agent architectures, knowledge channel configurations, and application domains.

\subsection{Experimental Setup}

\paragraph{Benchmark.}
We evaluate on OSWorld~\cite{osworld}, a benchmark of 369 real-world computer tasks executed inside live Ubuntu virtual machines.
Following common practice, we exclude 8 Google-Drive-dependent tasks and report the \emph{average score} (\%) on the remaining 361 tasks spanning 10 application domains; full benchmark details are provided in the supplementary material.

\paragraph{Evaluation Protocol.}
We evaluate the effectiveness of knowledge injection by comparing \emph{the same agent under different configurations}:
\textbf{No Annotation} (baseline), \textbf{+\,\plan{}} (planning only), and \textbf{+\,\plan{}\,\&\,\ground{}} (full dual-channel, default $k{=}7$ grounding elements).
Three agents are evaluated:
(i)~Seed-1.8~\cite{seed18} (Mode~B, single-model);
(ii)~Qwen3-VL-8B~\cite{qwen3vl} (Mode~B, single-model);
and (iii)~AgentS3~\cite{agents3}, a multi-agent system with GPT-5.2~\cite{gpt5} Worker + Seed-1.8 Grounding (Mode~A).

\paragraph{Implementation Details.}
The annotation pipeline uses GPT-5.1~\cite{gpt5} for both frame-pair VLM pairwise annotation (Eq.\,\ref{eq:pairwise-annotation}) and planning-grounding decomposition, with OmniParser~\cite{omniparser} for UI element detection.
The retrieval pipeline uses GPT-4.1~\cite{gpt4o} for query generation and GPT-4.1-mini for classification and scoring.
Full configuration details are provided in the supplementary material.

\paragraph{Practical Notes.}
Two factors affect absolute scores:
(i)~the Seed-1.8 API endpoint employs anti-distillation measures that intermittently produce malformed outputs; we apply lightweight parsing fixes, though the \emph{relative} improvements from \guidename{} are unaffected;
(ii)~intermittent network issues prevent a subset of Chrome tasks from loading target webpages, slightly depressing scores in this domain across all configurations.

\newcommand{\hc}{\cellcolor{placeholderBg}}
\begin{table*}[!htbp]
  \caption{\textbf{\guidenameC{} main results on OSWorld} (average score, \%).
    Configurations: No Annotation (baseline), +\,\textcolor{planblue}{Plan.}\ (planning only), +\,\textcolor{planblue}{Plan.}\,\&\,\textcolor{groundorange}{Gnd.}\ (full dual-channel).
    $^\dagger$Baseline from~\cite{qwen3vl}.
    $^\ddagger$GPT-5.2 Worker + Seed-1.8 Grounding (Mode~A); $k{=}5$.
    \textbf{Bold}: best per column per agent; {\scriptsize\textcolor{stagegreen}{green}}: absolute gain over baseline.}
  \label{tab:main-results}
  \centering\small
  \setlength{\tabcolsep}{2.2pt}
  \renewcommand{\arraystretch}{1.05}
  \resizebox{\textwidth}{!}{%
  \begin{tabular}{@{}llc cccccccccc@{}}
    \toprule
    Agent & Configuration & \textbf{Overall} & \textbf{Chrome} & \textbf{GIMP} & \textbf{Calc} & \textbf{Impress} & \textbf{Writer} & \textbf{OS} & \textbf{ThBrd} & \textbf{VLC} & \textbf{VSCode} & \textbf{Multi} \\
    & & \textbf{(361)} & (46) & (26) & (47) & (45) & (23) & (24) & (15) & (17) & (23) & (93) \\
    \midrule
    \multirow{3}{*}{Seed-1.8}
      & No Annotation      & 37.14 & 36.87 & 26.92 & 29.79 & 43.09 & 34.77 & 45.83 & 66.67 & 47.06 & 60.87 & 26.88 \\
      & +\,\textcolor{planblue}{Planning}           & 43.93\,{\scriptsize\textcolor{stagegreen}{+6.79}} & \textbf{47.74} & 34.62 & 42.55 & \textbf{46.51} & 56.38 & \textbf{50.00} & \textbf{73.33} & \textbf{52.32} & \textbf{65.22} & \textbf{27.89} \\
      & \hc +\,\textcolor{planblue}{Plan.}\,\&\,\textcolor{groundorange}{Gnd.} & \hc \textbf{44.62}\,{\scriptsize\textcolor{stagegreen}{+7.48}} & \hc \textbf{47.74} & \hc \textbf{42.31} & \hc \textbf{48.94} & \hc 45.31 & \hc \textbf{56.51} & \hc \textbf{50.00} & \hc \textbf{73.33} & \hc \textbf{52.32} & \hc \textbf{65.22} & \hc 25.74 \\
    \midrule
    \multirow{3}{*}{\shortstack[l]{Qwen3-VL\\-8B}}
      & No Annotation$^\dagger$ & 33.90 & -- & -- & -- & -- & -- & -- & -- & -- & -- & -- \\
      & +\,\textcolor{planblue}{Planning}           & 38.93\,{\scriptsize\textcolor{stagegreen}{+5.03}} & \textbf{41.22} & \textbf{48.00} & 27.66 & \textbf{47.59} & \textbf{52.16} & 50.00 & 73.33 & 40.56 & \textbf{52.17} & 21.71 \\
      & \hc +\,\textcolor{planblue}{Plan.}\,\&\,\textcolor{groundorange}{Gnd.} & \hc \textbf{39.73}\,{\scriptsize\textcolor{stagegreen}{+5.83}} & \hc 36.87 & \hc 46.15 & \hc \textbf{34.78} & \hc 38.48 & \hc \textbf{52.16} & \hc \textbf{54.17} & \hc \textbf{80.00} & \hc \textbf{51.77} & \hc 43.48 & \hc \textbf{25.98} \\
    \midrule
    \multirow{2}{*}{AgentS3$^\ddagger$}
      & No Annotation      & 50.18 & 41.18 & 38.46 & 51.06 & 44.62 & 52.17 & \textbf{70.83} & 73.33 & \textbf{73.91} & 73.91 & \textbf{40.32} \\
      & \hc +\,\textcolor{planblue}{Plan.}\,\&\,\textcolor{groundorange}{Gnd.} & \hc \textbf{54.65}\,{\scriptsize\textcolor{stagegreen}{+4.47}} & \hc \textbf{49.85} & \hc \textbf{53.85} & \hc \textbf{65.96} & \hc \textbf{46.88} & \hc \textbf{65.22} & \hc \textbf{70.83} & \hc \textbf{80.00} & \hc 56.25 & \hc \textbf{82.61} & \hc 37.10 \\
    \bottomrule
  \end{tabular}}
\end{table*}

\subsection{Main Results}

\cref{tab:main-results} presents the main results on the full OSWorld benchmark.
As a plug-and-play component, \guidename{} yields consistent and significant improvements across all three agents without modifying any model parameters.

\paragraph{Dual-Channel Analysis.}
\plan{} knowledge alone produces the dominant share of improvement: $+$6.79\% on Seed-1.8 and $+$5.03\% on Qwen3-VL-8B, accounting for ${\sim}$86--91\% of the total gain, confirming that the primary bottleneck is \emph{planning-level domain bias}.
Adding \ground{} yields further complementary gains ($+$0.69--0.80\%), strongest in domains with complex UI layouts (GIMP $+$7.69pp, Calc $+$6.39pp on Seed-1.8).

\paragraph{Architecture-Agnostic Generality.}
\guidename{} also generalizes to multi-agent systems: AgentS3~\cite{agents3} (GPT-5.2 Worker + Seed-1.8 Grounding) improves from 50.18\% to 54.65\% ($+$4.47pp), with the largest gains in GIMP ($+$15.39pp), Calc ($+$14.90pp), and Writer ($+$13.05pp).
On WindowsAgentArena (WAA)~\cite{windowsagentarena}, the same coordinate-free knowledge format transfers to native Windows widgets without tuning, improving Agents3+GPT-5.2 by $+$10.21pp and Qwen3-VL-32B-Instruct by $+$12.46pp (\cref{tab:waa-results}).
This cross-benchmark evidence confirms that GUIDE's gain is not an OSWorld artifact.

\paragraph{Per-Domain Analysis and Execution Efficiency.}
Domains with severe domain bias benefit most (Writer $+$21.74pp, Calc $+$19.15pp on Seed-1.8).
\ground{} knowledge reduces exploration on successful tasks (VLC $-$5.0, Calc $-$3.7 steps) by enabling direct element identification.
Knowledge injection increases per-step latency (9.5$\to$11.6\,s, $+$2.1\,s), but grounding compensates through fewer steps (15.7 vs.\ 16.6 on successful tasks); overall, success count grows from 134 to 161 ($+$20.1\%) at modest wall-clock overhead.
\vspace{-1mm}
\subsection{Ablation Studies and Additional Controls}
\label{sec:ablation}

\begin{table*}[tb]
  \caption{\textbf{Cross-benchmark transfer on WindowsAgentArena~\cite{windowsagentarena}} (154 tasks, average score, \%).
    GUIDE uses the same planning and grounding knowledge format as in OSWorld, without WAA-specific tuning.
    Task counts: Office 45, Web 30, System 24, Code 24, Media 21, Utility 10.
    \textbf{Bold}: best per backbone group.}
  \label{tab:waa-results}
  \centering\small
  \setlength{\tabcolsep}{2.5pt}
  \renewcommand{\arraystretch}{1.06}
  \resizebox{\textwidth}{!}{%
  \begin{tabular}{@{}llcccccccc@{}}
    \toprule
    Backbone & Configuration & Office & Web & System & Code & Media & Utility & \textbf{Overall} & $\Delta$ \\
    \midrule
    \multirow{2}{*}{Agents3+GPT-5.2}
      & No Annotation & 47.80 & 38.60 & 69.00 & 62.10 & 32.30 & 41.40 & 49.00 & +0.00 \\
      & \hc +\,\textcolor{planblue}{Plan.}\,\&\,\textcolor{groundorange}{Gnd.}
        & \hc \textbf{57.78} & \hc \textbf{46.67} & \hc \textbf{83.33} & \hc \textbf{75.00} & \hc \textbf{38.99} & \hc \textbf{50.00} & \hc \textbf{59.21} & \hc \textcolor{stagegreen}{+10.21} \\
    \midrule
    \multirow{2}{*}{Qwen3-VL-32B}
      & No Annotation & 27.10 & 35.90 & 53.80 & 20.90 & 30.80 & 14.40 & 31.70 & +0.00 \\
      & \hc +\,\textcolor{planblue}{Plan.}\,\&\,\textcolor{groundorange}{Gnd.}
        & \hc \textbf{37.78} & \hc \textbf{50.00} & \hc \textbf{75.00} & \hc \textbf{29.17} & \hc \textbf{42.89} & \hc \textbf{20.00} & \hc \textbf{44.16} & \hc \textcolor{stagegreen}{+12.46} \\
    \bottomrule
  \end{tabular}}%
\end{table*}

\paragraph{Cross-Benchmark Transfer.}
\cref{tab:waa-results} evaluates GUIDE on a different benchmark and OS stack.
OSWorld runs in Linux desktop applications, whereas WAA uses native Windows widgets and a different action interface.
Despite this shift, GUIDE improves every WAA domain under both a strong multi-agent backbone and an open 32B VLM backbone.
The open backbone receives the larger absolute gain ($+$12.46pp, 39\% relative), indicating that video-derived procedural knowledge remains useful precisely where built-in domain expertise is weaker.

\begin{table*}[tb]
  \caption{\textbf{Same-backbone OSWorld controls on Qwen3-VL-8B-thinking} (361 tasks, average score, \%).
    The Watch\,\&\,Learn~\cite{watchlearn}-style row uses the same retrieved videos as GUIDE, but injects raw text trajectories rather than decomposed planning/grounding knowledge.
    Domain scores for the published no-annotation baseline are not available in~\cite{qwen3vl}.}
  \label{tab:osworld-controls}
  \centering\small
  \setlength{\tabcolsep}{1.8pt}
  \renewcommand{\arraystretch}{1.06}
  \resizebox{\textwidth}{!}{%
  \begin{tabular}{@{}lcccccccccccc@{}}
    \toprule
    Configuration & \textbf{Overall} & Chrome & GIMP & Calc & Impress & Writer & OS & ThBrd & VLC & VSCode & Multi & $\Delta$ \\
    \midrule
    No Annotation$^\dagger$ & 33.90 & -- & -- & -- & -- & -- & -- & -- & -- & -- & -- & +0.00 \\
    W\&L raw traj. ICL & 31.96 & 45.56 & 38.46 & 17.02 & 29.58 & 43.47 & 45.83 & 66.67 & 40.56 & 30.43 & 18.94 & \textcolor{red}{-1.94} \\
    Grounding-only ($k{=}7$) & 34.31 & 41.22 & 46.15 & 23.40 & 38.17 & 47.82 & 45.83 & 66.67 & 28.79 & 39.13 & 19.44 & \textcolor{stagegreen}{+0.41} \\
    +\,\textcolor{planblue}{Planning} & 38.93 & \textbf{41.22} & \textbf{48.00} & 27.66 & \textbf{47.59} & \textbf{52.16} & 50.00 & 73.33 & 40.56 & \textbf{52.17} & 21.71 & \textcolor{stagegreen}{+5.03} \\
    \hc +\,\textcolor{planblue}{Plan.}\,\&\,\textcolor{groundorange}{Gnd.}
      & \hc \textbf{39.73} & \hc 36.87 & \hc 46.15 & \hc \textbf{34.78} & \hc 38.48 & \hc \textbf{52.16} & \hc \textbf{54.17} & \hc \textbf{80.00} & \hc \textbf{51.77} & \hc 43.48 & \hc \textbf{25.98} & \hc \textcolor{stagegreen}{+5.83} \\
    \bottomrule
  \end{tabular}}%
\end{table*}

\paragraph{Attribution and Raw-Context Controls.}
\cref{tab:osworld-controls} isolates the knowledge-injection design on the same backbone and split.
Planning-only accounts for $5.03/5.83=86.3\%$ of the full Qwen3-VL-8B gain, while grounding alone gives only $+$0.41pp; grounding is therefore complementary, adding $+$0.80pp on top of an already correct workflow plan.
The Watch\,\&\,Learn-style ICL baseline uses the same 299 retrieved-video tasks, the same GPT-5.1 trajectory source before decomposition, $K{=}2$ demos, 12 steps per demo, and a 24K-character cap.
It scores 31.96, below both no annotation and GUIDE, showing that raw tutorial-derived trajectories do not explain the gain; the useful signal is the structured conversion into \plan{} and \ground{} knowledge.

\begin{table*}[tb]
  \caption{\textbf{Annotation-pipeline ablation on a balanced OSWorld subset} (120 tasks, 12 per domain, Qwen3-VL-8B-thinking, average over three random seeds; score \%).
    Subtitle-only removes the visual annotation chain and injects transcript-derived planning knowledge.
    No-OmniParser keeps the visual pairwise annotation chain but removes the UI element graph.}
  \label{tab:pipeline-ablation}
  \centering\scriptsize
  \setlength{\tabcolsep}{3pt}
  \renewcommand{\arraystretch}{0.88}
  \begin{tabular}{@{}lr@{\hspace{8pt}}rrrr@{}}
    \toprule
    \multirow{2}{*}{Domain} & \multirow{2}{*}{$n$} & \multicolumn{4}{c}{Score (\%)} \\
    \cmidrule(l){3-6}
    & & None & Sub. & w/o OP & \textbf{Full} \\
    \midrule
    Chrome & 12 & 58.33 & 58.33 & 58.33 & \textbf{58.33} \\
    GIMP & 12 & 33.33 & 50.00 & \textbf{58.33} & 50.00 \\
    Calc & 12 & 8.33 & 16.67 & \textbf{25.00} & 16.67 \\
    Impress & 12 & 33.33 & 41.91 & \textbf{58.57} & \textbf{58.57} \\
    Writer & 12 & 66.65 & 58.31 & 58.31 & \textbf{66.65} \\
    Multi-apps & 12 & \textbf{41.67} & 33.33 & \textbf{41.67} & \textbf{41.67} \\
    OS & 12 & 25.00 & \textbf{41.67} & \textbf{41.67} & \textbf{41.67} \\
    Thunderbird & 12 & \textbf{75.00} & 58.33 & 66.67 & 66.67 \\
    VLC & 12 & 16.67 & 39.65 & 33.33 & \textbf{40.89} \\
    VSCode & 12 & 41.67 & 25.00 & 33.33 & \textbf{50.00} \\
    \midrule
    \textbf{Average} & \textbf{120} & 40.00 & 42.32 & 47.52 & \textbf{49.11} \\
    $\Delta$ vs. No Annotation & -- & +0.00 & \textcolor{stagegreen}{+2.32} & \textcolor{stagegreen}{+7.52} & \textcolor{stagegreen}{+9.11} \\
    \bottomrule
  \end{tabular}%
\end{table*}

\paragraph{Annotation-Pipeline Components.}
\cref{tab:pipeline-ablation} separates the sources of improvement under repeated seeds.
Subtitle-only already improves the 120-task subset by $+$2.32pp, confirming that retrieved ASR transcripts contain useful procedural cues.
Restoring visual pairwise annotation without OmniParser raises the gain to $+$7.52pp, showing that adjacent-frame visual reasoning contributes substantially beyond transcripts.
Full GUIDE reaches $+$9.11pp; the additional $+$1.59pp from OmniParser is positive but modest under a strong GPT-5.2 annotator, so we treat it as complementary rather than decisive in this setting.
The design motivation remains robustness: when the annotator is weaker, the explicit UI graph provides the structural prior needed to produce reliable coordinate-free element descriptions.

\begin{table*}[tb]
  \caption{\textbf{Number of retrieved video demonstrations} on a 50-task OSWorld subset with Qwen3-VL-8B-thinking and full GUIDE knowledge (score \%).
    All settings complete 50/50 tasks; only the number of retrieved demonstrations changes.}
  \label{tab:k-video-ablation}
  \centering\scriptsize
  \setlength{\tabcolsep}{6pt}
  \renewcommand{\arraystretch}{0.9}
  \begin{tabular}{@{}lrrrr@{}}
    \toprule
    Domain & \#Tasks & $K{=}1$ & $K{=}2$ & $K{=}3$ \\
    \midrule
    Chrome & 7 & 71.43 & \textbf{85.71} & 57.14 \\
    GIMP & 8 & \textbf{50.00} & \textbf{50.00} & \textbf{50.00} \\
    Calc & 6 & \textbf{16.67} & \textbf{16.67} & \textbf{16.67} \\
    Impress & 6 & 33.33 & 50.34 & \textbf{50.84} \\
    Writer & 5 & 20.00 & \textbf{40.00} & 20.00 \\
    Multi-apps & 6 & 50.00 & 33.33 & \textbf{66.67} \\
    OS & 1 & \textbf{100.00} & \textbf{100.00} & \textbf{100.00} \\
    Thunderbird & 3 & 33.33 & \textbf{66.67} & \textbf{66.67} \\
    VLC & 5 & \textbf{58.13} & 20.00 & 40.00 \\
    VSCode & 3 & \textbf{0.00} & \textbf{0.00} & \textbf{0.00} \\
    \midrule
    \textbf{Overall} & \textbf{50} & 41.81 & 44.04 & \textbf{44.10} \\
    \bottomrule
  \end{tabular}%
\end{table*}

\paragraph{Retrieved-Video Count.}
\cref{tab:k-video-ablation} supports the default $K{=}2$ design.
Adding a second relevant video improves the 50-task score by $+$2.23pp over $K{=}1$, while a third video adds only $+$0.06pp.
The per-domain pattern is non-monotonic because additional videos can introduce alternative workflows or UI variants that are not helpful for a particular task; overall, $K{=}2$ provides the best accuracy/context tradeoff.

\begin{table}[tb]
  \caption{\textbf{Retrieval matching controls} under the same 300-video, 3-annotator protocol as the paper's retrieval audit.
    All modes choose from the same per-task \texttt{yt-dlp} candidate pool; scores use the $1.0/0.5/0.0$ direct/partial/failed scale.}
  \label{tab:retrieval-controls}
  \centering\scriptsize
  \setlength{\tabcolsep}{4pt}
  \renewcommand{\arraystretch}{0.9}
  \begin{tabular}{@{}lcccc@{}}
    \toprule
    Retrieval mode & \#Videos & Mean & Acc.$\geq0.5$ & 1.0 / 0.5 / 0.0 \\
    \midrule
    Full GUIDE & 300 & \textbf{0.867} & \textbf{96.00\%} & \textbf{77.33} / 18.67 / 4.00 \\
    Title-only & 300 & 0.782 & 88.67\% & 67.67 / 21.00 / 11.33 \\
    Random & 300 & 0.628 & 80.33\% & 45.33 / 35.00 / 19.67 \\
    \bottomrule
  \end{tabular}
\end{table}

\paragraph{Retrieval Controls.}
\cref{tab:retrieval-controls} evaluates whether the retrieval pipeline adds value beyond the candidate pool itself.
The random baseline is already reasonably high because the initial task-derived \texttt{yt-dlp} query filters for basic relevance.
Title-only ranking adds a substantial improvement over random, but Full GUIDE further improves both mean match and acceptable rate by using subtitle-derived topics.
Most importantly, the strictly direct-match share rises from 45.33\% (random) and 67.67\% (title-only) to 77.33\%, isolating the subtitle/topic signal as the source of retrieval quality rather than merely the candidate pool.

\vspace{-2mm}
\subsection{Qualitative Analysis}
\label{sec:qualitative}

\begin{figure*}[tb]
  \centering
  \resizebox{0.78\textwidth}{!}{%
  \begin{tikzpicture}[
      x=1mm, y=1mm,
      font=\sffamily,
      badge/.style={fill=carrow, text=white, font=\bfseries\fontsize{7pt}{7pt}\selectfont, rounded corners=3pt, inner sep=3pt, align=center},
      card/.style={draw=none, fill=white, rounded corners=2pt, inner sep=3pt},
      planbox/.style={draw=cplan, fill=cbgplan, rounded corners=4pt, line width=1pt},
      groundbox/.style={draw=cground, fill=cbgground, rounded corners=4pt, line width=1pt},
      thoughtbox/.style={fill=placeholderBg, rounded corners=3pt, inner sep=3pt, font=\ttfamily\fontsize{6.5pt}{7pt}\selectfont, align=left},
      screenshot/.style={draw=carrow!30, line width=0.5pt, inner sep=0pt, rounded corners=2pt},
      labeltext/.style={font=\bfseries\fontsize{7pt}{7pt}\selectfont, text=ctext},
      subtext/.style={font=\fontsize{6pt}{6.5pt}\selectfont, text=ctext},
      arrowconnect/.style={-{Triangle[length=4pt,width=4pt]}, carrow, dashed, line width=0.8pt},
      highlightplan/.style={fill=cplan!15, text=black},
      highlightground/.style={fill=cground!15, text=black}
  ]
  
  \definecolor{cplan}{RGB}{31,119,180}
  \definecolor{cground}{RGB}{214,125,36}
  \definecolor{cstage}{RGB}{44,160,44}
  \definecolor{cfail}{RGB}{200,50,50}
  \definecolor{carrow}{RGB}{68,68,68}
  \definecolor{ctext}{RGB}{26,26,26}
  \definecolor{cmuted}{RGB}{120,120,120}
  \definecolor{cbgplan}{RGB}{235,243,250}
  \definecolor{cbgground}{RGB}{253,243,231}
  \definecolor{placeholderBg}{RGB}{245,245,245}
  
  \node[fill=placeholderBg, rounded corners=4pt, minimum width=174mm, inner sep=2.5mm, anchor=north west] (header) at (0, 137) {
      \begin{tabular}{p{168mm}}
      \fontsize{7.5pt}{9pt}\selectfont \textbf{Task Instruction:} ``I'd like to make the picture's contrast stronger to really bring out the main subject.''\\
      \fontsize{7.5pt}{9pt}\selectfont \textbf{Application:} GIMP 2.10 \hspace{3mm} \textbf{Retrieved Video:} ``How to Add Contrast to Photos in Gimp''
      \end{tabular}
  };
  
  \node[anchor=north west, font=\fontsize{9pt}{9pt}\bfseries\selectfont] at (0, 123) {(a) Step 1 / 6: Planning Guides Workflow};
  
  \fill[cbgplan, rounded corners=4pt] (0, 65) rectangle (174, 118);
  \draw[cplan, line width=1pt, rounded corners=4pt] (0, 65) rectangle (174, 118);
  
  \node[fill=white, draw=cplan!30, rounded corners=3pt, minimum width=43mm, minimum height=49mm, anchor=north west] (pk) at (1.5, 116.5) {};
  \fill[cplan, rounded corners=2pt] (1.5, 116.5) rectangle (2.3, 67.5); 
  \node[fill=cplan, text=white, rounded corners=2pt, font=\bfseries\fontsize{6.5pt}{6.5pt}\selectfont, anchor=north west] at (3, 115) {Planning Knowledge};
  
  \node[anchor=north west, text width=38mm, align=left, font=\fontsize{7pt}{8pt}\selectfont] at (3, 110) {
      \textbf{Execution Flow:}\\
      1. Duplicate original layer\\[0.5mm]
      \colorbox{cplan!15}{\parbox{35mm}{\textbf{2. Open ``Brightness-Contrast''}\\ \textbf{via Colors menu} \textcolor{cstage}{$\blacktriangleleft$}}} \\[0.5mm]
      3. Top menu bar navigation\\
      4. Adjust contrast slider\\
      5. Confirm OK\\[1.5mm]
      \textbf{Key Note:}\\
      ``In GIMP, contrast is under \textbf{Colors}, NOT Image menu.''
  };

  \node[anchor=north west, inner sep=0] (screen1) at (49, 116) {
      \includegraphics[width=78mm]{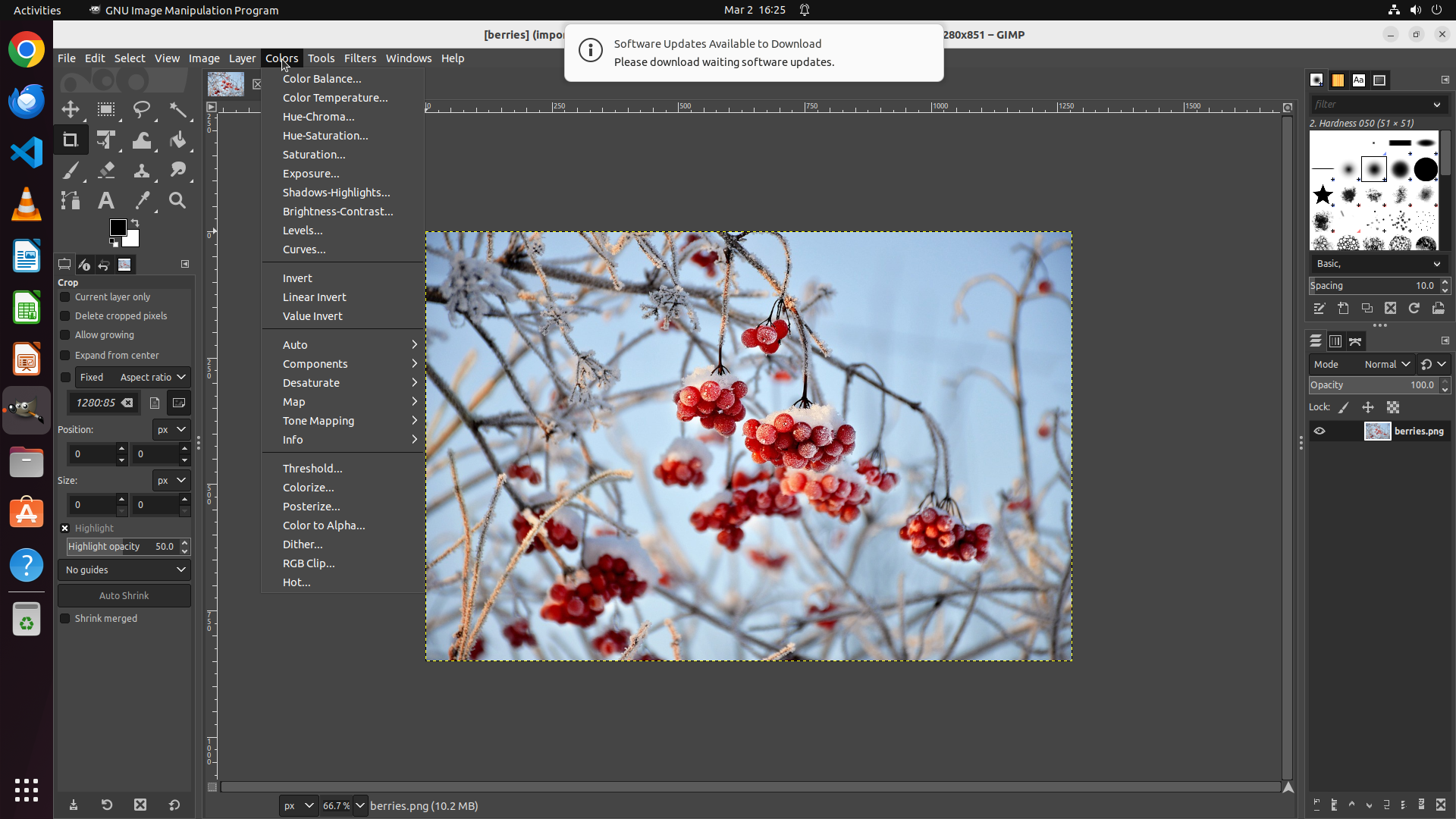}
  };
  \draw[carrow!30, line width=0.5pt, rounded corners=2pt] (screen1.south west) rectangle (screen1.north east);
  
  \begin{scope}[shift={(screen1.south west)}, x={(screen1.south east)}, y={(screen1.north west)}]
      \draw[cstage, line width=1.5pt, rounded corners=1.5pt] (0.18, 0.716) rectangle (0.295, 0.755);
      \node[fill=cstage, text=white, circle, inner sep=0.2pt, font=\bfseries\fontsize{3.5pt}{3.5pt}\selectfont] at (0.31, 0.735) {2};

      \draw[cstage, line width=1.5pt, rounded corners=1.5pt] (0.165, 0.90) rectangle (0.223, 0.94);
      \node[fill=cstage, text=white, circle, inner sep=0.2pt, font=\bfseries\fontsize{3.5pt}{3.5pt}\selectfont] at (0.235, 0.92) {1};

      \draw[-{Triangle[length=3pt,width=3pt]}, cstage, dashed, line width=1.2pt] (0.20, 0.90) -- (0.237, 0.755) node[midway, right, fill=cstage, text=white, rounded corners=1.5pt, font=\bfseries\fontsize{4.5pt}{4.5pt}\selectfont, inner sep=1pt, xshift=1mm] {Correct Path};
      
      \node[text=cfail, font=\bfseries\fontsize{7pt}{7pt}\selectfont] at (0.141, 0.92) {$\times$};
  \end{scope}
  
  \node[thoughtbox, text width=38mm, anchor=north west, minimum height=49mm] (thought1) at (133.5, 116.5) {
      \textbf{\textcolor{carrow}{Agent Thought:}}\\[1mm]
      I am at stage 2. According to \textcolor{cplan}{planning}:\\[1.5mm]
      \textit{``Key note: In GIMP, contrast adjustment is under the 'Colors' menu''}\\[1.5mm]
      \colorbox{cplan!15}{\parbox{34mm}{This tells me to navigate to\\Colors, NOT Image menu.}}\\[1.5mm]
      Looking at screenshot, I see ``Colors'' between ``Image'' and ``Tools''.\\[1.5mm]
      \textbf{Action:} Click ``Colors'' menu.\\[0.5mm]
      \hspace*{4mm}\texttt{<left\_click> [192, 69]}
  };
  
  \begin{scope}[shift={([xshift=-7mm, yshift=-6.0mm]thought1.north east)}, scale=2.4]
      \fill[cyan!40, rounded corners=2pt] (-1.2, -2.5) rectangle (1.2, -0.2);
      \fill[cyan!80!black] (-0.8, -2) rectangle (0.8, -0.7);
      \draw[white, line width=1pt] (-0.4, -1.2) -- (-0.1, -1.5) -- (0.4, -0.9);
      \fill[cyan!40, rounded corners=1.5pt] (-1.7, -0.5) rectangle (-1.2, -1.8);
      \fill[cyan!40, rounded corners=1.5pt] (1.2, -0.5) rectangle (1.7, -1.8);
      \fill[gray!50] (-0.4, -0.2) rectangle (0.4, 0);
      \fill[cyan!40, rounded corners=3pt] (-1.4, 0) rectangle (1.4, 2.2);
      \fill[gray!50, rounded corners=1pt] (-1.7, 0.6) rectangle (-1.4, 1.4);
      \fill[gray!50, rounded corners=1pt] (1.4, 0.6) rectangle (1.7, 1.4);
      \fill[white] (-0.6, 1.2) circle (0.4);
      \fill[white] (0.6, 1.2) circle (0.4);
      \fill[cyan!80!black] (-0.6, 1.2) circle (0.15);
      \fill[cyan!80!black] (0.6, 1.2) circle (0.15);
      \draw[cyan!80!black, line width=1.2pt] (-0.4, 0.5) .. controls (-0.1, 0.2) and (0.1, 0.2) .. (0.4, 0.5);
      \draw[gray!80, line width=1.5pt] (0, 2.2) -- (0, 3.0);
      \fill[orange] (0, 3.2) circle (0.25);
  \end{scope}

  \draw[-{Triangle}, cplan, dashed, line width=0.8pt] (pk.east) -- (screen1.west);
  \draw[-{Triangle}, cplan, dashed, line width=0.8pt] (screen1.east) -- (thought1.west);

  \draw[arrowconnect] (87, 65) -- (87, 53);
  \node[font=\itshape\fontsize{7.5pt}{7.5pt}\selectfont, text=cmuted, fill=white] at (87, 59) {Step 1 succeeded, proceeding to Step 3...};

  \node[anchor=north west, font=\fontsize{9pt}{9pt}\bfseries\selectfont] at (0, 51) {(b) Step 3 / 6: Grounding Locates UI Element};

  \fill[cbgground, rounded corners=4pt] (0, -3) rectangle (174, 46);
  \draw[cground, line width=1pt, rounded corners=4pt] (0, -3) rectangle (174, 46);

  \node[fill=white, draw=cground!30, rounded corners=3pt, minimum width=43mm, minimum height=46mm, anchor=north west] (gk) at (1.5, 44.5) {};
  \fill[cground, rounded corners=2pt] (1.5, 44.5) rectangle (2.3, -1.5); 
  \node[fill=cground, text=white, rounded corners=2pt, font=\bfseries\fontsize{6.5pt}{6.5pt}\selectfont, anchor=north west] at (3, 43) {Grounding Knowledge};

  \node[anchor=north west, text width=38mm, align=left, font=\fontsize{7pt}{8pt}\selectfont] at (3, 38) {
      \textbf{Target Element 3:}\\
      Role: Contrast Slider\\[1.5mm]
      \colorbox{cground!15}{\parbox{35mm}{\textbf{Visual Desc: Horizontal bar}\\ \textbf{labeled ``Contrast'', beneath}\\ \textbf{the ``Brightness'' slider.} \textcolor{cground}{$\blacktriangleright$}}} \\[1.5mm]
      Function: Drag to boost contrast.
  };

  \node[anchor=north west, inner sep=0] (screen2) at (49, 45) {
      \includegraphics[width=78mm]{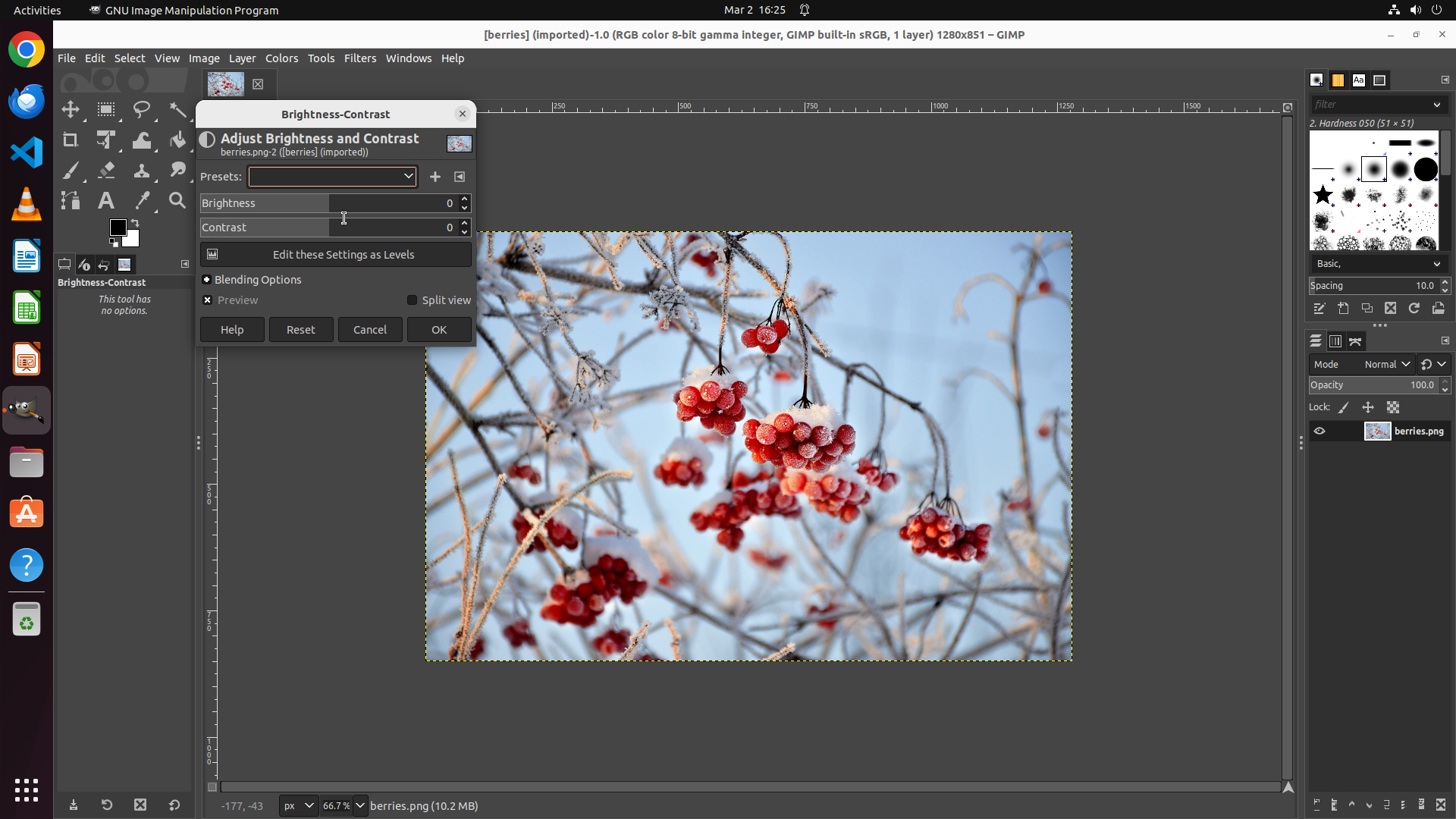}
  };
  \draw[carrow!30, line width=0.5pt, rounded corners=2pt] (screen2.south west) rectangle (screen2.north east);

  \begin{scope}[shift={(screen2.south west)}, x={(screen2.south east)}, y={(screen2.north west)}]
      \draw[cstage, line width=1.5pt, rounded corners=1.5pt] (0.125, 0.699) rectangle (0.341, 0.745);
      \node[fill=cstage, text=white, rounded corners=1.5pt, font=\bfseries\fontsize{5pt}{5pt}\selectfont, inner sep=1pt] at (0.43, 0.722) {Target: Contrast};
  \end{scope}
  
  \node[thoughtbox, text width=38mm, anchor=north west, minimum height=46mm] (thought2) at (133.5, 44.5) {
      \textbf{\textcolor{carrow}{Agent Thought:}}\\[1mm]
      Need to identify Contrast slider.\\
      \textcolor{cground}{Grounding} says it is \textit{``labeled 'Contrast', beneath 'Brightness' slider''}.\\[1.5mm]
      \colorbox{cground!15}{\parbox{34mm}{I identify the lower slider\\matching this description.}}\\[1.5mm]
      \textbf{Action:} Drag Contrast slider handle.\\
      \hspace*{4mm}\texttt{<drag>} \texttt{[234, 276]} \texttt{\scalebox{0.8}{$\to$}} \texttt{[274, 276]}
  };
  
  \begin{scope}[shift={([xshift=-7mm, yshift=-6.0mm]thought2.north east)}, scale=2.4]
      \fill[cyan!40, rounded corners=2pt] (-1.2, -2.5) rectangle (1.2, -0.2);
      \fill[cyan!80!black] (-0.8, -2) rectangle (0.8, -0.7);
      \draw[white, line width=1pt] (-0.4, -1.2) -- (-0.1, -1.5) -- (0.4, -0.9);
      \fill[cyan!40, rounded corners=1.5pt] (-1.7, -0.5) rectangle (-1.2, -1.8);
      \fill[cyan!40, rounded corners=1.5pt] (1.2, -0.5) rectangle (1.7, -1.8);
      \fill[gray!50] (-0.4, -0.2) rectangle (0.4, 0);
      \fill[cyan!40, rounded corners=3pt] (-1.4, 0) rectangle (1.4, 2.2);
      \fill[gray!50, rounded corners=1pt] (-1.7, 0.6) rectangle (-1.4, 1.4);
      \fill[gray!50, rounded corners=1pt] (1.4, 0.6) rectangle (1.7, 1.4);
      \fill[white] (-0.6, 1.2) circle (0.4);
      \fill[white] (0.6, 1.2) circle (0.4);
      \fill[cyan!80!black] (-0.6, 1.2) circle (0.15);
      \fill[cyan!80!black] (0.6, 1.2) circle (0.15);
      \draw[cyan!80!black, line width=1.2pt] (-0.4, 0.5) .. controls (-0.1, 0.2) and (0.1, 0.2) .. (0.4, 0.5);
      \draw[gray!80, line width=1.5pt] (0, 2.2) -- (0, 3.0);
      \fill[orange] (0, 3.2) circle (0.25);
  \end{scope}

  \draw[-{Triangle}, cground, dashed, line width=0.8pt] (gk.east) -- (screen2.west);
  \draw[-{Triangle}, cground, line width=1pt] (39.5, 23) to[out=0, in=180] (58.5, 31.5); 

  \draw[-{Triangle}, cground, dashed, line width=0.8pt] (screen2.east) -- (thought2.west);

  \node[anchor=south east, font=\bfseries\fontsize{8.5pt}{8.5pt}\selectfont, text=cstage] at (172, 0) {
      \Huge \textcolor{cstage}{$\checkmark$} \normalsize Task Succeeded (1.0)
  };

  \end{tikzpicture}
  }
  \definecolor{cfail}{RGB}{200,50,50}
  \definecolor{cstage}{RGB}{44,160,44}
  \caption{\textbf{Qualitative example on a GIMP contrast task.}
    \textbf{(a)}~\textcolor{planblue}{Planning} redirects the agent from the conventional ``Image'' menu to GIMP's correct ``Colors'' menu.
    \textbf{(b)}~\textcolor{groundorange}{Grounding} identifies the Contrast slider among visually similar controls.}
  \label{fig:qualitative}
\end{figure*}

\cref{fig:qualitative} presents a representative example from the GIMP domain to illustrate how \guidename{}'s dual-channel knowledge injection resolves domain bias at both the planning and grounding levels during task execution.
\vspace{-1mm}
\paragraph{Planning Knowledge Resolves Workflow Ambiguity.}
In \cref{fig:qualitative}(a), a general-purpose agent would choose the ``Image'' menu, the conventional location in other editors.
GIMP instead places contrast controls under ``Colors'', creating a typical \emph{planning-level bias}.
The retrieved planning knowledge states this distinction, letting the agent select the correct menu path without trial-and-error.
\vspace{-1mm}
\paragraph{Grounding Knowledge Enables Precise Element Identification.}
In \cref{fig:qualitative}(b), the dialog presents multiple visually similar sliders.
Grounding knowledge provides a discriminative description---``horizontal bar labeled `Contrast', beneath the `Brightness' slider''---enabling the agent to identify and manipulate the correct control.
Together, planning addresses \emph{what to do} and grounding addresses \emph{where to act}---neither alone suffices, consistent with the ablation results in \cref{sec:ablation}.
\vspace{-1mm}
\paragraph{Failure Cases.}
Injected knowledge can also slightly degrade performance when (i)~the retrieved video is procedurally mismatched to the target task, misleading the agent with an inapplicable workflow, or (ii)~the tutorial's UI layout differs substantially from the evaluation environment, invalidating grounding descriptions.
Further examples are provided in the supplementary material.
\vspace{-2mm}
\subsection{Annotation Pipeline Quality}
\label{sec:pipeline-quality}
\vspace{-1mm}
We audit retrieval coverage and two pipeline stages with human labels.
\vspace{-1mm}
\paragraph{Retrieval Coverage.}
Of 361 OSWorld tasks, 299 (82.8\%) retrieve at least one relevant tutorial video, and 42.7\% of covered tasks retrieve two videos.
All covered tasks complete download, keyframe extraction, annotation, and knowledge decomposition; the remaining 62 tasks fall back to baseline behavior without injected knowledge.

\begin{figure}[tb]
  \centering
  \definecolor{myblue}{RGB}{235, 243, 250}
  \definecolor{mydarkblue}{RGB}{31, 119, 180}
  \definecolor{myorange}{RGB}{253, 243, 231}
  \definecolor{mydarkorange}{RGB}{214, 125, 36}
  \definecolor{mygreen}{RGB}{232, 245, 233}
  \definecolor{mydarkgreen}{RGB}{44, 160, 44}
  \definecolor{myred}{RGB}{253, 235, 235}
  \definecolor{mydarkred}{RGB}{214, 39, 40}
  \definecolor{mygray}{RGB}{245, 245, 245}
  \definecolor{mydarkgray}{RGB}{150, 150, 150}
  
  \resizebox{0.70\textwidth}{!}{
  \begin{tikzpicture}[font=\sffamily\scriptsize]

    \begin{scope}[xshift=0cm]
      \node[font=\bfseries\small, anchor=center] at (1.8, 4.5) {(a) GUI Video Classification};

      \draw[fill=mygreen, draw=mydarkgreen, thick, rounded corners=2pt] (0, 1.8) rectangle (1.8, 3.6);
      \node[align=center, text=mydarkgreen] at (0.9, 2.7) {\textbf{TP}\\[2pt] 171 (57\%)};
      
      \draw[fill=myorange, draw=mydarkorange, thick, rounded corners=2pt] (1.8, 1.8) rectangle (3.6, 3.6);
      \node[align=center, text=mydarkorange] at (2.7, 2.7) {\textbf{FN}\\[2pt] 17 (6\%)};

      \draw[fill=mygray, draw=mydarkgray, thick, rounded corners=2pt] (0, 0) rectangle (1.8, 1.8);
      \node[align=center, text=mydarkgray] at (0.9, 0.9) {\textbf{FP}\\[2pt] 0 (0\%)};

      \draw[fill=mygreen, draw=mydarkgreen, thick, rounded corners=2pt] (1.8, 0) rectangle (3.6, 1.8);
      \node[align=center, text=mydarkgreen] at (2.7, 0.9) {\textbf{TN}\\[2pt] 112 (37\%)};

      \node[anchor=south, font=\scriptsize\bfseries, align=center] at (0.9, 3.65) {Pred:\\GUI};
      \node[anchor=south, font=\scriptsize\bfseries, align=center] at (2.7, 3.65) {Pred:\\Non-GUI};
      
      \node[rotate=90, anchor=south, font=\scriptsize\bfseries, align=center] at (-0.15, 2.7) {Actual:\\GUI};
      \node[rotate=90, anchor=south, font=\scriptsize\bfseries, align=center] at (-0.15, 0.9) {Actual:\\Non-GUI};

      \node[anchor=west, align=center, font=\scriptsize] at (3.8, 2.7) {Accuracy:\\[2pt]\textbf{94.3\%}};
      \node[anchor=west, align=center, font=\scriptsize] at (3.8, 1.8) {Precision:\\[2pt]\textbf{100\%}};
      \node[anchor=west, align=center, font=\scriptsize] at (3.8, 0.9) {Recall:\\[2pt]91.0\%};
    \end{scope}

    \begin{scope}[xshift=6.5cm]
      \node[font=\bfseries\small, anchor=center] at (2.4, 4.5) {(b) Topic Extraction Scores};

      \begin{scope}[yshift=0.3cm]
      \draw[->, thick, mydarkgray] (0,0) -- (4.8,0);
      \draw[->, thick, mydarkgray] (0,0) -- (0,3.3) node[above, text=black] {\%};

      \foreach \y/\ytext in {1/25, 2/50, 3/75} {
        \draw[mydarkgray] (-0.1, \y) -- (0, \y);
        \node[anchor=east, font=\scriptsize, text=mydarkgray] at (-0.1, \y) {\ytext};
      }

      \draw[fill=mygreen, draw=mydarkgreen, thick, rounded corners=1pt] (0.4, 0) rectangle (1.4, 3.09);
      \node[anchor=south, font=\scriptsize, text=black] at (0.9, 3.09) {77.3\%};
      \node[anchor=north, align=center, font=\scriptsize, text=black] at (0.9, -0.1) {\textbf{1.0}\\[1pt](Accurate)};

      \draw[fill=myorange, draw=mydarkorange, thick, rounded corners=1pt] (1.9, 0) rectangle (2.9, 0.75);
      \node[anchor=south, font=\scriptsize, text=black] at (2.4, 0.75) {18.7\%};
      \node[anchor=north, align=center, font=\scriptsize, text=black] at (2.4, -0.1) {\textbf{0.5}\\[1pt](Partial)};

      \draw[fill=myred, draw=mydarkred, thick, rounded corners=1pt] (3.4, 0) rectangle (4.4, 0.16);
      \node[anchor=south, font=\scriptsize, text=black] at (3.9, 0.16) {4.0\%};
      \node[anchor=north, align=center, font=\scriptsize, text=black] at (3.9, -0.1) {\textbf{0.0}\\[1pt](Failed)};

      \draw[mydarkgray, rounded corners=2pt, densely dashed] (1.6, 2.0) rectangle (4.5, 3.2);
      \node[anchor=center, font=\scriptsize, align=center, text=black] at (3.05, 2.6) {Mean Score: \textbf{0.867}\\[3pt]Acceptable: \textbf{96\%}};
      \end{scope}

    \end{scope}

  \end{tikzpicture}
  }
  \vspace{-4mm}
  \caption{\textbf{Human evaluation of retrieval/annotation quality.}
    Three annotators score 300 videos per stage:
    \textbf{(a)}~GUI classification keeps 100\% precision;
    \textbf{(b)}~topic extraction reaches mean 0.867 and 96\% acceptable.}
  \label{fig:rag-eval}
\end{figure}

\paragraph{GUI Domain Classification (Stage~1).}
Three annotators judge 300 candidate videos for actual GUI demonstrations (\cref{fig:rag-eval}a).
The subtitle-assisted classifier reaches \textbf{100\%} precision and 90.96\% recall (F1\,$=$\,95.27\%), preventing non-GUI videos from entering annotation; false negatives are mostly visual-only tutorials whose narration lacks GUI interaction terms.
\vspace{-1mm}
\paragraph{Topic Extraction (Stage~2).}
On 300 confirmed GUI videos, annotators score extracted topics as 1.0/0.5/0.0 for accurate/partial/failed matching.
The mean score is 0.867 with 96.00\% acceptable topics (\cref{fig:rag-eval}b); only 4.00\% fail, mainly due to poor auto-generated subtitles, confirming that subtitle-derived topics provide reliable anchors for relevance matching (\cref{sec:video-rag}).
\vspace{-1mm}

\section{Conclusion}
\label{sec:conclusion}
\vspace{-1mm}
We presented \guidename{}, a training-free framework that mitigates GUI-agent domain bias by acquiring procedural expertise from web tutorial videos.
Its subtitle-driven Video-RAG and VLM pairwise video annotation produce transferable \plan{} and \ground{} knowledge without changing model weights.
OSWorld gains of \textbf{+4.47--+7.48 percentage points}, WindowsAgentArena transfer~\cite{windowsagentarena}, and controlled ablations show that structured video knowledge is a scalable resource for GUI agents.

\paragraph{Acknowledgments.}
We thank the anonymous reviewers, area chairs, and human annotators for their constructive feedback and evaluation support.

\bibliographystyle{splncs04}
\bibliography{main}

\clearpage
\begin{center}
{\Large\bfseries Supplementary Material}
\end{center}
\vspace{1em}

\appendix
\section{Prompt Structures for Agent Integration}
\label{sec:appendix-prompts}

This appendix provides the full prompt structures used for injecting \guidename{} knowledge into both integration modes.

\subsection{Mode~A: AgentS Multi-Agent System}

\paragraph{Worker System Prompt (Planning Injection).}
The Worker system prompt is constructed from a base procedural memory template (action API documentation and response format instructions).
When \plan{} knowledge is available, the following block is inserted verbatim into the guidelines section before being passed to the model:
\begin{promptbox}
[Agent guidelines and action API]

### Reference Plan from a Similar Task
A step-by-step plan was extracted from a demonstration
video showing how a similar task was completed. Use it as
a starting point to guide your approach -- follow the
general workflow, and adapt each step to the current task
and UI state as needed.

VIDEO_PLANNING

> If the plan does not match your current UI state or task
> requirements, trust what you see in the screenshot and
> proceed independently.

### END OF GUIDELINES
[Screenshot + interaction history]
\end{promptbox}
\noindent
\texttt{VIDEO\_PLANNING} is replaced at runtime with the \plan{} text retrieved for the current task.
When no \plan{} knowledge is retrieved the entire block is omitted.

\paragraph{Grounding Agent Prompt (Grounding Injection).}
The visual grounding agent is queried once per action to resolve a natural-language element description into screen coordinates.
When \ground{} knowledge is available, the prompt is:
\begin{promptbox}
Reference knowledge from similar tasks (element
locations and descriptions):
{video_grounding}

Based on the screenshot, locate the target element:
{element_description}
Output only the coordinate of one point in your response.
[Current screenshot]
\end{promptbox}
\noindent
When no \ground{} knowledge is retrieved, the first paragraph is omitted and only the \texttt{element\_description} and screenshot are sent.

\subsection{Mode~B: Single-Model Agent (Qwen3-VL / Seed-1.8)}

\paragraph{Unified System Prompt.}
Both knowledge streams are injected into a single system prompt that also contains the tool definitions.
The knowledge block appended after the tool schema is shown below (square-bracket notes are editorial):
\begin{promptbox}
[Tool definitions and function call schema]

# External Knowledge from Similar Tasks

## Video Planning Reference (Optional Guidance)
You are provided with planning trajectories from
demonstration videos of similar tasks. These serve as
**reference materials only** and should be used with
caution.

**IMPORTANT: Use as Reference, Not Instruction**:
- The video planning is a **suggestion**, not a
  requirement -- always prioritize your own analysis
  of the current task and screenshot
- Video demonstrations may differ from your actual task
  in important ways (different data, different UI state,
  different requirements)
- **Only use video planning when**:
  * The task is highly similar (same application, same
    type of operation, same general workflow)
  * The planning clearly aligns with your current task
    requirements
  * You can verify each suggested step applies to your
    specific situation

**How to Use Video Planning Safely**:
- **Extract General Patterns**: Focus on the overall
  workflow structure, not specific UI elements
- **Verify Applicability**: Confirm each suggestion
  makes sense for your current screenshot and task
- **Adapt Critically**: Modify the approach to fit your
  specific requirements -- never blindly follow steps
- **Trust Your Analysis**: If the video planning
  conflicts with what you observe in the screenshot,
  trust the screenshot

**When to Ignore Video Planning**:
- Low relevance: different application, operation type,
  or workflow
- Conflicts with current state: suggestions don't match
  your screenshot
- Unclear or ambiguous: default to your own analysis

**Video Planning for Similar Tasks**:
{video_planning}

**Critical Reminder**: Always verify each step against
the current screenshot and task requirements before
proceeding. Your independent analysis takes priority.

## Video Grounding Reference (UI Elements from Similar
   Tasks)
You are provided with GUI element descriptions from
demonstration videos. These include element appearance,
position, and function.

**IMPORTANT: Use with Caution**:
- Descriptions come from **similar but different tasks**
  and may not match your current UI
- Use as **hints about what to look for**, not as
  precise instructions

**How to Use Video Grounding Safely**:
- **Verify Against Screenshot**: Always check if
  described elements actually exist in your current
  screenshot before acting on them
- **Look for Analogous Elements**: If described elements
  don't exist, look for similar ones serving the same
  purpose

**GUI Elements from Similar Tasks**:
{video_grounding}

**Critical Reminder**: Always verify against the actual
screenshot before taking any action. The screenshot is
your source of truth.
\end{promptbox}
\noindent
When only one channel is retrieved the corresponding subsection is omitted entirely.

\paragraph{Structured Thought Template.}
The response format enforces step-by-step reasoning that actively references both knowledge channels.
The full format (for the case where both \plan{} and \ground{} knowledge are available) is:
\begin{promptbox}
# Response format

For each step, output exactly in this order:

1) **Thought**: Analyze the current screenshot. Use the
   video planning to identify which stage of the
   workflow you are in and what to do next. Use the
   video grounding to help locate the relevant UI
   element -- then verify it exists in your current
   screenshot. Write 2-4 concise sentences.
2) **Action**: A short imperative describing what to do.
3) A single <tool_call>...</tool_call> block with JSON:
   {"name": "computer_use", "arguments": {...}}.

Rules:
- Always start your response with "Thought:".
- Use video planning for workflow understanding; use
  grounding to identify elements.
- Always verify elements in your current screenshot
  before acting. If they differ from the video, trust
  the screenshot.
- Output order: Thought -> Action -> tool_call.
- To finish the task, use action=terminate.

Example:
Thought: The video planning shows inserting a chart as
the next step after setting up the data. I am at that
stage now. The video grounding describes the "Insert"
menu tab as a text label near the top of the LibreOffice
Calc window below the title bar. I can see the Insert
tab in the menu bar of my current screenshot.
Action: Click the Insert tab to open chart insertion
options.
<tool_call>
{"name": "computer_use", "arguments":
  {"action": "left_click", "coordinate": [200, 45]}}
</tool_call>
\end{promptbox}
\noindent
This format instructs the model to: (1)~identify its current position in the video workflow using \plan{} knowledge, (2)~cross-reference \ground{} element descriptions against the live screenshot, and (3)~emit a grounded action.
Flexible degradation is built in: when only \plan{} is available the Thought step reads ``Reference the video planning to identify your current stage in the workflow''; when only \ground{} is available it reads ``Check the video grounding descriptions to understand what the element looks like, then locate it in your current screenshot''; when neither is retrieved the Thought simplifies to a one- to two-sentence screenshot analysis.

\section{Implementation Details and Benchmark}
\label{sec:appendix-impl}

\paragraph{Benchmark Details.}
OSWorld~\cite{osworld} executes tasks inside live Ubuntu virtual machines at $1920{\times}1080$ resolution.
Each task specifies an initial environment state and a natural-language instruction; task completion is assessed by execution-based evaluation scripts that compare the resulting system state against a gold reference, yielding a score in $[0,1]$.
The 361 evaluated tasks span 10 application domains: LibreOffice~Calc (47), Impress (47), Chrome (46), GIMP (26), OS operations (24), Writer (23), VS~Code (23), VLC (17), Thunderbird (15), and cross-application workflows (93).
The maximum number of interaction steps per task is set to 50.

\paragraph{Pipeline Configuration.}
\guidename{}'s retrieval pipeline uses GPT-4.1~\cite{gpt4o} for query generation, GPT-4.1-mini for query simplification, GUI domain classification, and relevance scoring, and OpenAI Whisper~\cite{whisper} (base model) for ASR with word-level timestamps.
The annotation pipeline uses OmniParser~\cite{omniparser} for UI element detection, GPT-5.1~\cite{gpt5} for frame-pair VLM pairwise annotation, and GPT-5.1 for planning-grounding decomposition.
Keyframe extraction uses MOG2~\cite{mog2} with foreground threshold\,$=$\,10\,000 for videos under 600\,s.
Adaptive top-$K$ retains at most $K{=}2$ videos with relevance $\geq 0.5$ for non-top-1 candidates.
For grounding injection, the default maximum number of UI elements is $k{=}7$; the main paper reports the corresponding ablation.
Seed-1.8 uses temperature\,$=$\,1.0 with thinking mode enabled; Qwen3-VL-8B uses temperature\,$=$\,1.0 with thinking mode enabled.

\section{Full Per-Domain Ablation Results}
\label{sec:appendix-ablation}

\cref{tab:ablation-full} provides the full per-domain breakdowns for the two ablation studies summarized in the main text.

\begin{table*}[tb]
  \caption{\textbf{Full per-domain ablation results} on the 50-task subset (Seed-1.8, score~\%).
    \textbf{(a)}~\textcolor{groundorange}{Grounding} element count $k$ ($k{=}0$: \textcolor{planblue}{planning}-only).
    \textbf{(b)}~Annotation model ($k{=}5$); all annotators share the same \guidename{} pipeline.
    Best result per domain in \textbf{bold}.
    Domain task counts: Chrome (6), GIMP (8), Calc (8), Impress (5), Writer (5), Multi (6), OS (1), ThBrd (3), VLC (5), VSCode (3).}
  \label{tab:ablation-full}
  \centering\small

  \textbf{(a) Element count $k$}\par\vspace{3pt}
  \resizebox{\textwidth}{!}{%
  \begin{tabular}{@{}cc cccccccccc@{}}
    \toprule
    $k$ & \textbf{Overall} & Chrome & GIMP & Calc & Impress & Writer & Multi & OS & ThBrd & VLC & VSCode \\
    \midrule
    0  & 41.82 & 33.3 & 37.5 & 37.5 & 20.0 & 39.4 & 58.7 & 100.0 & 66.7 & 20.0 & 66.7 \\
    1  & 45.00 & 33.3 & 50.0 & 37.5 & 40.0 & 80.0 & 25.0 & 100.0 & 33.3 & 40.0 & 66.7 \\
    3  & 49.69 & 50.0 & \textbf{55.0} & 25.0 & 20.0 & 40.0 & 65.6 & 100.0 & 66.7 & 38.1 & 66.7 \\
    \rowcolor{placeholderBg}
    \textbf{5} & \textbf{53.81} & 50.0 & 37.5 & 37.5 & \textbf{60.0} & \textbf{80.0} & 50.0 & 100.0 & 66.7 & 58.1 & 66.7 \\
    7  & 47.88 & 33.3 & 50.0 & 37.5 & 20.0 & 60.0 & 65.7 & 100.0 & 66.7 & 60.0 & 33.3 \\
    10 & 48.66 & \textbf{50.0} & 25.0 & 37.5 & 40.0 & 40.0 & \textbf{73.7} & 100.0 & 66.7 & 58.1 & 66.7 \\
    15 & 48.66 & \textbf{50.0} & 25.0 & 37.5 & \textbf{60.0} & 60.0 & 57.1 & 100.0 & 66.7 & 58.1 & 33.3 \\
    \bottomrule
  \end{tabular}}

  \vspace{8pt}
  \textbf{(b) Annotation model}\par\vspace{3pt}
  \resizebox{\textwidth}{!}{%
  \begin{tabular}{@{}lc cccccccccc@{}}
    \toprule
    Annotator & \textbf{Overall} & Chrome & GIMP & Calc & Impress & Writer & Multi & OS & ThBrd & VLC & VSCode \\
    \midrule
    No Annotation & 32.0 & 16.7 & 25.0 & 25.0 & 20.0 & 20.0 & 50.0 & 100.0 & 66.7 & 20.0 & 66.7 \\
    GPT-4.1 Mini  & 47.7 & 50.0 & 37.5 & 37.5 & 20.0 & 40.0 & \textbf{65.6} & 100.0 & 66.7 & 58.1 & 66.7 \\
    Qwen3-VL-8B   & 51.8 & \textbf{50.0} & \textbf{62.5} & 25.0 & 40.0 & 60.0 & 50.0 & 100.0 & 66.7 & 58.1 & 66.7 \\
    GPT-5.1        & 53.8 & \textbf{50.0} & 37.5 & 37.5 & 60.0 & \textbf{80.0} & 50.0 & 100.0 & 66.7 & 58.1 & 66.7 \\
    \rowcolor{placeholderBg}
    \textbf{Seed-1.8} & \textbf{57.8} & 33.3 & \textbf{62.5} & \textbf{50.0} & \textbf{80.0} & 60.0 & 50.0 & 100.0 & 66.7 & 58.1 & 66.7 \\
    \bottomrule
  \end{tabular}}
\end{table*}

\section{Meaningful Filter Evaluation}
\label{sec:appendix-meaningful}

We conduct a human verification study to evaluate whether the \texttt{Meaningful} filter correctly removes visually changing but semantically invalid annotation candidates.
Three annotators each label 100 randomly sampled tutorial videos, with 15 extracted keyframes per video, yielding 4,500 evaluated frames in total.
Each frame is labeled as \emph{GUI-valid}, \emph{non-GUI}, or \emph{idle no-action} (static content without an actual GUI operation).
We treat the latter two categories as invalid frames that should be filtered out by \texttt{Meaningful}.
Labeling one video's 15 extracted keyframes takes approximately 4.5 minutes, yielding 1,350 minutes (22.5 hours) for the 300-video study, or 450 minutes (7.5 hours) per annotator.

Among the 4,500 evaluated frames, 1,227 are labeled invalid by humans, including 1,031 non-GUI frames and 196 idle no-action frames.
The system correctly filters 1,124 invalid frames and mistakenly filters 47 valid GUI frames.
This yields 96.0\% precision, 91.6\% recall, and 93.7\% F1 for invalid-frame filtering overall.
The filter is especially effective on non-GUI content, correctly removing 1,019/1,031 non-GUI frames (98.8\%), while idle no-action frames are more challenging at 105/196 (53.6\%).
Most false positives come from genuinely valid frames with only slight cursor movement or limited visible change, which makes the interaction appear semantically weak despite being part of a real operation.

\begin{figure*}[t]
  \centering
  \setlength{\tabcolsep}{0pt}
  \renewcommand{\arraystretch}{1.03}
  \begin{tabular}{@{}c@{\hspace{1.2mm}}c@{\hspace{1.2mm}}c@{}}
    \multicolumn{3}{c}{\textbf{(a) Non-GUI frame pair correctly filtered by \texttt{Meaningful}}} \\
    $s_t$ & $s_{t+1}$ & OmniParser on $s_t$ \\
    \includegraphics[width=0.323\linewidth]{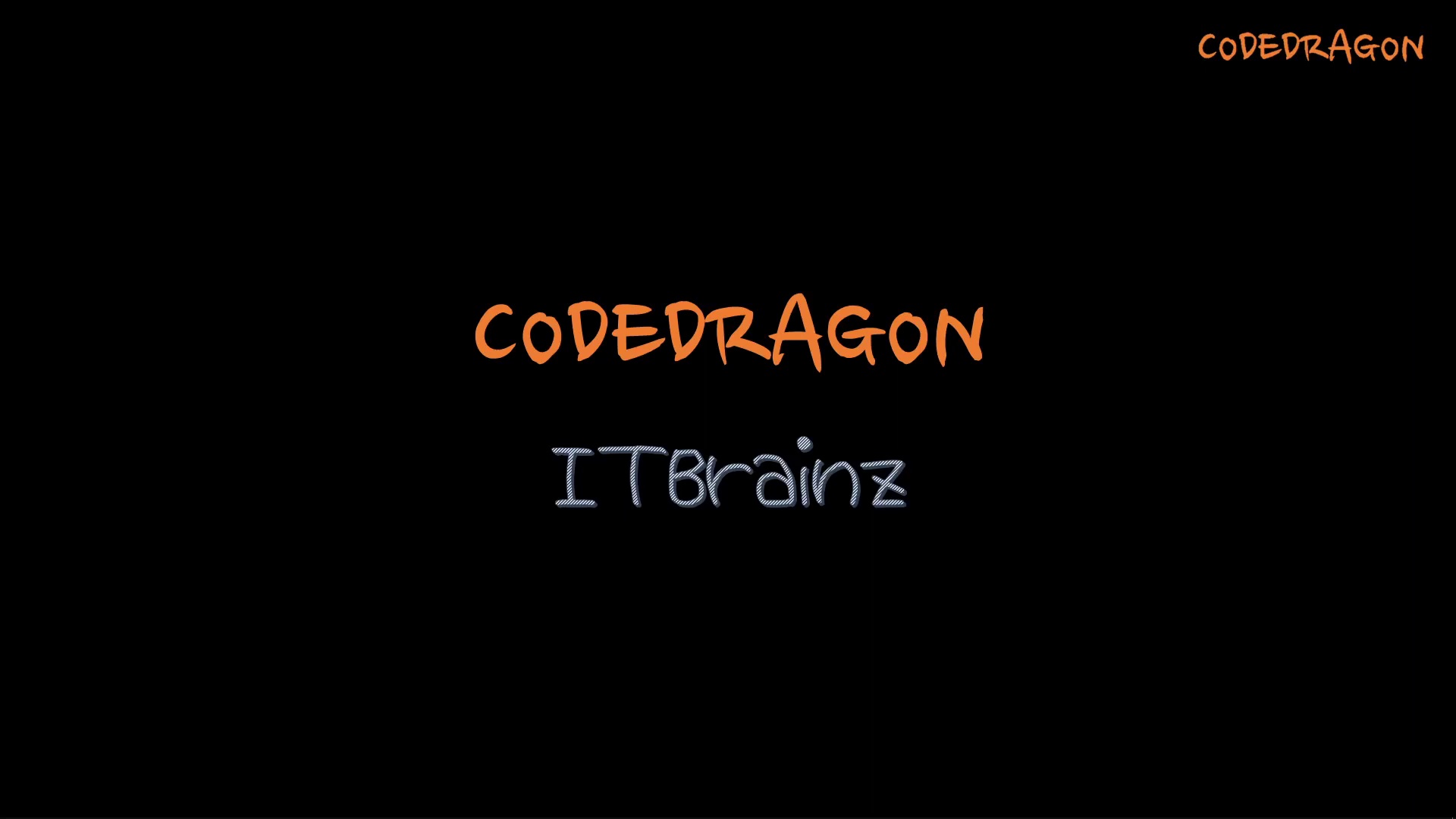} &
    \includegraphics[width=0.323\linewidth]{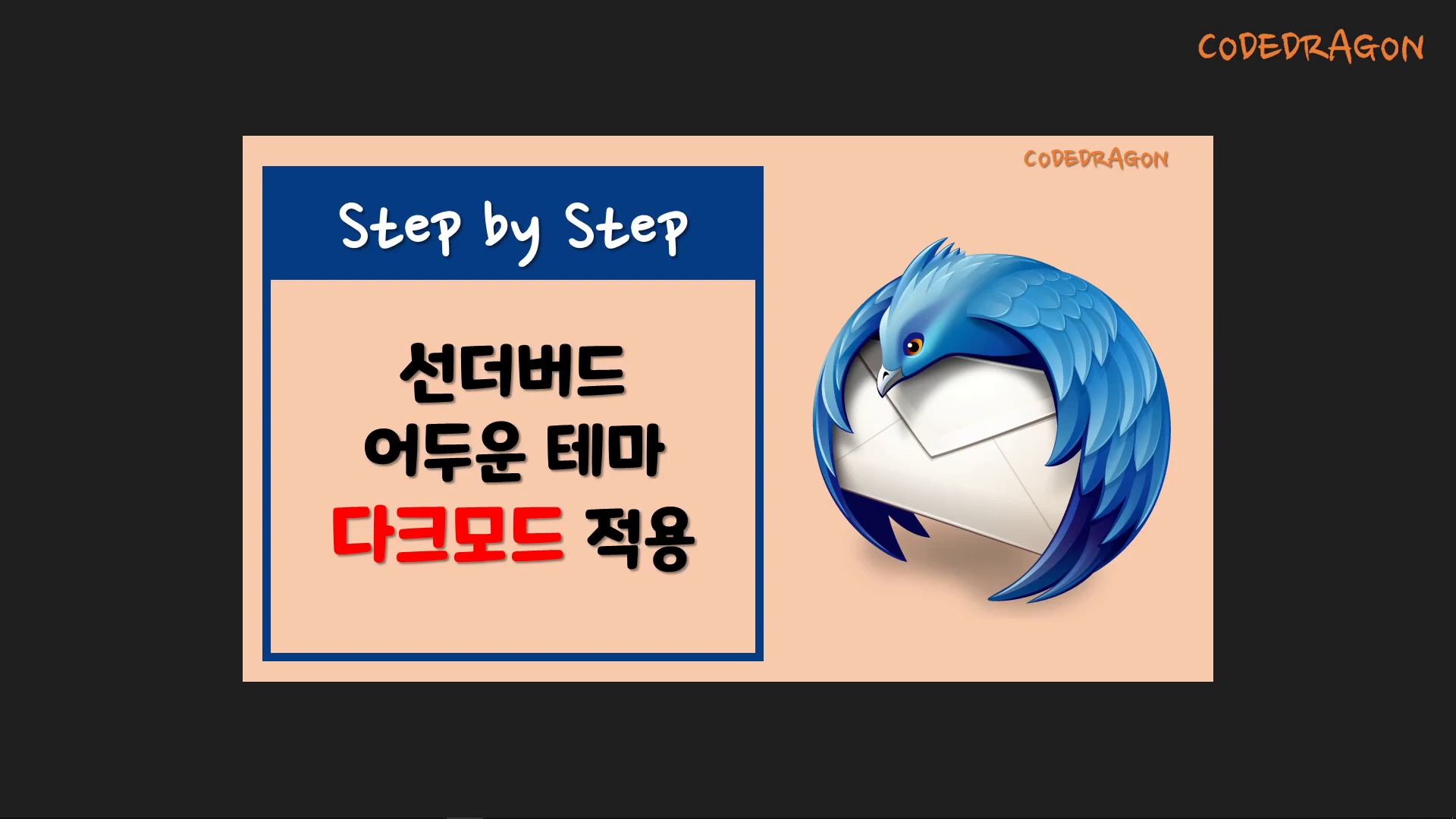} &
    \includegraphics[width=0.323\linewidth]{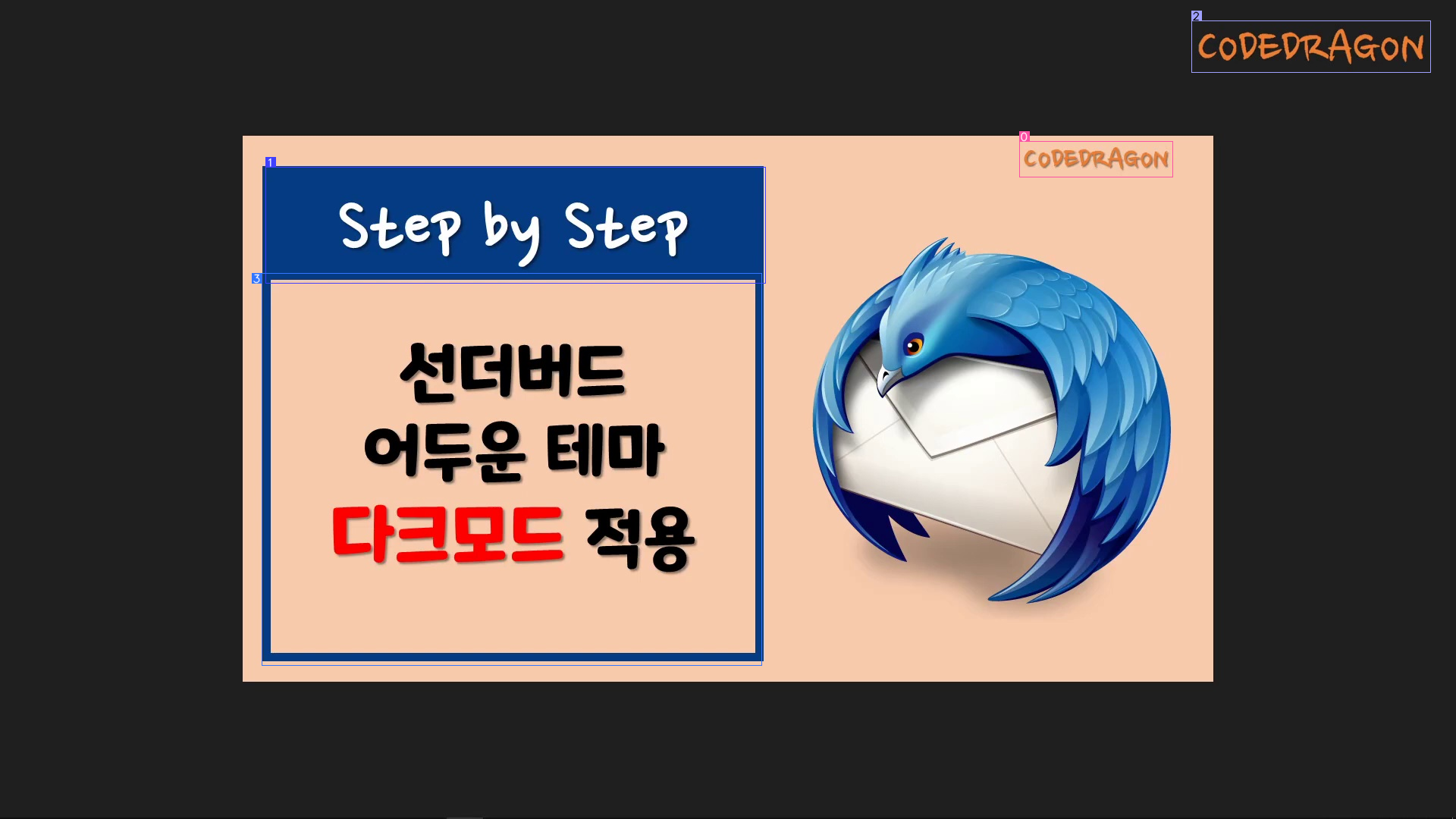} \\
    \multicolumn{3}{c}{\parbox{0.985\linewidth}{\centering\scriptsize OmniParser detects only a few generic text regions, providing little useful interactive structure for GUI grounding.}} \\[1.5mm]
    \multicolumn{3}{c}{\textbf{(b) Idle no-action GUI pair correctly filtered by \texttt{Meaningful}}} \\
    $s_t$ & $s_{t+1}$ & OmniParser on $s_t$ \\
    \includegraphics[width=0.323\linewidth]{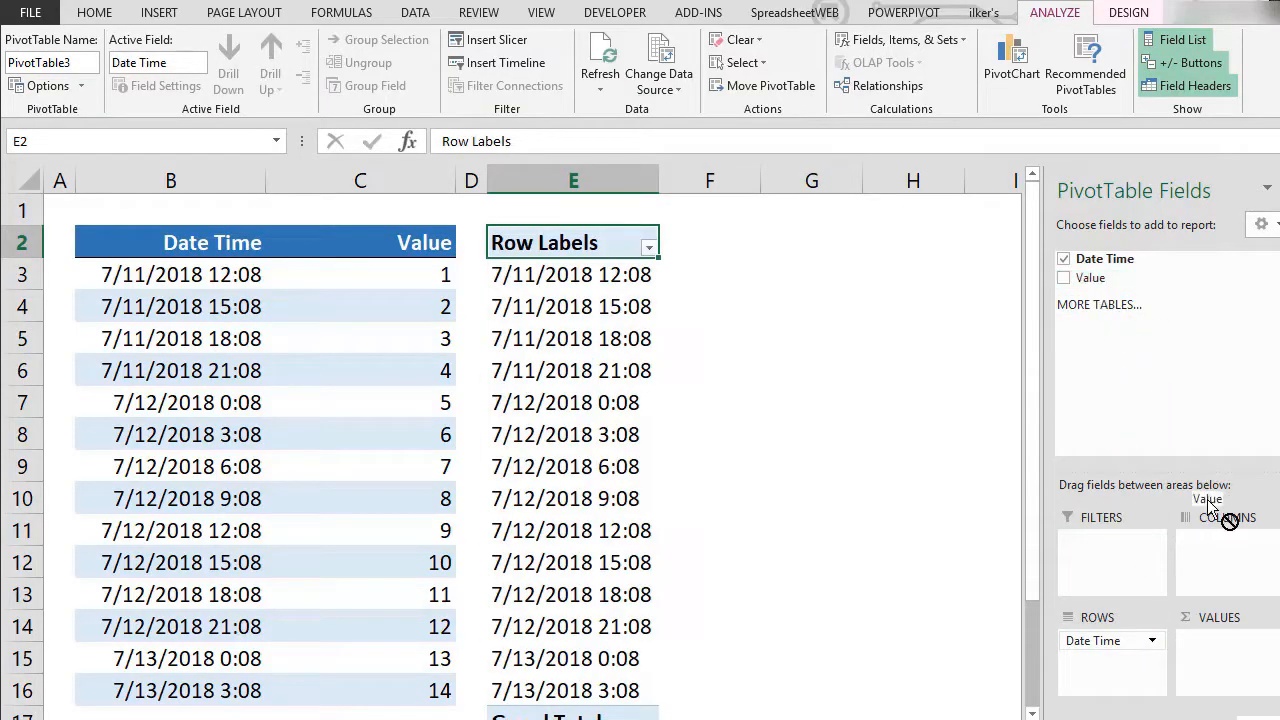} &
    \includegraphics[width=0.323\linewidth]{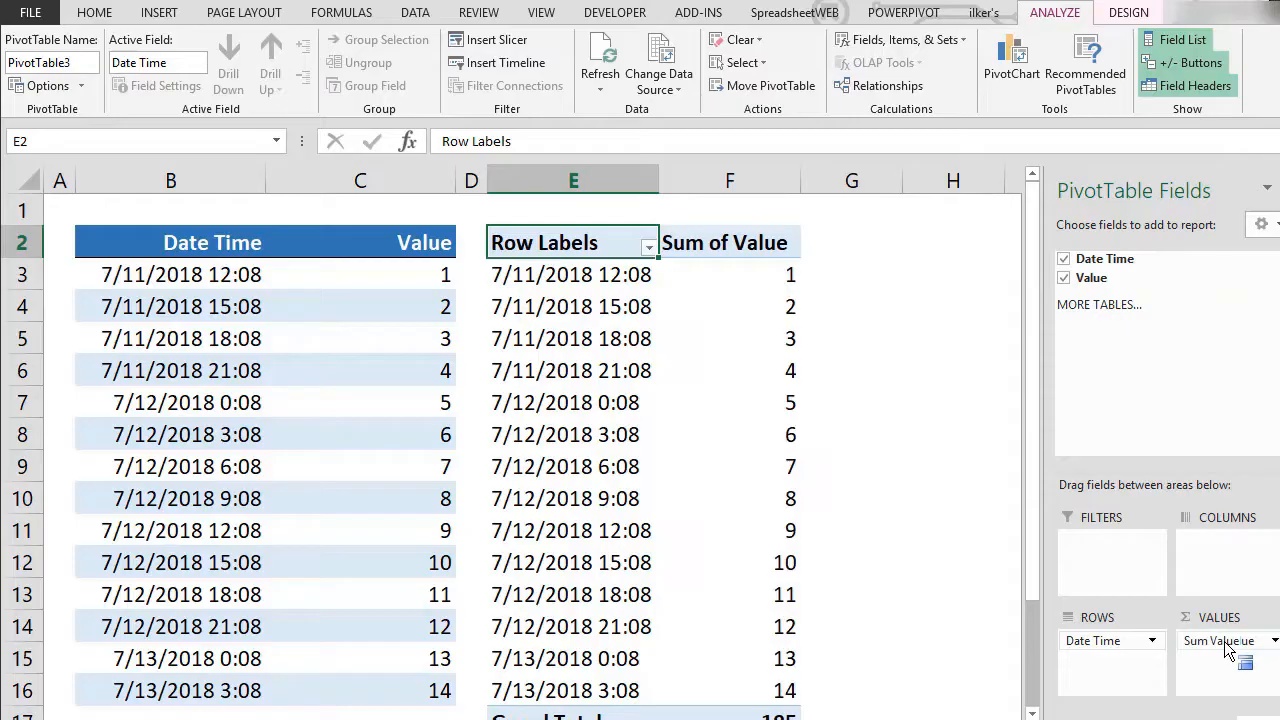} &
    \includegraphics[width=0.323\linewidth]{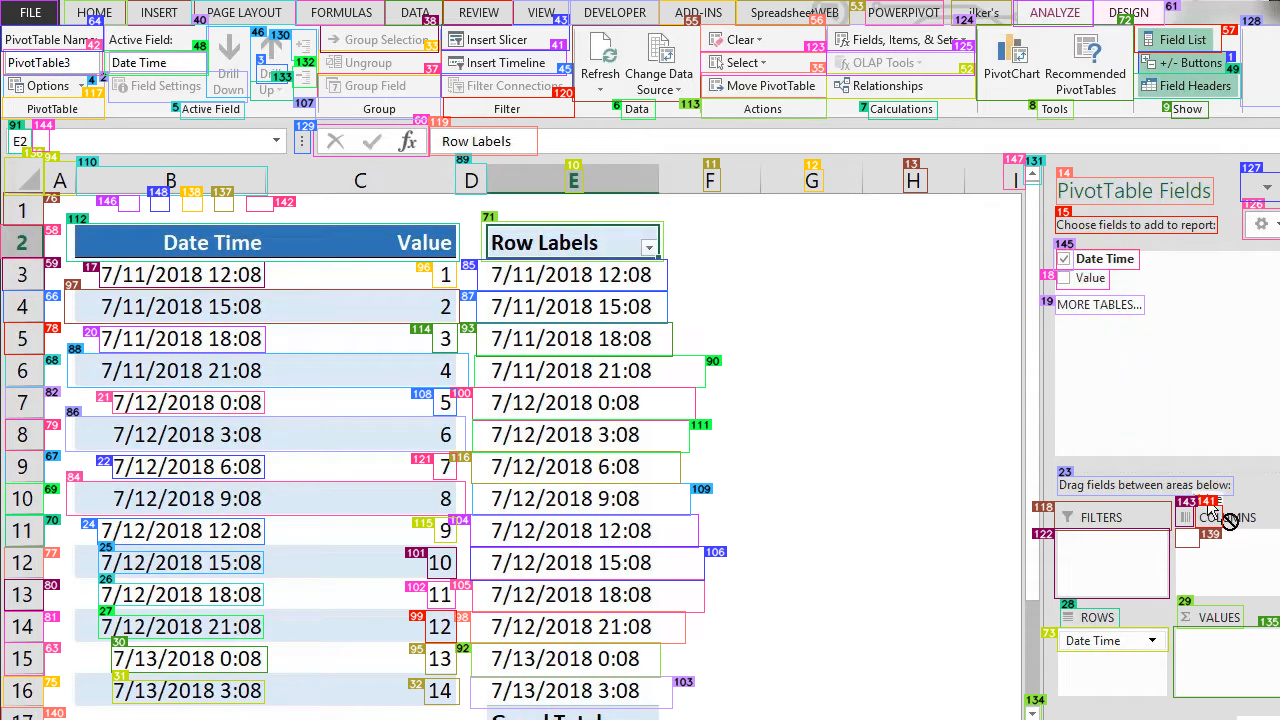} \\
    \multicolumn{3}{c}{\parbox{0.985\linewidth}{\centering\scriptsize Despite dense UI element detections, no task-progressing operation occurs between the two states.}}
  \end{tabular}
  \caption{\textbf{Representative examples for the \texttt{Meaningful} evaluation.}
  Top: a non-GUI intro pair from a retrieved video segment, where OmniParser captures only sparse text regions rather than actionable UI structure.
  Bottom: an idle no-action GUI pair, where the interface remains semantically valid and richly parsed, but the transition does not contribute transferable planning or grounding knowledge.}
  \label{fig:appendix-meaningful-examples}
\end{figure*}

\paragraph{Result Analysis.}
The observed error pattern is consistent with the design of \guidename{}'s annotation pipeline and the representative examples in \cref{fig:appendix-meaningful-examples}.
\texttt{Meaningful} is predicted jointly from the visual pair $(s_t, s_{t+1})$, the OmniParser element graphs $(E_t, E_{t+1})$, and the semantic context $(T_{\mathrm{topic}}, C_{\mathrm{sub}})$ within the pairwise video annotator $f_{\mathrm{VPA}}$.
Accordingly, non-GUI frames are comparatively easy to reject: their OmniParser outputs are often nearly empty or lack plausible interactive structure, and this weak element evidence is further reinforced by subtitle/topic context that does not describe an operation, yielding 98.8\% recall.

By contrast, idle no-action frames remain visually valid GUI states, with non-empty element graphs and domain-consistent interface structure.
Their difficulty lies not in GUI recognition but in temporal action inference: the model must determine that no task-progressing operation occurred between two visually similar interface states.
This is precisely the failure mode left unresolved by pixel-change-based keyframe extraction alone, and explains the lower recall on idle no-action frames (53.6\%).

The 47 false positives are mainly valid operations with weak cursor motion or limited appearance change, indicating a conservative bias when temporal evidence is subtle.
For the downstream pipeline, this trade-off is acceptable: excluding a small number of weak-but-valid steps is less harmful than allowing invalid transitions to contaminate the ``Thought \& Action NLP'' trajectory and the subsequent \plan{} / \ground{} decomposition.
Thus, the filter primarily serves to protect downstream knowledge quality.

\begin{table}[t]
  \caption{\textbf{Human evaluation of the \texttt{Meaningful} filter.}
  Category-specific numbers report filtering recall because the provided annotations support a clean subtype breakdown on false negatives but not on false positives.}
  \label{tab:appendix-meaningful}
  \centering\footnotesize
  \setlength{\tabcolsep}{5pt}
  \begin{tabular}{@{}lccc@{}}
    \toprule
    Evaluation target & Count & Correctly filtered & Metric \\
    \midrule
    Invalid frames (overall) & 1,227 & 1,124 & P\,=\,96.0, R\,=\,91.6, F1\,=\,93.7 \\
    Non-GUI frames & 1,031 & 1,019 & Recall\,=\,98.8 \\
    Idle no-action frames & 196 & 105 & Recall\,=\,53.6 \\
    \bottomrule
  \end{tabular}
\end{table}

\section{Annotation Cost Analysis}
\label{sec:appendix-cost}

We report the \emph{full pipeline API cost} under the default configuration described in \cref{sec:appendix-impl}: GPT-4.1 for query generation, GPT-4.1-mini for query simplification, GUI classification, and relevance scoring, and GPT-5.1 for frame-pair pairwise video annotation plus planning-grounding decomposition.
Whisper ASR and OmniParser are executed locally, so they are part of the pipeline but do not contribute API charges in this accounting.
All numbers below therefore measure \emph{external model API consumption only}.

\paragraph{Retrieval-stage accounting.}
The retrieval pipeline has four API stages: one GPT-4.1 query-generation call, one GPT-4.1-mini query-simplification call, approximately 15 GPT-4.1-mini GUI-classification/topic-extraction calls after metadata pre-filtering, and one GPT-4.1-mini batch relevance-scoring call over approximately 8 surviving GUI candidates.
The stage counts follow the empirical retrieval funnel: 50+ initial candidates $\rightarrow$ $\sim$15 metadata-valid candidates $\rightarrow$ $\sim$8 GUI videos $\rightarrow$ top-$K\leq2$ retained videos.
Prompt lengths are estimated directly from the implementation, using the subtitle truncation cap of 10,000 characters in the retrieval pipeline.

\begin{table}[t]
  \caption{\textbf{Retrieval API usage per task.}
    Query generation uses GPT-4.1 at \$2.00 / \$8.00 per 1M input / output tokens; the remaining retrieval calls use GPT-4.1-mini at \$0.40 / \$1.60.}
  \label{tab:appendix-retrieval-cost}
  \centering\footnotesize
  \setlength{\tabcolsep}{4pt}
  \begin{tabular}{@{}lcccc@{}}
    \toprule
    Stage & Calls & In/Out Tok. & Price & Cost \\
    \midrule
    Query generation & 1 & 109 / 10 & 2.00 / 8.00 & \$0.0003 \\
    Query simplification & 1 & 268 / 20 & 0.40 / 1.60 & \$0.00014 \\
    GUI class.\,+\,topic & $\sim$15 & 43.38K / 525 & 0.40 / 1.60 & \$0.0182 \\
    Relevance scoring & 1 & 436 / 25 & 0.40 / 1.60 & \$0.00021 \\
    \midrule
    \textbf{Total / task} & \textbf{$\sim$18} & \textbf{44.19K / 580} & \textbf{mixed} & \textbf{\$0.0188} \\
    \bottomrule
  \end{tabular}
\end{table}

\noindent
With current OpenAI API pricing, the retrieval stage costs about \$0.0188 per task.
This is small relative to annotation, but not negligible at benchmark scale, so we include it explicitly instead of reporting annotation cost alone.

\paragraph{Annotation-stage accounting.}
For annotation, we use file-measured statistics from an actual processed tutorial video in the current pipeline.
Each selected video yields 15 frame-pair GPT-5.1 calls (30 keyframes total), of which about 11 are meaningful and 4 are invalid/non-GUI transitions filtered by \texttt{Meaningful}.
The consolidated trajectory is then sent to GPT-5.1 once for planning decomposition and once for grounding decomposition.
Under the representative \emph{typical-task} regime, each valid frame uses approximately 10K OmniParser characters and each invalid frame uses approximately 100 characters.
Using the same file-level token protocol as in the measured statistics ($4$ characters $\approx 1$ token, 2,125 tokens per $1920{\times}1080$ image, 550 static prompt tokens per frame-pair call, and a 2,728-token consolidated trajectory), the 127,200 frame-pair VPA input tokens are composed of 63,750 image tokens (30 images), 55,200 OmniParser JSON tokens (two parsed element files per call), and 8,250 static prompt tokens.
The remaining video-level inputs come from the two decomposition calls, contributing 3,178 input tokens for planning and 3,178 for grounding, yielding 133,556 total input tokens; the corresponding outputs are 6,350 for frame-pair VPA, 546 for planning, and 1,650 for grounding.

\begin{table}[t]
  \caption{\textbf{Annotation API usage per selected video} in the typical-task regime.
    Costs use GPT-5.1 at \$1.25 / \$10.00 per 1M input / output tokens.}
  \label{tab:appendix-annotation-cost}
  \centering\footnotesize
  \setlength{\tabcolsep}{4pt}
  \begin{tabular}{@{}lcccc@{}}
    \toprule
    Stage & Calls & In/Out Tok. & Price & Cost \\
    \midrule
    Frame-pair VPA & 15 & 127.20K / 6.35K & 1.25 / 10.00 & \$0.2225 \\
    Planning split & 1 & 3.18K / 546 & 1.25 / 10.00 & \$0.0094 \\
    Grounding split & 1 & 3.18K / 1.65K & 1.25 / 10.00 & \$0.0205 \\
    \midrule
    \textbf{Total / video} & \textbf{17} & \textbf{133.56K / 8.55K} & \textbf{1.25 / 10.00} & \textbf{\$0.252} \\
    \bottomrule
  \end{tabular}
\end{table}

\noindent
Thus, annotation dominates the end-to-end spend: one selected video costs about \$0.252 with GPT-5.1, while one retrieval attempt costs only about \$0.019.
Within annotation, the repeated frame-pair pairwise video annotation stage is the main driver, accounting for 127.20K / 6.35K of the total 133.56K / 8.55K input / output tokens and \$0.2225 of the \$0.252 per-video spend.
Under a more demanding \emph{complex-GUI} regime, where valid frames carry about 17K OmniParser characters on average, the per-video total rises to 172,452 input and 8,546 output tokens, corresponding to about \$0.301 with GPT-5.1, \$0.048 with GPT-4.1-mini, \$0.029 with Seed-1.8, and \$0.014 with Qwen3-VL-8B.

\paragraph{Benchmark-scale total cost.}
On OSWorld, retrieval is attempted for all 361 evaluation tasks.
In the full OSWorld run, 299 tasks retrieve at least one relevant video and 42.7\% of those retrieve a second one, yielding approximately 427 selected videos for annotation.
Combining \cref{tab:appendix-retrieval-cost} and \cref{tab:appendix-annotation-cost}, the default end-to-end API spend for the full benchmark is summarized in \cref{tab:appendix-total-cost}.

\begin{table}[t]
  \caption{\textbf{Full benchmark API cost} under the default configuration.}
  \label{tab:appendix-total-cost}
  \centering\footnotesize
  \setlength{\tabcolsep}{4pt}
  \begin{tabular}{@{}lccc@{}}
    \toprule
    Module & Units & Cost & Share \\
    \midrule
    Retrieval & $361 \times \$0.0188$ & \$6.8 & 5.9\% \\
    Annotation & $\sim427 \times \$0.252$ & \$107.8 & 94.1\% \\
    \midrule
    \textbf{Total} & -- & \textbf{\$114.6} & \textbf{100\%} \\
    \bottomrule
  \end{tabular}
\end{table}

\paragraph{Advantage over manual annotation.}
Compared with the standard GUI-agent pipeline of manually annotating task trajectories and then fine-tuning on those demonstrations, \guidename{} is substantially more efficient.
Each manually labeled trajectory requires a human annotator to execute the task, record the correct action sequence, and often revise the trace for training use, making the per-example cost much higher than the API-only cost reported above.
In addition, manual-annotation-and-fine-tuning pipelines are inherently offline: adapting to a new task or a changed software interface requires collecting new demonstrations and repeating data curation and training.
By contrast, \guidename{} retrieves up-to-date web tutorials and converts them into task-specific knowledge automatically at inference time, providing a faster and more practical mechanism for real-time adaptation.

\section{Failure Case Analysis}
\label{sec:appendix-failure}
\begin{figure*}[tb]
  \centering
  \resizebox{0.9\textwidth}{!}{%
  \begin{tikzpicture}[
      x=1mm, y=1mm,
      font=\sffamily,
      badge/.style={fill=carrow, text=white, font=\bfseries\fontsize{7pt}{7pt}\selectfont, rounded corners=3pt, inner sep=3pt, align=center},
      card/.style={draw=none, fill=white, rounded corners=2pt, inner sep=3pt},
      planbox/.style={draw=cplan, fill=cbgplan, rounded corners=4pt, line width=1pt},
      thoughtbox/.style={fill=placeholderBg, rounded corners=3pt, inner sep=3pt, font=\ttfamily\fontsize{6.5pt}{7pt}\selectfont, align=left},
      screenshot/.style={draw=carrow!30, line width=0.5pt, inner sep=0pt, rounded corners=2pt},
      labeltext/.style={font=\bfseries\fontsize{7pt}{7pt}\selectfont, text=ctext},
      subtext/.style={font=\fontsize{6pt}{6.5pt}\selectfont, text=ctext},
      arrowconnect/.style={-{Triangle[length=4pt,width=4pt]}, carrow, dashed, line width=0.8pt},
      cfailbox/.style={draw=cfail, fill=cbgfail, rounded corners=4pt, line width=1pt}
  ]
  
  \definecolor{cplan}{RGB}{31,119,180}
  \definecolor{cground}{RGB}{214,125,36}
  \definecolor{cstage}{RGB}{44,160,44}
  \definecolor{cfail}{RGB}{200,50,50}
  \definecolor{carrow}{RGB}{68,68,68}
  \definecolor{ctext}{RGB}{26,26,26}
  \definecolor{cmuted}{RGB}{120,120,120}
  \definecolor{placeholderBg}{RGB}{245,245,245}
  \definecolor{cbgfail}{RGB}{253,236,234}
  \definecolor{cwarn}{RGB}{255,127,14}
  \definecolor{cbgwarn}{RGB}{255,243,235}
  
  \node[fill=placeholderBg, rounded corners=4pt, minimum width=174mm, inner sep=2.5mm, anchor=north west] (header) at (0, 137) {
      \begin{tabular}{p{168mm}}
      \fontsize{7.5pt}{9pt}\selectfont \textbf{Task Instruction:} ``Can you enhance this low-resolution photo to high-resolution without increasing file size?''\\
      \fontsize{7.5pt}{9pt}\selectfont \textbf{App:} GIMP 2.10 \hspace{3mm} \textbf{Retrieved Video:} ``How to Get Higher Quality Image Prints with Resolution''
      \end{tabular}
  };
  
  \node[anchor=north west, font=\fontsize{9pt}{9pt}\bfseries\selectfont] at (0, 123) {(a) Planning Mismatch $\to$ Wrong Goal (ppi vs quality)};
  
  \fill[cbgfail, rounded corners=4pt] (0, 65) rectangle (174, 118);
  \draw[cfail, line width=1pt, rounded corners=4pt] (0, 65) rectangle (174, 118);
  
  \node[fill=white, draw=cfail!30, rounded corners=3pt, minimum width=43mm, minimum height=49mm, anchor=north west] (pk) at (1.5, 116.5) {};
  \fill[cfail, rounded corners=2pt] (1.5, 116.5) rectangle (2.3, 67.5); 
  \node[fill=cfail, text=white, rounded corners=2pt, font=\bfseries\fontsize{6.5pt}{6.5pt}\selectfont, anchor=north west] at (3, 115) {Planning $\times$ Mismatch};
  
  \node[anchor=north west, text width=38mm, align=left, font=\fontsize{7pt}{8pt}\selectfont] at (3, 110) {
      \textbf{Execution Flow:} (Retrieved)\\[0.5mm]
      1. Open image window in GIMP\\[0.5mm]
      \colorbox{cfail!15}{\parbox{35mm}{\textbf{2. Open Image $\to$ Scale}\\ \textbf{Image... review X/Y ppi} \textcolor{cfail}{$\blacktriangleleft$}}} \\[0.5mm]
      \colorbox{cfail!15}{\parbox{35mm}{\textbf{3. Change ppi to 150ppi}\\ \textbf{for print output} \textcolor{cfail}{$\blacktriangleleft$}}} \\[0.5mm]
      4. Click Scale to apply\\[1.5mm]
      \textbf{Task Needs vs Retrieved:}\\
      \textcolor{cfail}{\textbf{$\times$} Needs visual enhancement}\\
      \textcolor{cfail}{(sharpen), NOT print ppi tweak.}
  };

  \node[anchor=north west, inner sep=0] (screen1) at (49, 116) {
      \includegraphics[width=78mm,height=44mm,keepaspectratio]{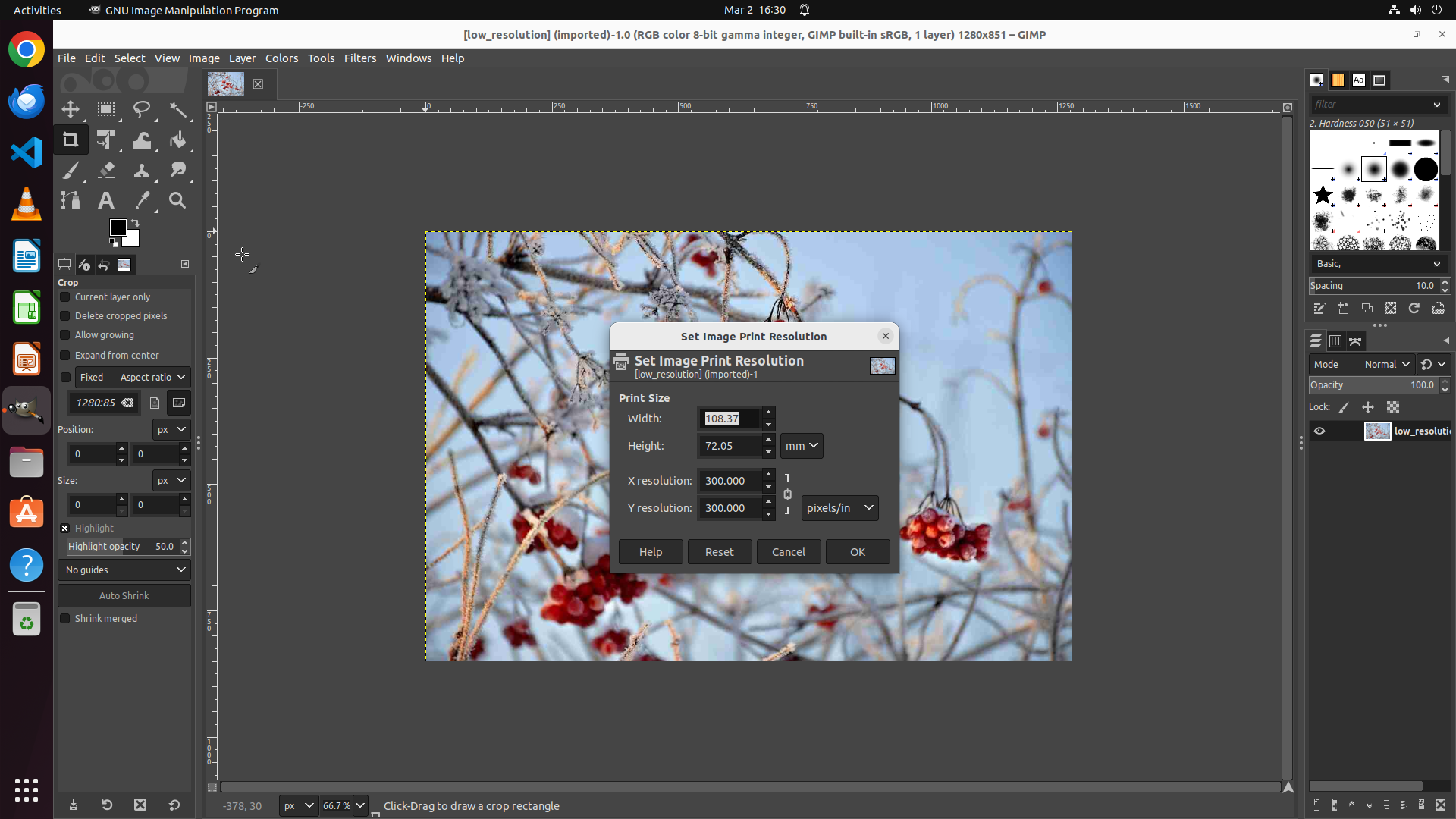}
  };
  \draw[carrow!30, line width=0.5pt, rounded corners=2pt] (screen1.south west) rectangle (screen1.north east);
  \node[fill=white, draw=carrow, rounded corners=2pt, inner sep=1.5pt, font=\bfseries\fontsize{5.5pt}{5.5pt}\selectfont, anchor=north east] at ($ (screen1.north east) + (-1, -1) $) {Step 2 / 8};

  \begin{scope}[shift={(screen1.south west)}, x={(screen1.south east)}, y={(screen1.north west)}]
      \draw[cfail, dashed, line width=1.5pt] (0.5, 0.45) ellipse (0.35 and 0.2);
      \node[fill=cfail, text=white, rounded corners=1pt, font=\bfseries\fontsize{5pt}{5pt}\selectfont, anchor=south] at (0.5, 0.65) {$\times$ ppi check (dead end)};
  \end{scope}
  
  \node[thoughtbox, text width=38mm, anchor=north west, minimum height=49mm] (thought1) at (133.5, 116.5) {
      \textbf{\textcolor{carrow}{Agent Thought:}}\\[1mm]
      The task is to enhance this photo...\\
      According to retrieved planning:\\[1.5mm]
      \colorbox{cbgwarn}{\parbox{34mm}{"The user opens Image $\to$ Scale Image... review X/Y resolution... choose ppi..."}}\\[1.5mm]
      Let me open Image $\to$ Print Size.\\[1mm]
      \colorbox{cfail!15}{\parbox{34mm}{The image is already at 300ppi. Adjusting ppi won't help.\\Planning path is a dead end.}}\\[1mm]
      \textbf{Action:} Fallback to Filters $\to$\\Enhance $\to$ Unsharp Mask\\
      \textcolor{cfail}{[Score 0: Failed]}
  };
  
  \begin{scope}[shift={([xshift=-7mm, yshift=-6.0mm]thought1.north east)}, scale=2.4]
      \fill[cyan!40, rounded corners=2pt] (-1.2, -2.5) rectangle (1.2, -0.2);
      \fill[cyan!80!black] (-0.8, -2) rectangle (0.8, -0.7);
      \draw[cfail, line width=1.5pt] (-0.3, -1.1) -- (0.3, -1.7);
      \draw[cfail, line width=1.5pt] (-0.3, -1.7) -- (0.3, -1.1);
      \fill[cyan!40, rounded corners=1.5pt] (-1.7, -0.5) rectangle (-1.2, -1.8);
      \fill[cyan!40, rounded corners=1.5pt] (1.2, -0.5) rectangle (1.7, -1.8);
      \fill[gray!50] (-0.4, -0.2) rectangle (0.4, 0);
      \fill[cyan!40, rounded corners=3pt] (-1.4, 0) rectangle (1.4, 2.2);
      \fill[gray!50, rounded corners=1pt] (-1.7, 0.6) rectangle (-1.4, 1.4);
      \fill[gray!50, rounded corners=1pt] (1.4, 0.6) rectangle (1.7, 1.4);
      \fill[white] (-0.6, 1.2) circle (0.4);
      \fill[white] (0.6, 1.2) circle (0.4);
      \fill[cyan!80!black] (-0.6, 1.2) circle (0.15);
      \fill[cyan!80!black] (0.6, 1.2) circle (0.15);
      \draw[cyan!80!black, line width=1.2pt] (-0.4, 0.3) .. controls (-0.1, 0.5) and (0.1, 0.5) .. (0.4, 0.3);
      \draw[gray!80, line width=1.5pt] (0, 2.2) -- (0, 3.0);
      \fill[orange] (0, 3.2) circle (0.25);
  \end{scope}

  \draw[-{Triangle}, cfail, dashed, line width=0.8pt] (pk.east) -- (screen1.west);
  \draw[-{Triangle}, cfail, dashed, line width=0.8pt] (screen1.east) -- (thought1.west);

  \draw[carrow!50, dashed, line width=1pt] (0, 60) -- (174, 60);

  \node[fill=placeholderBg, rounded corners=4pt, minimum width=174mm, inner sep=2.5mm, anchor=north west] (headerb) at (0, 56) {
      \begin{tabular}{p{168mm}}
      \fontsize{7.5pt}{9pt}\selectfont \textbf{Task Instruction:} ``Could you turn my image into CYMK mode within GIMP?''\\
      \fontsize{7.5pt}{9pt}\selectfont \textbf{App:} GIMP 2.10 \hspace{3mm} \textbf{Retrieved Video:} ``0501 Gimp RGB to CMYK for color Separations''
      \end{tabular}
  };

  \node[anchor=north west, font=\fontsize{9pt}{9pt}\bfseries\selectfont] at (0, 42) {(b) Grounding Source Mismatch $\to$ Browser vs GIMP Detour};

  \fill[cbgfail, rounded corners=4pt] (0, -16) rectangle (174, 37);
  \draw[cfail, line width=1pt, rounded corners=4pt] (0, -16) rectangle (174, 37);

  \node[fill=white, draw=cfail!30, rounded corners=3pt, minimum width=43mm, minimum height=49mm, anchor=north west] (gk) at (1.5, 35.5) {};
  \fill[cfail, rounded corners=2pt] (1.5, 35.5) rectangle (2.3, -13.5); 
  \node[fill=cfail, text=white, rounded corners=2pt, font=\bfseries\fontsize{6.5pt}{6.5pt}\selectfont, anchor=north west] at (3, 34) {Grounding $\times$ Mismatch};

  \node[anchor=north west, text width=38mm, align=left, font=\fontsize{7pt}{8pt}\selectfont] at (3, 29) {
      \textbf{Described Elements:}\\
      $\triangleright$ E1: "Partha's PL... Tab" (Browser)\\
      $\triangleright$ E2: "Address Bar" (Browser)\\
      $\triangleright$ E3: "Search Field: adobe cmyk..."\\
      $\triangleright$ E4: "Download Now Button"\\[1.5mm]
      \textbf{Actual GIMP UI state:}\\
      \textcolor{cfail}{\textbf{$\times$} GIMP environment only.}\\
      \textcolor{cfail}{No browser UI elements exist.}\\
      \textcolor{cfail}{Agent thinks it must open}\\
      \textcolor{cfail}{browser to align with video.}
  };

  \node[anchor=north west, inner sep=0] (screen2) at (49, 35) {
      \includegraphics[width=78mm,height=44mm,keepaspectratio]{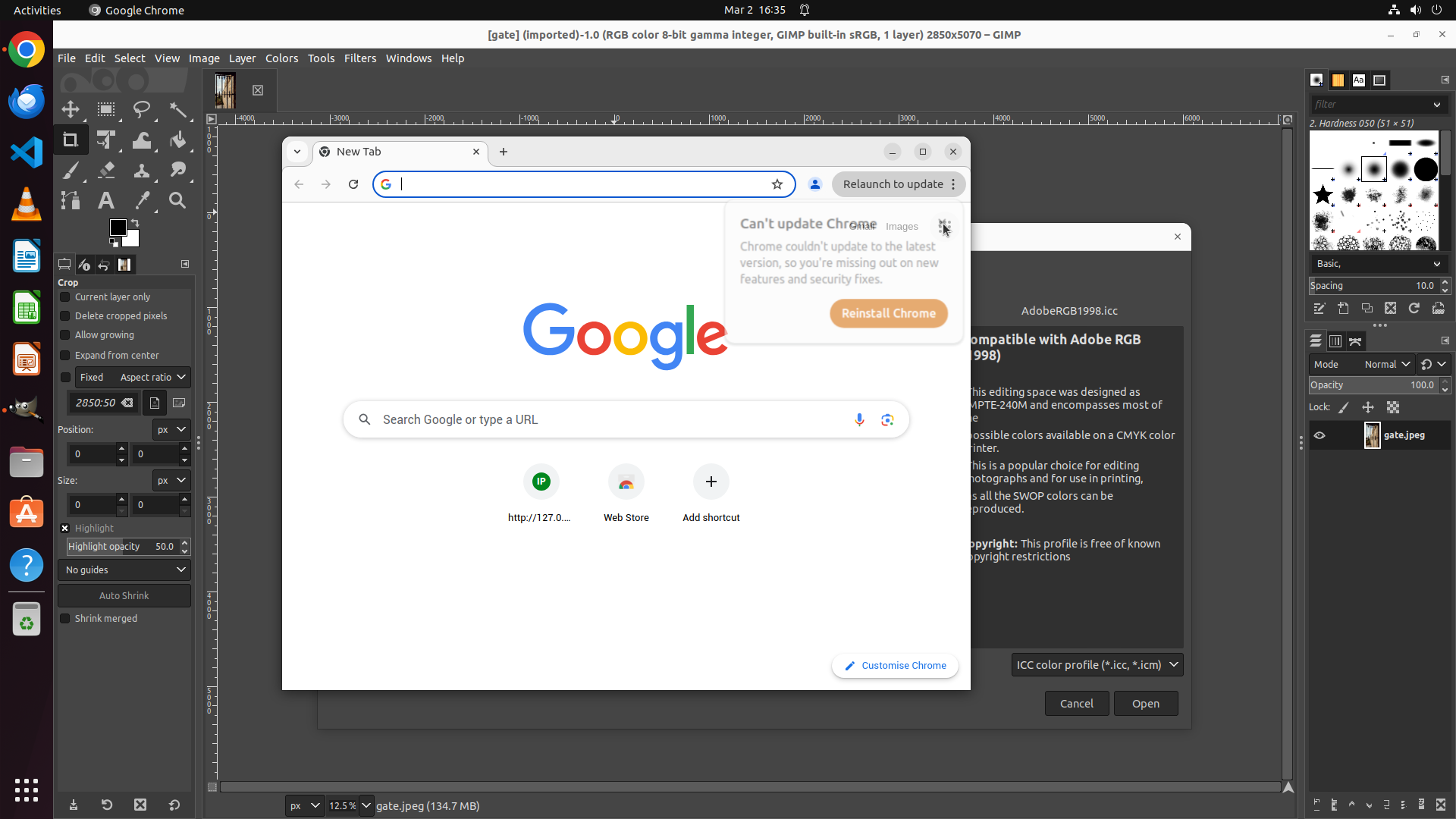}
  };
  \draw[carrow!30, line width=0.5pt, rounded corners=2pt] (screen2.south west) rectangle (screen2.north east);
  \node[fill=white, draw=carrow, rounded corners=2pt, inner sep=1.5pt, font=\bfseries\fontsize{5.5pt}{5.5pt}\selectfont, anchor=north east] at ($ (screen2.north east) + (-1, -1) $) {Step 14 / 50};

  \begin{scope}[shift={(screen2.south west)}, x={(screen2.south east)}, y={(screen2.north west)}]
      \draw[cfail, dashed, line width=1.5pt, rounded corners=1.5pt] (0.18, 0.15) rectangle (0.68, 0.84);
      \node[fill=cfail!90, text=white, rounded corners=1.5pt, font=\bfseries\fontsize{6.5pt}{6.5pt}\selectfont, inner sep=2pt, align=center] at (0.5, 0.5) {$\times$ Browser opened needlessly};
  \end{scope}
  
  \node[thoughtbox, text width=38mm, anchor=north west, minimum height=49mm] (thought2) at (133.5, 35.5) {
      \textbf{\textcolor{carrow}{Agent Thought:}}\\[1mm]
      No CMYK profiles found locally.\\[1mm]
      \colorbox{cfail!15}{\parbox{34mm}{[Retrieved grounding describes browser UI: "Address Bar", "Search Field"... none exist in GIMP.]}}\\[1mm]
      The video reference said to get Adobe CMYK profiles. So we need to open Chrome to download them.\\[1mm]
      \textbf{Action:} Click Chrome icon $\to$\\
      Open browser $\to$ Search "Adobe..."\\
      \textcolor{cfail}{[50 steps failed detour]}
  };
  
  \begin{scope}[shift={([xshift=-7mm, yshift=-6.0mm]thought2.north east)}, scale=2.4]
      \fill[cyan!40, rounded corners=2pt] (-1.2, -2.5) rectangle (1.2, -0.2);
      \fill[cyan!80!black] (-0.8, -2) rectangle (0.8, -0.7);
      \draw[cfail, line width=1.5pt] (-0.3, -1.1) -- (0.3, -1.7);
      \draw[cfail, line width=1.5pt] (-0.3, -1.7) -- (0.3, -1.1);
      \fill[cyan!40, rounded corners=1.5pt] (-1.7, -0.5) rectangle (-1.2, -1.8);
      \fill[cyan!40, rounded corners=1.5pt] (1.2, -0.5) rectangle (1.7, -1.8);
      \fill[gray!50] (-0.4, -0.2) rectangle (0.4, 0);
      \fill[cyan!40, rounded corners=3pt] (-1.4, 0) rectangle (1.4, 2.2);
      \fill[gray!50, rounded corners=1pt] (-1.7, 0.6) rectangle (-1.4, 1.4);
      \fill[gray!50, rounded corners=1pt] (1.4, 0.6) rectangle (1.7, 1.4);
      \fill[white] (-0.6, 1.2) circle (0.4);
      \fill[white] (0.6, 1.2) circle (0.4);
      \fill[cyan!80!black] (-0.6, 1.2) circle (0.15);
      \fill[cyan!80!black] (0.6, 1.2) circle (0.15);
      \draw[cyan!80!black, line width=1.2pt] (-0.4, 0.3) .. controls (-0.1, 0.5) and (0.1, 0.5) .. (0.4, 0.3);
      \draw[gray!80, line width=1.5pt] (0, 2.2) -- (0, 3.0);
      \fill[orange] (0, 3.2) circle (0.25);
  \end{scope}

  \draw[-{Triangle}, cfail, dashed, line width=0.8pt] (gk.east) -- (screen2.west);
  \draw[-{Triangle}, cfail, dashed, line width=0.8pt] (screen2.east) -- (thought2.west);

  \node[anchor=south east, font=\bfseries\fontsize{8.5pt}{8.5pt}\selectfont, text=cfail] at (172, 67) {
      \Huge \textcolor{cfail}{$\times$} \normalsize Task Failed (0.0)\quad
      \mdseries\itshape\fontsize{7.5pt}{7.5pt}\selectfont 8 steps wasted on wrong workflow
  };

  \node[anchor=south east, font=\bfseries\fontsize{8.5pt}{8.5pt}\selectfont, text=cfail] at (172, -14) {
      \Huge \textcolor{cfail}{$\times$} \normalsize Task Failed (0.0)\quad
      \mdseries\itshape\fontsize{7.5pt}{7.5pt}\selectfont 50 steps lost to browser detour
  };

  \end{tikzpicture}
  }
  \caption{\textbf{Failure Cases: When Annotations Mislead the Agent.}
  \textbf{(a)} \textbf{Planning mismatch:} The retrieved video targets print-resolution (ppi) tuning while the task requires pixel-level visual quality enhancement without increasing file size.
  Following the retrieved planning, the agent opens the Print Resolution dialog---only to find the image is already at 300\,ppi, a dead end.
  It falls back to Unsharp Mask but still fails (8 steps wasted).
  \textbf{(b)} \textbf{Grounding mismatch:} The retrieved video demonstrates CMYK profile installation via a web browser; the grounding knowledge therefore describes browser UI elements (tabs, address bar, search field) entirely absent from the GIMP-only task environment.
  The agent opens Chrome to download the profiles and exhausts all 50 steps in a fruitless browser detour.
  }
  \label{fig:failure}
\end{figure*}

\begin{figure*}[htbp]
  \centering
  \resizebox{0.9\textwidth}{!}{%
  \begin{tikzpicture}[
      x=1mm, y=1mm,
      font=\sffamily,
      badge/.style={fill=carrow, text=white, font=\bfseries\fontsize{7pt}{7pt}\selectfont, rounded corners=3pt, inner sep=3pt, align=center},
      card/.style={draw=none, fill=white, rounded corners=2pt, inner sep=3pt},
      planbox/.style={draw=cplan, fill=cbgplan, rounded corners=4pt, line width=1pt},
      thoughtbox/.style={fill=placeholderBg, rounded corners=3pt, inner sep=3pt, font=\ttfamily\fontsize{6.5pt}{7pt}\selectfont, align=left},
      screenshot/.style={draw=carrow!30, line width=0.5pt, inner sep=0pt, rounded corners=2pt},
      labeltext/.style={font=\bfseries\fontsize{7pt}{7pt}\selectfont, text=ctext},
      subtext/.style={font=\fontsize{6pt}{6.5pt}\selectfont, text=ctext},
      arrowconnect/.style={-{Triangle[length=4pt,width=4pt]}, carrow, dashed, line width=0.8pt},
      cfailbox/.style={draw=cfail, fill=cbgfail, rounded corners=4pt, line width=1pt}
  ]
  
  \definecolor{cplan}{RGB}{31,119,180}
  \definecolor{cground}{RGB}{214,125,36}
  \definecolor{cstage}{RGB}{44,160,44}
  \definecolor{cfail}{RGB}{200,50,50}
  \definecolor{carrow}{RGB}{68,68,68}
  \definecolor{ctext}{RGB}{26,26,26}
  \definecolor{cmuted}{RGB}{120,120,120}
  \definecolor{placeholderBg}{RGB}{245,245,245}
  \definecolor{cbgfail}{RGB}{253,236,234}
  \definecolor{cwarn}{RGB}{255,127,14}
  \definecolor{cbgwarn}{RGB}{255,243,235}
  
  \node[fill=placeholderBg, rounded corners=4pt, minimum width=174mm, inner sep=2.5mm, anchor=north west] (header) at (0, 137) {
      \begin{tabular}{p{168mm}}
      \fontsize{7.5pt}{9pt}\selectfont \textbf{Task Instruction:} ``Convert my new raw image into jpeg.'' \hspace{2mm} \textbf{App:} GIMP 2.10\\
      \fontsize{7.5pt}{9pt}\selectfont \textbf{Retrieved Video:} ``How to Open RAW Photos with GIMP \& Darktable or RawTherapee''
      \end{tabular}
  };
  
  \node[anchor=north west, font=\fontsize{9pt}{9pt}\bfseries\selectfont] at (0, 123) {(a) Planning Mismatch $\to$ Wrong Workflow};
  
  \fill[cbgfail, rounded corners=4pt] (0, 65) rectangle (174, 118);
  \draw[cfail, line width=1pt, rounded corners=4pt] (0, 65) rectangle (174, 118);
  
  \node[fill=white, draw=cfail!30, rounded corners=3pt, minimum width=43mm, minimum height=49mm, anchor=north west] (pk) at (1.5, 116.5) {};
  \fill[cfail, rounded corners=2pt] (1.5, 116.5) rectangle (2.3, 67.5); 
  \node[fill=cfail, text=white, rounded corners=2pt, font=\bfseries\fontsize{6.5pt}{6.5pt}\selectfont, anchor=north west] at (3, 115) {Planning $\times$ Mismatch};
  
  \node[anchor=north west, text width=38mm, align=left, font=\fontsize{7pt}{8pt}\selectfont] at (3, 110) {
      \textbf{Execution Flow:} (Retrieved)\\[0.5mm]
      1. Research: browse RawTherapee...\\[0.5mm]
      \colorbox{cfail!15}{\parbox{35mm}{\textbf{2. Install darktable as RAW}\\ \textbf{processor plugin} \textcolor{cfail}{$\blacktriangleleft$}}} \\[0.5mm]
      3. Open "Open Image" dialog...\\[2mm]
      \textbf{Task Needs vs Retrieved:}\\
      \textcolor{cfail}{\textbf{$\times$} Needs format conversion}\\
      \textcolor{cfail}{(Export As), NOT plugin install.}
  };

  \node[anchor=north west, inner sep=0] (screen1) at (49, 116) {
      \includegraphics[width=78mm]{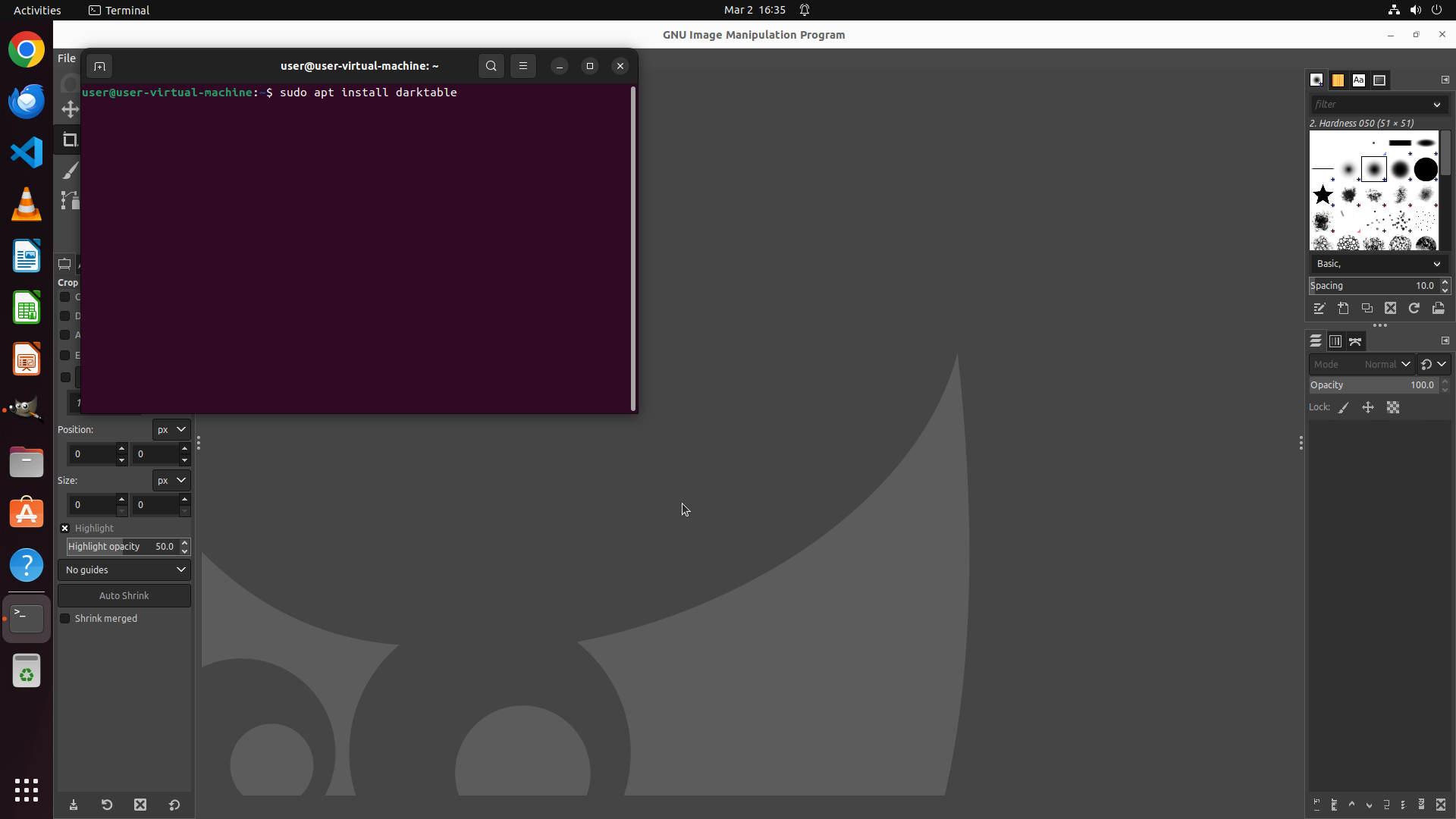}
  };
  \draw[carrow!30, line width=0.5pt, rounded corners=2pt] (screen1.south west) rectangle (screen1.north east);
  \node[fill=white, draw=carrow, rounded corners=2pt, inner sep=1.5pt, font=\bfseries\fontsize{5.5pt}{5.5pt}\selectfont, anchor=north east] at ($ (screen1.north east) + (-1, -1) $) {Step 7 / 13};

  \begin{scope}[shift={(screen1.south west)}, x={(screen1.south east)}, y={(screen1.north west)}]
      \draw[cfail, dashed, line width=1.5pt, rounded corners=1.5pt] (0.052, 0.492) rectangle (0.438, 0.938);
      \node[fill=cfail, text=white, rounded corners=1pt, font=\bfseries\fontsize{5pt}{5pt}\selectfont, anchor=north] at (0.245, 0.478) {$\times$ Wrong action};
  \end{scope}
  
  \node[thoughtbox, text width=38mm, anchor=north west, minimum height=49mm] (thought1) at (133.5, 116.5) {
      \textbf{\textcolor{carrow}{Agent Thought:}}\\[1mm]
      The GIMP error says no RAW loader is installed.\\
      According to the retrieved planning:\\[1.5mm]
      \colorbox{cbgwarn}{\parbox{34mm}{"The workflow begins by gathering information... load Darktable installation page..."}}\\[1.5mm]
      Following this, I need to install darktable as the RAW loader.\\[1.5mm]
      \colorbox{cfail!15}{\parbox{34mm}{\textbf{Action:} Open terminal $\to$\\sudo apt install darktable\\{[Password rejected]}}}
  };
  
  \begin{scope}[shift={([xshift=-7mm, yshift=-6.0mm]thought1.north east)}, scale=2.4]
      \fill[cyan!40, rounded corners=2pt] (-1.2, -2.5) rectangle (1.2, -0.2);
      \fill[cyan!80!black] (-0.8, -2) rectangle (0.8, -0.7);
      \draw[cfail, line width=1.5pt] (-0.3, -1.1) -- (0.3, -1.7);
      \draw[cfail, line width=1.5pt] (-0.3, -1.7) -- (0.3, -1.1);
      \fill[cyan!40, rounded corners=1.5pt] (-1.7, -0.5) rectangle (-1.2, -1.8);
      \fill[cyan!40, rounded corners=1.5pt] (1.2, -0.5) rectangle (1.7, -1.8);
      \fill[gray!50] (-0.4, -0.2) rectangle (0.4, 0);
      \fill[cyan!40, rounded corners=3pt] (-1.4, 0) rectangle (1.4, 2.2);
      \fill[gray!50, rounded corners=1pt] (-1.7, 0.6) rectangle (-1.4, 1.4);
      \fill[gray!50, rounded corners=1pt] (1.4, 0.6) rectangle (1.7, 1.4);
      \fill[white] (-0.6, 1.2) circle (0.4);
      \fill[white] (0.6, 1.2) circle (0.4);
      \fill[cyan!80!black] (-0.6, 1.2) circle (0.15);
      \fill[cyan!80!black] (0.6, 1.2) circle (0.15);
      \draw[cyan!80!black, line width=1.2pt] (-0.4, 0.3) .. controls (-0.1, 0.5) and (0.1, 0.5) .. (0.4, 0.3);
      \draw[gray!80, line width=1.5pt] (0, 2.2) -- (0, 3.0);
      \fill[orange] (0, 3.2) circle (0.25);
  \end{scope}

  \draw[-{Triangle}, cfail, dashed, line width=0.8pt] (pk.east) -- (screen1.west);
  \draw[-{Triangle}, cfail, dashed, line width=0.8pt] (screen1.east) -- (thought1.west);

  \draw[carrow!50, dashed, line width=1pt] (0, 60) -- (174, 60);

  \node[fill=placeholderBg, rounded corners=4pt, minimum width=174mm, inner sep=2.5mm, anchor=north west] (headerb) at (0, 56) {
      \begin{tabular}{p{168mm}}
      \fontsize{7.5pt}{9pt}\selectfont \textbf{Task Instruction:} ``Can you assist me in shifting the text box to the left?...'' \hspace{2mm} \textbf{App:} GIMP 2.10\\
      \fontsize{7.5pt}{9pt}\selectfont \textbf{Retrieved Video:} ``Moving Text Boxes in GIMP'' (\textit{Animated tutorial slides video, NOT actual GIMP operation})
      \end{tabular}
  };

  \node[anchor=north west, font=\fontsize{9pt}{9pt}\bfseries\selectfont] at (0, 42) {(b) Grounding Mismatch $\to$ Unusable UI Descriptions};

  \fill[cbgfail, rounded corners=4pt] (0, -16) rectangle (174, 37);
  \draw[cfail, line width=1pt, rounded corners=4pt] (0, -16) rectangle (174, 37);

  \node[fill=white, draw=cfail!30, rounded corners=3pt, minimum width=43mm, minimum height=49mm, anchor=north west] (gk) at (1.5, 35.5) {};
  \fill[cfail, rounded corners=2pt] (1.5, 35.5) rectangle (2.3, -13.5); 
  \node[fill=cfail, text=white, rounded corners=2pt, font=\bfseries\fontsize{6.5pt}{6.5pt}\selectfont, anchor=north west] at (3, 34) {Grounding $\times$ Mismatch};

  \node[anchor=north west, text width=38mm, align=left, font=\fontsize{7pt}{8pt}\selectfont] at (3, 29) {
      \textbf{Described Elements:}\\
      $\triangleright$ E1: "Main Title Header Box"\\
      $\triangleright$ E3: "Slide Navigation Button"\\
      $\triangleright$ E13: "THANK YOU Box"\\[2.5mm]
      \textbf{Actual GIMP UI state:}\\
      \textcolor{cfail}{\textbf{$\times$} Canvas \& Layers pane only.}\\
      \textcolor{cfail}{NONE of the 13 described}\\
      \textcolor{cfail}{grounding elements match}\\
      \textcolor{cfail}{any item in the workspace.}
  };

  \node[anchor=north west, inner sep=0] (screen2) at (49, 35) {
      \includegraphics[width=78mm]{figures/e2dd0213_step1_gimp_textlayer.png}
  };
  \draw[carrow!30, line width=0.5pt, rounded corners=2pt] (screen2.south west) rectangle (screen2.north east);

  \begin{scope}[shift={(screen2.south west)}, x={(screen2.south east)}, y={(screen2.north west)}]
      \draw[cfail, dashed, line width=1.5pt, rounded corners=1.5pt] (0.02, 0.02) rectangle (0.98, 0.98);
      \node[fill=cfail!90, text=white, rounded corners=1.5pt, font=\bfseries\fontsize{6.5pt}{6.5pt}\selectfont, inner sep=2pt, align=center] at (0.5, 0.5) {Grounding elements\\NOT FOUND in UI};
  \end{scope}
  
  \node[thoughtbox, text width=38mm, anchor=north west, minimum height=49mm] (thought2) at (133.5, 35.5) {
      \textbf{\textcolor{carrow}{Agent Thought:}}\\[1mm]
      I need to shift the "Thanks World" text box left.\\[1mm]
      \colorbox{cfail!15}{\parbox{34mm}{[Retrieved grounding describes: tutorial title cards, slide nav buttons... NONE exist here. Info unusable.]}}\\[1mm]
      Using basic knowledge: select Move Tool from left toolbar.\\[1mm]
      \textbf{Action:} Drag text:\\
      \texttt{<drag> [538, 634]} \texttt{\scalebox{0.8}{$\to$}} \texttt{[338, 634]}\\[0.5mm]
      \textcolor{cfail}{[Evaluator result: FAILED]}
  };
  
  \begin{scope}[shift={([xshift=-7mm, yshift=-6.0mm]thought2.north east)}, scale=2.4]
      \fill[cyan!40, rounded corners=2pt] (-1.2, -2.5) rectangle (1.2, -0.2);
      \fill[cyan!80!black] (-0.8, -2) rectangle (0.8, -0.7);
      \draw[cfail, line width=1.5pt] (-0.3, -1.1) -- (0.3, -1.7);
      \draw[cfail, line width=1.5pt] (-0.3, -1.7) -- (0.3, -1.1);
      \fill[cyan!40, rounded corners=1.5pt] (-1.7, -0.5) rectangle (-1.2, -1.8);
      \fill[cyan!40, rounded corners=1.5pt] (1.2, -0.5) rectangle (1.7, -1.8);
      \fill[gray!50] (-0.4, -0.2) rectangle (0.4, 0);
      \fill[cyan!40, rounded corners=3pt] (-1.4, 0) rectangle (1.4, 2.2);
      \fill[gray!50, rounded corners=1pt] (-1.7, 0.6) rectangle (-1.4, 1.4);
      \fill[gray!50, rounded corners=1pt] (1.4, 0.6) rectangle (1.7, 1.4);
      \fill[white] (-0.6, 1.2) circle (0.4);
      \fill[white] (0.6, 1.2) circle (0.4);
      \fill[cyan!80!black] (-0.6, 1.2) circle (0.15);
      \fill[cyan!80!black] (0.6, 1.2) circle (0.15);
      \draw[cyan!80!black, line width=1.2pt] (-0.4, 0.3) .. controls (-0.1, 0.5) and (0.1, 0.5) .. (0.4, 0.3);
      \draw[gray!80, line width=1.5pt] (0, 2.2) -- (0, 3.0);
      \fill[orange] (0, 3.2) circle (0.25);
  \end{scope}

  \draw[-{Triangle}, cfail, dashed, line width=0.8pt] (gk.east) -- (screen2.west);
  \draw[-{Triangle}, cfail, dashed, line width=0.8pt] (screen2.east) -- (thought2.west);

  \node[anchor=south east, font=\bfseries\fontsize{8.5pt}{8.5pt}\selectfont, text=cfail] at (172, 67) {
      \Huge \textcolor{cfail}{$\times$} \normalsize Task Failed (0.0)\quad
      \mdseries\itshape\fontsize{7.5pt}{7.5pt}\selectfont 13 steps wasted on wrong workflow
  };

  \node[anchor=south east, font=\bfseries\fontsize{8.5pt}{8.5pt}\selectfont, text=cfail] at (172, -14) {
      \Huge \textcolor{cfail}{$\times$} \normalsize Task Failed (0.0)\quad
      \mdseries\itshape\fontsize{7.5pt}{7.5pt}\selectfont 0 usable UI descriptions
  };

  \end{tikzpicture}
  }
  \caption{\textbf{Additional Failure Cases.}
  \textbf{(a)} \textbf{Planning mismatch:} The retrieved video describes installing a RAW-processing plugin (darktable) rather than direct format conversion.
  The agent follows the retrieved planning and wastes 13 steps executing terminal commands to install software before ultimately failing.
  \textbf{(b)} \textbf{Grounding mismatch:} The retrieved video is an animated tutorial slide deck rather than an actual GIMP screencast.
  The grounding knowledge describes presentation elements (title cards, slide navigation buttons) absent from the real GIMP workspace; the agent falls back on basic knowledge and fails the task.
  }
  \label{fig:failure-additional}
\end{figure*}

In this section, we present concrete failure examples where injected knowledge misleads the agent into erroneous task execution.
We identify two primary failure modes:
(1)~\emph{planning mismatch}, where retrieved planning knowledge prescribes an incorrect workflow or goal for the given task; and
(2)~\emph{grounding mismatch}, where retrieved grounding knowledge describes UI elements absent from the actual task environment.
\Cref{fig:failure} illustrates representative examples of each mode.
In panel~(a), the retrieved video targets print-resolution (ppi) tuning while the task requires visual quality enhancement without increasing file size;
the agent follows the planning into a dead end, wasting 8 steps.
In panel~(b), the retrieved video covers CMYK profile installation via a web browser, so grounding describes browser-only UI elements;
finding no browser in the GIMP-only environment, the agent opens Chrome itself and exhausts all 50 steps in a fruitless detour.
\Cref{fig:failure-additional} presents additional examples of the same two failure modes with distinct characteristics.
In panel~(a), planning knowledge describes a plugin-installation workflow (darktable) while the task is direct RAW-to-JPEG conversion;
13 steps are wasted on terminal commands.
In panel~(b), the retrieved video is an animated slide deck rather than a software screencast;
grounding describes presentation elements (title cards, navigation buttons) entirely absent from the GIMP workspace,
leaving the agent without actionable visual guidance.

\paragraph{Discussion.}
These failure cases collectively reveal \guidename{}'s dependence on retrieval quality.
When the retrieval engine selects a semantically adjacent but procedurally mismatched video---as in the ppi-tuning case (\cref{fig:failure}(a)) or the plugin-installation case (\cref{fig:failure-additional}(a))---the injected planning knowledge directly derails the agent by prescribing the wrong goal or workflow.
When the retrieved video's format fundamentally differs from a standard software screencast---whether it involves browser-based downloads (\cref{fig:failure}(b)) or animated slide presentations (\cref{fig:failure-additional}(b))---the grounding stage produces descriptions of UI elements absent from the task environment, leaving the agent without actionable guidance.
These observations motivate two complementary mitigations:
(1)~stricter procedural consistency filtering at the semantic retrieval stage to exclude videos whose described workflow differs from the task objective, and
(2)~video format classifiers to screen out non-screencast sources prior to annotation.

\end{document}